\documentclass{article}

\usepackage[main,final]{neurips_2026}

\usepackage{enumitem}

\usepackage[utf8]{inputenc} %
\usepackage[T1]{fontenc}    %
\usepackage{hyperref}       %
\usepackage{url}            %
\usepackage{booktabs}       %
\usepackage{amsfonts}       %
\usepackage{nicefrac}       %
\usepackage{microtype}      %
\usepackage{xcolor}         %
\usepackage{some_macros}
\usepackage{hyperref}
\usepackage{url}
\usepackage{enumitem,kantlipsum}
\usepackage{todonotes}
\usepackage{amsthm}
\usepackage{wrapfig}

\usepackage{amsthm,amsfonts,amsmath,amssymb,epsfig,color,float,graphicx,verbatim}
\usepackage{hyperref}
\usepackage{cleveref}

\hypersetup{
	colorlinks   = true, %
	urlcolor     = blue, %
	linkcolor    = blue, %
	citecolor   = black %
}
\usepackage{bbm}
\usepackage{newtxtext}
\usepackage[varbb]{newtxmath}
\usepackage{caption}
\usepackage{subcaption}
\usepackage{enumitem}
\usepackage{multicol}
\usepackage{mathtools}
\usepackage{tikz}
\usetikzlibrary{arrows.meta,fit,backgrounds,positioning}

\newtheorem{theorem}{Theorem}[section]

\newtheorem{lemma}[theorem]{Lemma}
\newtheorem{corollary}[theorem]{Corollary}
\newtheorem{definition}[theorem]{Definition}

\newtheorem{assumption}{Assumption}
\newtheorem*{assumption*}{Assumption}
\Crefname{assumption}{assumption}{assumptions}
\Crefname{assumption}{Assumption}{Assumptions}
\Crefname{assumption*}{assumption}{assumptions}

\newcommand{\reals}{\mathbb{R}}

\newcommand{\be}{\mathbf{e}}
\newcommand{\bx}{\mathbf{x}}
\newcommand{\bw}{\mathbf{w}}

\newcommand{\bv}{\mathbf{v}}

\newcommand{\bh}{\mathbf{h}}

\newcommand{\bq}{\mathbf{q}}
\newcommand{\bp}{\mathbf{p}}

\newcommand{\loss}{\mathcal{L}}

\newcommand{\norm}[1]{\left\|#1\right\|}
\newcommand{\inner}[1]{\left\langle#1\right\rangle}

\newcommand{\abs}[1]{\left|#1\right|}

\newcommand{\kl}[2]{\mathrm{KL}\left(#1\parallel #2\right)}
\newcommand{\conv}{\mathrm{Conv}}

\renewcommand{\eqref}[1]{Eq.~(\ref{#1})}

\newcommand{\naturals}{\mathbb{N}}

\makeatletter
\newcommand{\printfnsymbol}[1]{%
  \textsuperscript{\@fnsymbol{#1}}%
}
\makeatother

\usetikzlibrary{positioning,decorations.pathreplacing}
\title{Implicit Bias in State Space Models and \\
Linear Autoregressive Training}

\author{%
  Gal Katzhendler\\
  Faculty of Computer Science\\
  Hebrew University of Jerusalem, Israel \\
  \texttt{gal.katzhendler@mail.huji.ac.il} \\
  \And
  Gilad Yehudai \\
  Faculty of Computer Science\\
  Faculty of Computer and Electrical Engineering\\
  Technion, Israel \\
  \And
  Gal Vardi \\
  Department of Computer Science and Applied Mathematics\\
  Weizmann Institute of Science, Israel \\
}

\begin{document}

\maketitle
\begin{abstract}
Linear State Space Models (SSMs) and linear autoregressive sequence models reuse the same parameters across a sample-dependent number of steps. As a result, different examples impose
constraints through different predictors, even though all predictors share the same weights. We study this phenomenon through a general model of multiple non-homogeneous polynomial predictors with shared parameters, in a classification setting. Under standard interpolation and directional-convergence assumptions, together with a trajectory-dependent effective-degree condition, we show that gradient descent on the exponential loss converges in direction to a KKT point of an \emph{effective-degree max-margin problem}.
In this problem, only examples with minimal effective degree impose hard margin constraints; examples with faster-growing margins are asymptotically inactive except for feasibility.
Our result provides a first-order characterization of the implicit bias in linear SSMs and linear autoregressive models. It exposes a mechanism of length bias: a longer input sequence or additional autoregressive steps may increase the example's effective degree, and the examples with the smallest effective degree dominate the limiting classifier. Our result highlights how non-homogeneity and parameter sharing alter classical homogeneous implicit-bias results.
\end{abstract}

\section{Introduction}
Ever since the introduction of Transformers \citep{vaswani2017attention},
they have been widely used for autoregressive sequence generation, and have come to dominate many areas of machine learning
\citep{10.7554/eLife.82819, Erol2024AudioMBA, firstfew}. Recently, State Space Models (SSMs) have emerged as a promising alternative. By offering improved computational efficiency while maintaining competitive performance, SSMs are increasingly being adopted across a range of tasks \citep{gu2021efficiently,gu2023mamba,ren2025sambasimplehybridstate, dao2024transformers}.

Autoregressive models reuse the same parameters across repeated computation steps. When examples have different sequence lengths or require different numbers of rollout steps, the final prediction for each example may be a different function of the same shared parameters. Even if the per-step computation is linear, the resulting margin is generally a non-homogeneous polynomial in the parameters, with a degree that depends on the number of autoregressive applications.

This creates a setting that is not captured by the existing max-margin implicit-bias theory. Classical results show that gradient descent on separable linear models with an exponential-type loss converges in direction to a hard-margin classifier \citep{soudry2018implicit}, and later work extends this picture to homogeneous and some non-homogeneous parameterizations \citep{gunasekar2018implicit,lyu2019gradient, ji2020directional, kunin2022asymmetric, bartlett}. In those settings, however, the training examples are typically evaluated by a single shared predictor, or by predictors with a common homogeneity structure. In contrast, variable-step autoregressive models lead to multiple polynomial predictors, one per example or per length class, all sharing parameters but potentially growing at different asymptotic rates during training.

We study this phenomenon through the lens of shared-parameter, linear-in-the-input polynomial predictors, in a classification setting. Our main observation is that the relevant degree of a predictor is not necessarily its algebraic degree. Along a gradient-descent trajectory, leading homogeneous components may vanish in the limiting direction, causing a high-degree margin polynomial to grow like a lower-degree polynomial. We call this trajectory-dependent quantity the effective degree. Since exponential-loss gradients weight examples by their current margins, examples with larger effective degree can become asymptotically suppressed, while examples with smaller effective degree can determine the limiting classifier.

Our main result gives a conditional max-margin characterization for this setting. Assuming that the loss converges to zero, the parameter direction converges, and each margin has a well-defined integer effective degree along the trajectory, we prove that the limiting direction of gradient descent satisfies the KKT conditions of an effective-degree max-margin problem. The unit-margin constraints are imposed only by examples with minimum effective degree, and higher-effective-degree examples enter only through feasibility constraints. Thus the limiting classifier can be governed by the slowest-growing margins rather than by all examples equally.

We instantiate the theorem in scalar linear autoregressive models, linear state-space models (SSMs), and linear autoregressive SSMs. These models are not intended to capture all features of modern sequence architectures; rather, they isolate the effect of parameter reuse across variable numbers of steps.
Indeed, the implicit bias of linear models and linear neural networks has been extensively studied as a testbed for understanding the implicit bias in more complex architectures \citep{vardi2023implicit}, and we adopt this approach here by considering linear autoregressive models and linear SSMs.
In our simplified settings, when additional recurrent or autoregressive steps increase effective degree, our theorem predicts a form of length-sensitive implicit bias: shorter or lower-effective-degree examples can dominate the asymptotic margin solution.

\paragraph{Our contribution:}
\begin{itemize}[leftmargin=*]
\item We introduce a trajectory-dependent effective degree for non-homogeneous polynomial predictors and show that it can differ sharply from the algebraic degree.
\item Under standard asymptotic convergence assumptions and an effective-degree regularity assumption, we prove a max-margin/KKT characterization of the gradient-descent limit for exponential loss. This characterization is significantly different from the standard max-margin characterization of homogeneous models.
\item We show that, in the linear autoregressive and SSM instantiations, this characterization yields a length-sensitive bias whenever additional computation steps increase effective degree.
\end{itemize}
\section{Related Work}

\paragraph{Implicit bias in homogeneous and near-homogeneous models.}
There is a large body of work studying the implicit bias of gradient-based optimization.
\citet{soudry2018implicit} initiated this line of work by showing that gradient descent on separable data with the logistic loss converges in direction to the max-margin classifier.
Subsequent works extended this perspective to broader classes of models, including linear convolutional networks and homogeneous polynomials \citep{gunasekar2018implicit,gunasekar2018characterizing}.
For homogeneous models, \citet{lyu2019gradient} showed that any limit point of the optimization path is in the direction of a KKT point of the corresponding margin-constrained Euclidean-norm minimization problem, and \citet{ji2020directional} established directional convergence of parameters and gradients under additional structural assumptions. \citet{ji2020characterizingimplicitbiasprimaldual} used the induced dynamics on dual variables to characterize convergence rates to the implicit bias in linear models; we use a related observation to prove directional convergence of the gradient along a subsequence.

Several works go beyond standard homogeneity.
\citet{nacson2019lexicographicdepthsensitivemarginshomogeneous} studied regularization paths for sums of homogeneous models with different degrees, but their setting uses separate parameters for each homogeneous component and does not analyze gradient descent dynamics.
\citet{kunin2022asymmetric} generalized this framework to \emph{quasi-homogeneous} models, where different parameters have different homogeneity degrees, and showed that the resulting implicit bias can be asymmetric.
Our polynomial models, as well as their multidimensional extensions, do not fall under this quasi-homogeneous framework; see \Cref{subsection:kunin}.
\citet{lippl2022implicitbiasgradientdescent} studied gradient descent on parameter-sharing homogeneous polynomials, allowing the polynomial to vary across examples.

Most closely related to our treatment of non-homogeneous structure is \citet{bartlett}, who established implicit bias results for \emph{nearly-homogeneous} models.
Although polynomials satisfy their definition of near-homogeneity, their analysis considers a single predictor shared across all data points, whereas we study a more general setting in which each data point may induce a different predictor.
Moreover, their result relies on a strong assumption on the optimization dynamics, which we relax for polynomials; see \Cref{Section:comparison} for a detailed comparison.

\paragraph{Implicit bias in attention and sequence models.}
Recent works have begun to study implicit bias in sequence models.
For self-attention, \citet{tarzanagh2023transformers,ataee2023max,vasudeva2025implicitbiasfastconvergence} analyze the implicit bias and convergence behavior of attention-based architectures.
These works are complementary to ours: they focus on attention mechanisms, whereas our analysis concerns polynomial and state-space/autoregressive structures.

\paragraph{State-space models and linear autoregressive models.}
Several recent theoretical works study state-space models and related linear autoregressive models in settings close to ours.
\citet{merrill2025illusionstatestatespacemodels} analyze the expressivity of state-space models.
\citet{slutzky2025implicitbiasstructuredstate} study the effect of data poisoning on the implicit bias of diagonal SSMs with the square loss for sequences of fixed length.
Other works analyze SSMs with structured transition matrices or related architectures \citep{gu2021efficiently,dao2024transformers}, while \citet{smkal2024theorylearningdynamicsdeep} study the learning dynamics of SSMs under square loss.
For autoregressive models, \citet{joshi2025theorylearningautoregressivechain,joshi2026learningthinkmultiplethinkers} establish computational learning results for autoregressive linear threshold models in a setting similar to ours, and \citet{thrampoulidis2024implicitoptimizationbiasnexttoken} study implicit optimization bias in linear autoregressive next-token prediction with fixed-length sequences and cross-entropy loss.
Finally, \citet{emami2021implicitbiaslinearrnns} analyze linear RNNs at the NTK limit, showing that the effect of input tokens decays geometrically; the model they consider is reminiscent of the SSMs studied in \Cref{Section:SSM}.

\section{Preliminaries and Problem Definition}
\label{Section:Preliminaries}
\paragraph{Notations.}
We denote vectors $\bv\in \reals^d$ by boldface letters i.e. $\bv=(v_1,\ldots,v_d)$, and $\bar \bv:=\bv/\norm{\bv}$. We denote the sign function $\sign(x):=\begin{cases}
    1 & x\geq0 \\ -1  &\text{else}
\end{cases}$. We denote by $\norm{\cdot}$ the Euclidean norm, and $[n]:=\{1,\ldots, n\}$. For $\nu\in \mathbb{N}$, we denote by $\cp^\nu$ the sum of (multivariate, vector valued) monomials in $\cp$ of degree $\nu$, this is a $\nu$-homogeneous polynomial, which we call the $\nu$-homogenization of $\cp$.

\subsection{Problem Definition}
\paragraph{Model and Data.}
We consider a dataset $\{X_i,y_i\}_{i=1}^N$ where $X_i\in\reals^{  d\times l_i}$
and $y_i\in\{\pm 1\}$. We view each input sample $X_i$ as $l_i$ tokens, each of dimension $d$. Note that we allow for each sample to have a different number of tokens. The difficulty in determining the implicit bias of our models already appears in the case of $d=1$. Hence, for brevity, in the main part of the paper, we consider the case of $d=1$, namely, each token is a scalar in $\reals$, and we denote the input sequence as $\x_i \in \reals^{l_i}$. All our results are readily extended to the case where $d$ is some constant larger than $1$, and we prove these extensions formally in  \Cref{Section:MultidimToken}.

Let $\{\bx_i,y_i\}_{i=1}^N$ be a dataset where $\bx_i \in \reals^{l_i}$, $y_i\in\{\pm 1\}$.
We define a set of models $\{\cm_i:\reals^{D}\to\reals^{l_i}\}_{i=1}^N$, that share the same parameters $\bw\in \reals^D$, but might have different output dimensions. These models act linearly on the data $\bx_i$, namely their output is $\hat{y_i} := \inner{\cm_{i}(\bw),\x_i}$ where  $l_i$ is the dimension of the $i$-th sample,  i.e. $\bx_i\in \reals^{l_i}$, or in other words, the number of one dimensional tokens $\x_i$ has. Note that although $\{\cm_i:\reals^{D}\to\reals^{l_i}\}_{i=1}^N$ indexes $N$ models, they may not all be distinct, i.e., we allow $\cm_i = \cm_j$ for some $i\neq j$. This multi-model setting is a generalization of the standard implicit bias setting where all the samples share the same predictor $\cm$. This framing aligns with the setting of SSMs and autoregressive models, where each sample may consist of a different number of input tokens or autoregressive steps.
In this paper, we focus on two such models: (1) a linear autoregressive model and (2) a linear state-space model (SSM), formally defined in \Cref{Section:Linear,Section:SSM}.
Although the models acts linearly on the input data, their parameterizations can be highly non-linear, leading to implicit biases that differ substantially from those of standard linear predictors.

\paragraph{Training.}
In this paper we will consider the exponential loss and denote $\cl(\bw)=\sum_{i=1}^n\exp(-y_i \cm_{i}(\bw)^\top\x_i)$. Our results can be readily extended to the logistic loss as well. Throughout the paper, we consider gradient descent w.r.t. vectors or matrices of the parameters. We denote by $\bw^{(0)}$ the parameters at initialization, and
\begin{equation} \label{eq:gradient_updates}
    \bw^{(t)}=\bw^{(t-1)}-\eta_t\nabla\cl(\bw^{(t-1)})~,
\end{equation}
as the parameter vector after the $t^{\text{th}}$ step of gradient descent, where $\eta_t$ denotes the learning rate at time $t$. Our results can be applied to general learning rate schedulers, and we only assume the learning rate sequence $\{\eta_t\}_{t=1}^\infty$ does not increase.
We denote by $\bar \bw^\infty=\lim_{t\to\infty}\frac{\bw^{(t)}}{\norm{\bw^{(t)}}}$ the directional limit of $\bw$ when it exists.

\subsection{Linear Autoregressive Model}
\label{Section:Linear}
The first model we consider is a linear autoregressive model. This type of models is common in many  academic fields for the prediction of time series \cite{econometrics, psych, kirchgassner2012introduction, Dama2021PartiallyHMA, Chen2025InterpretableTSA}. Moreover, it is a basic simplification of large language models, where we isolate the autoregressive part of transformers from the non-linear parts which are the self-attention and MLP. The core mechanism is that the model generates new tokens, and then apply itself autoregressively on its own output.

We consider an example $\x\in \reals^l$ which we interpret as an input prompt, i.e., the initial part of some sequence. Here we will consider parameters $\bw \in \reals^D$ and assume for simplicity $D=l$, thus $\x^\top =\x^\top_{1:D} = (x_1,\ldots, x_D)$. We compute a sequence of $T$ autoregressive steps, where the final token's sign is the label. Formally, the model computes:
\begin{equation}
    x_{D+1}=\inner{\bw,\x},\quad x_{D+2}=\inner{\bw,\x_{2:D+1}},\quad\dots\quad, x_{D+T}=\inner{\bw,\x_{T:D+T-1}}~.
\end{equation}
The label of the sequence is defined as $\sign(x_{D+T})$. More generally, we may allow $l \neq D$: in the case $l<D$, we pad $\x=(0,\ldots, x_1,\ldots, x_l)$, and in the case $l>D$, apply $\bw$ only to the last $D$ elements of $\x$. This can be thought of as the input prompt being longer than the model's context length.

\begin{wrapfigure}[13]{R}{0.5\textwidth}
\centering
\includegraphics[width=\linewidth]{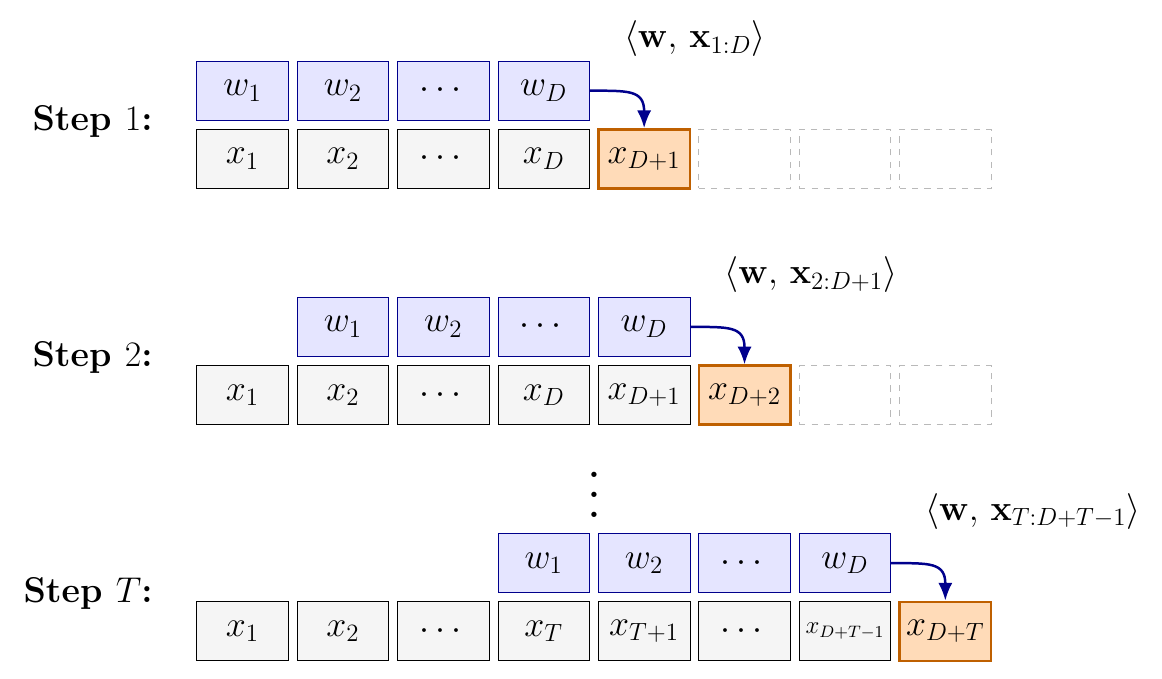}
\caption{Movie-strip view of the autoregressive construction.}
\label{fig:ar-movie-strip}
\end{wrapfigure}

We denote the vector of coefficients of $x_1, \ldots, x_D$ in $x_{D+T}$ as $\cp$, a polynomial in $\bw$. Using this notation, we have that $\inner{\cp(\bw),\x} = x_{D+T}$. See \Cref{fig:ar-movie-strip} for a demonstration.

We can interpret the setup in the context of language models as having $l$ initial one-dimensional tokens, $x_1,\ldots,x_l$, which serve as the starting sequence. In standard transformers, and autoregressive language models in general, text is generated token by token: the model predicts the next token based on the current sequence, appends this token, and then feeds the entire updated sequence back into the model to produce the next one, $T$ times. Our framework mirrors this process but uses a linear autoregressive model, which is more tractable to analyze.
We note that in our work the polynomials are usually associated with an indexed example $\x_i\in \reals^{l_i}$, hence we denote them $\cp_i$.
Note that crucially for a dataset $\{\x_i, y_i\}_{n=1}^N$, we have a number of autoregressive steps $T_i$ for each example, which together with $l_i$  changes $\cp_i$. That is, both elements embodied in the polynomial predictor. This also means that we may have $\cp_i=\cp_j$ for $i\neq j$, if both examples $i,j$ are labeled using the same number $T_i$ of autoregressive steps and share the same input length $l_i$.
We note that in this model our training setup corresponds to training directly on the output after $T_i$ steps, rather than training on the next-token-prediction.

\subsection{Linear State Space Models}
\label{Section:SSM}

The second model we consider, is a Linear State Space Model (SSM). SSMs are used to model natural phenomena and have been lately adopted by the Machine Learning community as a promising candidate for Large Language Models to replace and complement Transformers \citep{gu2021efficiently, gu2023mamba, dao2024transformers}.

We consider a sequence of tokens of variable lengths $X\in \reals^{d\times l}$ where $l$ is the length of the sequence. SSMs consist of parameter matrices $A\in \reals^{d_A\times d_A}$, $B\in \reals^{d_A\times d}$, $C\in \reals^{d_A \times d_C}$ where the model is given recursively by a sequence of ``hidden'' states $\bh$, i.e., for $1\leq i\leq l$, $\bh_i=A\cdot\bh_{i-1}+B(X)_i$, where $(X)_{i}$ denotes the $i^{\text{th}}$-column of $X$, with $\bh_0=\mathbf{0}$, and with the output being $y=\sign(C^\top\bh_{l})$.
Note that we consider binary classification and assume $d_C=1$. For the rest of the main part of the paper, we consider the case $d=1$, and let $B,C$ be vectors. We provide the generalization for a matrix $C$ and full proof for the more general case of $d \geq 1$ in \Cref{Section:MultidimToken}.

A useful convolutional way to denote it overall is
\[
y=\sign\left(C^\top \sum_{i=1}^{l}A^{l-i} B (X)_{i}\right)~.
\]
Note that SSMs in practice usually use nonlinearities, which we omit, and sometimes use special matrix structure for $A$, such as diagonal, or diagonal up to some low rank. We analyze the general case and note that in principle our analysis should apply to those cases as well.

With a dataset $\{\x_i,y_i\}_{n=1}^N$, denote the number of tokens of the $k^{\text{th}}$ sample as $l_k$, and consider the polynomial
\[
\cq_k(A,B,C)=\begin{pmatrix}
    C^\top A^{l_k-1} B \\
    \vdots \\
    C^\top A B\\
    C^\top B
\end{pmatrix}
\]
i.e. $y_k=\inner{\cq_k(A, B, C), \x_k}$, note that $\cq_k$ depends only on the number of steps taken by the SSM, that is the model works for a variable length input $\x_k\in \reals^{l_k}$. Similarly to the linear autoregressive case, although each sample $k$ is associated with a different polynomial $\cq_k$, we may have $\cq_k=\cq_j$ for $k\neq j$ if both examples $k,j$ have the same length, and $l_k$ is embodied in $\cq_k$.

\subsection{Linear Autoregressive SSM} \label{Section:AutoregressiveSSM}
Note that the SSM from the previous section does not use its own outputs as inputs for the next steps. That is, the SSM simply runs over the input sequence but does not extend it. Hence it is not autoregressive in this sense.
We now introduce the \emph{Autoregressive SSM}.
Consider an example $\x \in \reals^{l}$, that we can think of as an input prompt, or as a beginning of some sequence $\x^\top = \x^\top_{1:l} = (x_1,\ldots, x_l)$, we compute a sequence of $T$ autoregressive steps:
\begin{equation}
    x_{l+1}=C^\top \sum_{i=1}^{l}A^{l-i} B x_{i},\quad x_{l+2}=C^\top \sum_{i=1}^{l+1}A^{l+1-i} B x_{i},\quad\dots\quad, x_{l+T}=C^\top \sum_{i=1}^{l+T-1}A^{l+T-1-i} B x_{i}~.
\end{equation}

The label of the sequence $\x$ is defined as $\sign(x_{l+T})$, and we write the polynomial coefficients as a vector $\cq(A,B,C)$, i.e. $y=\sign\left(\inner{\cq(A, B, C), \x}\right)$. Note that $\cq$ depends both on the number of autoregressive steps $T$ \emph{and} the number of input tokens $l$.
As in previous sections, $\cq_i$ implicitly embeds $l_i$ and $T_i$, and we allow $\cq_i = \cq_j$ for $i\neq j$.

\section{Assumptions on the Dynamics}

In this section, we present the assumptions about the optimization dynamics that we will use to prove our main results.
Before introducing our assumptions, we define the ``effective degree'' of a polynomial over an optimization trajectory:

\begin{definition}[Effective Degree]
Let $p:\reals^D\to\reals$ be a polynomial with $\deg(p)=k$. Let $\bw^{(t)}\in \reals^D$ be some trajectory of the polynomial's inputs, such that $\norm{\bw^{(t)}}\to \infty$ and $\lim_t \frac{\bw^{(t)}}{\norm{\bw^{(t)}}}=\bar \bw^{\infty}$. We say $p$ has an \emph{effective degree} of $m$ over the trajectory $\bw^{(t)}$ (denoted $\effdeg_{\bw^{(t)}}(p)$, or simply $\effdeg(p)$ when $\bw^{(t)}$ is implicit) if
\[
\lim_{t\to\infty} \frac{p(\bw^{(t)})}{\norm{\bw^{(t)}}^j}=
\begin{cases}
Cp^j(\bar\bw^{\infty})\neq0 & j=m\\
0 & j>m
\end{cases}~,
\]
and similarly
\[
\lim_{t\to \infty} \frac{\nabla p(\bw^{(t)})}{\norm{\bw^{(t)}}^{j-1}}=
\begin{cases}
C\nabla p^{j}(\bar\bw^{\infty})\neq \mathbf{0} & j=m\\
\mathbf{0} & j>m
\end{cases}~.
\]
For some constant $C>0$.
\end{definition}
Intuitively, the definition states that along the trajectory, $\bw^{(t)}$ moves in a direction that cancels out higher degree monomials (if such monomials exist). A simple example would be the trajectory $\bw^{(t)}=(t,t,t)$ with the polynomial $p(\bw)=w_1+(w_2-w_3)^2$, then $p(\bw^{(t)})=t=w_i$ for all $i$. From now on, for a polynomial $p$, we refer to $\deg(p)$ as the algebraic degree of $p$, as to not confuse it with its effective degree.

We use the following assumptions throughout the paper.

\begin{assumption}[Loss minimization] \label{assmp:minimization}
The iterates $\bw^{(t)}$ minimize the objective, i.e. $\loss(\bw^{(t)})\underset{t\to\infty}{\rightarrow} 0$.
\end{assumption}
\begin{assumption}[Convergence in direction] \label{assmp:param_direction}
    The model's parameters converge in direction, i.e. $\lim_{t\to\infty}\frac{\bw^{(t)}}{\norm{\bw^{(t)}}}=\bar\bw^{\infty}$.
\end{assumption}

\begin{assumption} \label{assmp:effdeg}
$\forall i$, $\effdeg_{\bw^{(t)}}(\cp_i^\top \x_iy_i)=\nu_i$ for $\nu_i\in \mathbb{N}$\footnote{We note that a polynomial with $\effdeg_{\bw^{(t)}}(\cp^\top\x y)=0$ contradicts \Cref{assmp:minimization}, hence for the polynomials we will consider, $\effdeg\geq 1$.
}, with some $C_i> 0$.
\end{assumption}
In our first assumption, we assume the iterates of the parameters $\bw^{(t)}$ minimize the objective, i.e. we assume that the loss converges to zero. This is a practical assumption as in many cases deep learning exhibits complete interpolation i.e. zero or close to zero error on the data. It also allows us to separate the question of whether gradient descent interpolates from the question of where gradient descent converges when it interpolates. As such, this assumption is also common in literature (e.g., \cite{gunasekar2018implicit, gunasekar2018characterizing, vardi2021implicitregularizationrelunetworks, timor2022implicitregularizationrankminimization}).

Our second assumption states that the parameter vector $\bw^{(t)}$
converges in direction. This assumption means that the trajectory of $\bw^{(t)}$ is ``well behaved'', and does not oscillate indefinitely. This is a standard assumption in the literature, and although some works, such as \citet{ji2020directional,bartlett} have managed to prove it under some assumptions, many works use it implicitly or explicitly as an assumption (e.g., \citet{kunin2022asymmetric,gunasekar2018implicit,lyu2019gradient}).
Studying when directional convergence can be proved in our setting is a natural and important open question; we leave it to future work and focus here on characterizing the limit.

Our third assumption is more elaborate, but it is natural when we think of polynomials. If we fix some unit vector $\bar\bw^{\infty}$, then for any polynomial $p(\bw)$, the univariate polynomial $q\left(\norm{\bw^{(t)}}\right):=p\left(\norm{\bw^{(t)}}\bar \bw^{\infty}\right)=\Theta\left(p^k\left(\norm{\bw^{(t)}}\bar \bw^{\infty}\right)\right)$ when $\norm{\bw^{(t)}}\to \infty$, for the maximal $k$ for which  $p^{k}\left(\bar\bw^{\infty}\right)\neq 0$. In words, when we plug in a fixed direction to a multivariate polynomial, it starts behaving like a univariate polynomial, hence grows with its degree. In this case, its degree would be the highest degree that does not cancel when plugging the direction in. The gradient part of the definition ensures that the optimization does not go too far off this rather intuitive behavior.

\subsection{Initialization-Dependent Effective Degree}
Having the effective degree dependent on the gradient descent trajectory may seem like a burdensome complication, which makes the calculation of the implicit bias highly contextual. We show that
under the setting of our non-homogeneous polynomials, the effective degree may be different for initializations in two different open sets. Hence, instead of only depending on the model and training examples, it depends on the initialization as well.

\begin{lemma}\label{lem:different_init}
There exist a polynomial predictor $\cp:\reals^D\to\reals^n$, a dataset $\{\x_i, y_i\}_{i=1}^n$ and two disjoint open sets $U_2, U_4 \subset \reals^D$, such that when trained with gradient descent (\Cref{eq:gradient_updates}) with a constant step size $\eta>0$, the margins $m_i=y_i\inner{\cp, \x_i}$ have:
\[
\effdeg(m_i)= \begin{cases}
    2i & \bw^{(0)} \in U_2 \\
4i & \bw^{(0)} \in U_4
\end{cases}
\]
\end{lemma}
The full proof is in \Cref{subsec:init_dependence}. The proof holds for every $n$, and for any subset of the above dataset, and it can be extended to a larger family of predictors beyond that in the lemma. The example above requires careful analysis of the optimization trajectory, but this is not always the case. We provide a concrete, nontrivial family of predictors and datasets, for which the effective degree follows from a short calculation, in \Cref{subsec:simple_calc}.
\subsection{Comparison with \citet{bartlett}} \label{Section:comparison}
At this point, one might think that the implicit bias in our setting can be analyzed using the results from \citet{bartlett}, at least for the case where all examples share the same predictor. \citet{bartlett} established a result on the implicit bias for the class of \emph{nearly-homogeneous models}, which include our polynomial setting.
Perhaps surprisingly, their result is restrictive, as the assumptions of \citet{bartlett} implicitly imply that the effective degree of the model equals its algebraic degree. Thus, we can think of \Cref{assmp:effdeg} as a relaxation which captures a broader variety of dynamical behaviors.

\begin{assumption} [\citet{bartlett}, simplified for our setting] \label{assmp:bartlett}
At some point $t_0$ during optimization, $\cl(\bw^{(t_0)}) \leq \exp\left(-p\left(\norm{\bw^{(t_0)}}\right)\right)$ for some $p(x)$. If the predictor is a polynomial of the form $f(\bw;\x)=\inner{q(\bw),\x}$, then $0\leq\deg(p)\leq\deg(q)-1$, where for the special case of an homogeneous $q$, $p\equiv0$
\end{assumption}

\Cref{assmp:bartlett} is presented in \citet{bartlett} as a strong separation assumption on the data.
However, it is essentially an assumption on the optimization rate. Since it includes $t_0$ on both sides of the inequality, it requires that at some point, the optimization reaches a certain rate; specifically, it requires that at some point all margins behave as if they were some polynomial w.r.t. the model's parameters. The problem is that the condition implies that all margins progress at some maximal rate. Is it possible that this maximal rate is never reached, and hence \Cref{assmp:bartlett} does not hold? As we shall see next, the answer is positive even for a very simple setting with our linear autoregressive model.

\begin{lemma} \label{lemma:bartlett_fails}
Consider the linear autoregressive setting from \Cref{Section:Linear}, trained using gradient descent (\Cref{eq:gradient_updates}) with a constant step size $\eta>0$ and initialization at $\mathbf{0}$. Then, there exists a size-$1$ dataset $\{(\x,y)\}$, such that \Cref{assmp:bartlett} from \citet{bartlett} does not hold, while our \Cref{assmp:minimization,assmp:param_direction,assmp:effdeg} hold.
\end{lemma}
The example we provide has a margin (polynomial w.r.t. $\bw$), with algebraic degree $4$, but an effective degree of $1$. This is in contradiction to \Cref{assmp:bartlett} that implies the effective degree is higher.
Moreover, if we do not insist on the setting from \Cref{Section:Linear}, we can find such examples for every even algebraic degree $d$, and $k<d$ effective degree (see \Cref{Section:remark}), which in the case of even $k$ work for almost all initializations.
Therefore, a more nuanced approach is necessary to capture the range of possible dynamics of optimization, \Cref{assmp:effdeg} is one such approach.
In fact, in the following lemma we prove that in the case of polynomials,
\Cref{assmp:bartlett} implies \Cref{assmp:minimization,assmp:param_direction,assmp:effdeg} (see \Cref{Section:implication_proof} for the proof):
\begin{lemma}\label{lemma:assump_implication}
    In the setting where the predictor is polynomial, \Cref{assmp:bartlett} implies \Cref{assmp:minimization,assmp:param_direction,assmp:effdeg}.
\end{lemma}

\Cref{assmp:bartlett} essentially requires that the effective degree for every example is equal to the algebraic degree, while \Cref{assmp:effdeg} allows \emph{any} effective degree. As we will see in \Cref{Section:Results}, the implicit bias of gradient descent is different than the one presented in \citet{bartlett} for the cases where \Cref{assmp:bartlett} does not hold. In \Cref{Section:Experiments}, we empirically verify that our assumptions (including \Cref{assmp:effdeg}) indeed hold, at least in some cases.

We further emphasize that while the above discussion shows that for the case where all examples share the same predictor $\cp_i$ (i.e., the same $l_i$ and $T_i$), our \Cref{assmp:effdeg} is weaker than \Cref{assmp:bartlett} of \citet{bartlett}, cases where $l_i$ or $T_i$ (and hence $\cp_i$) are not uniform for all examples are not covered by the setting in \citet{bartlett} at all. On the other hand, \citet{bartlett} cover non-polynomial model, which we do not consider.

Finally, we note that it is theoretically possible that the dynamics do not satisfy our assumptions. This can happen when the effective degree of the polynomial along the trajectory behaves as a ``fractional-degree''. We elaborate on these cases in \Cref{Section:Fractional}. Studying such dynamics introduces a challenging analysis since the margin in these cases, as well as the KKT conditions, is not well-defined. We leave investigating those cases for future work.

\section{Main Results} \label{Section:Results}
We will now present our main result, and show how it applies to the various models we consider.
\begin{theorem} \label{thm:main}
Let $\{\cp_i\}_{i=1}^N$ a collection of polynomials with parameters $\bw\in \reals^D$. For a dataset $\{\x_i, y_i\}_{i=1}^N$ where $y_i=\sign \left(\inner{\cp_{i}(\bw) ,\x_i}\right)$, consider the sequence of gradient updates (\Cref{eq:gradient_updates}) with a non-increasing sequence of step sizes $\{\eta_t\}_t$, minimizing the empirical risk.
If \Cref{assmp:minimization,assmp:param_direction,assmp:effdeg} hold, we have that $\bar \bw^{\infty}:=\lim_{t\to \infty}\frac{\bw^{(t)}}{\norm{\bw^{(t)}}}$ is a positive scaling of a first order stationary point (KKT point) of the following optimization problem:
    \begin{align*}
    {\min}_{\bw\in \reals^D}\norm{\bw}_2^2 \text{ s.t. } &\forall i, y_i \inner{\x_i,\cp^{\nu_{i}}_{i}(\bw)} \geq 0 && \tag{\text{OPT}} \label{eq:OPT}\\
    &\forall i\in Z, y_i C_i\inner{\x_i,\cp^{\nu_{i}}_{i}(\bw)}\geq 1~,
    \end{align*}
     where $Z=\arg \min_{i\in[n]}\left\{\effdeg_{\bw}y_i\inner{\cp_i, \x_i}\right\}$.
\end{theorem}
The idea behind the proof is to establish the dominance of the lowest effective-degree margins on the gradient, as higher effective degrees decay faster. Then, analyzing the induced gradient-descent updates on the dual weights, i.e., the dual certificate, to show a subsequence of these weights converges. Finally, we use a convexity argument to show the gradient under this subsequence converges in direction to $\bar\bw^\infty$. An extended proof intuition and the full proof are in \Cref{proof:main}.

Note that this result is different in numerous ways from the
implicit bias of homogeneous models, given by
\begin{equation*}
    \min_{\bw}\norm{\bw} \text{ s.t. } \forall i,\; y_i f(\bw) \geq 1~.
\end{equation*}
\begin{enumerate}[leftmargin=*]
    \item The positive-margin constraint is only w.r.t. some of the samples, specifically the ones with lowest effective degree. The constraint on the rest of the samples only roughly requires that they will be classified correctly.
    \item Each sample is associated with a different predictor.
    \item  Importantly, the margins $y_iC_i \inner{\x_i,\cp^{\nu_{i}}_{i}(\bw)}$ use different predictors than those being optimized ($\cp_i$ rather than $\cp_i^{\nu_i}$), that is, the direction of convergence is towards a KKT point of an optimization problem which may a-priori seem unrelated to the optimized objective. This is similar to \citet{bartlett}; however, they use the algebraic-degree homogenization of the optimized predictors instead of the effective-degree homogenization.
    \item The implicit bias is different depending on the dynamics of gradient descent and depends on the effective degree, which is optimization dependent. This is similar in spirit to \citet{ataee2023max, tarzanagh2023transformers}, where multiple implicit biases are possible depending on the optimization dynamics.
\end{enumerate}

Note that \Cref{thm:main} is phrased in a rather general sense for polynomial predictors, but since we know each of our models is in fact a polynomial, we can see how it applies to them.
\begin{corollary}\label{coro:linear}
Under the settings of \Cref{thm:main},  $\bar \bw^{\infty}$ is a positive scaling of a first order stationary point of \Cref{eq:OPT}, where:
\begin{itemize}[leftmargin=*]
    \item For linear autoregressive models, the polynomial predictors are defined as in \Cref{Section:Linear}.
    \item For linear SSMs, the polynomial predictors are defined as in \Cref{Section:SSM}.
    \item For linear autoregressive SSMs, the polynomial predictors are defined as in  \Cref{Section:AutoregressiveSSM}.
\end{itemize}
\end{corollary}

\paragraph{Interpretation of the max-margin solution.}
We split the interpretation into two cases: (1) All polynomials have their maximal possible effective degree, i.e. $\forall i, \deg(\cp_i)=\effdeg(\cp_i)$. (2) Some polynomials do not achieve their maximal effective degree, i.e. $\exists i,\deg(\cp_i)>\effdeg(\cp_i)$.

In the first case, our results imply that the implicit bias for a max-margin solution is affected mainly by the shortest sequences seen during training (i.e. polynomials with lowest degree). The other sequences are only required to obtain a correct classification. In the second case, the solution is mainly affected by the examples with the lowest effective degree. This very roughly extends the interpretation of the first case, in the sense that fewer autoregressive steps, or lower-degree polynomials, are most likely to have a lower effective degree. For example, in the case of polynomials corresponding to only one autoregressive step, these polynomials are \emph{guaranteed} to have the minimal effective degree, and hence have greater effect on the max-margin.

\section{Experiments}
\label{Section:Experiments}
The discussion in \Cref{Section:comparison} alludes that our assumptions may not hold in general, hence it's customary to check them empirically. We performed experiments with autoregressive linear models and SSMs both on synthetic data and on MNIST \citep{lecun1998mnist}, and found many cases for which our assumptions hold.
In the main text, we present our experiments on autoregressive linear models with a synthetic teacher-student dataset. Here, we generate a random teacher and use it to label Gaussian examples, with different numbers of autoregressive steps $T \in \{1,3,5\}$. We refer the reader to \Cref{Section:Details_Experiments} for experiments with SSMs and linear autoregressive models in other settings, and for more information about the experiments.

\subsection{Verifying Our Assumptions}

\begin{figure}[t]
  \includegraphics[width=1\textwidth]{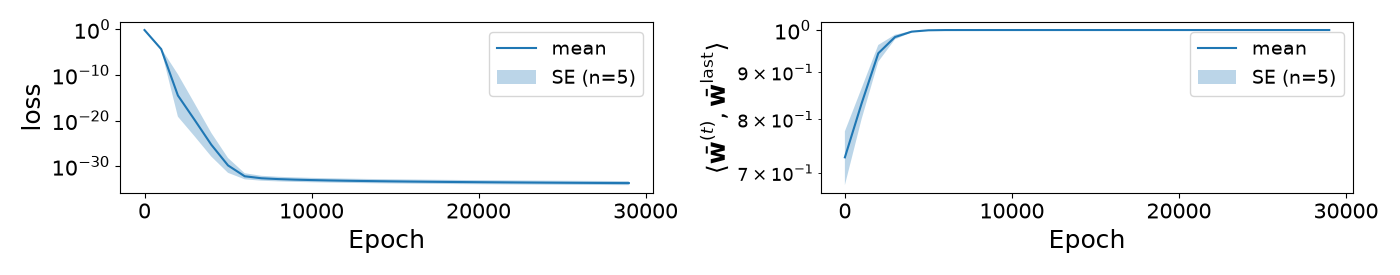}

  \caption{\textbf{Left:} The loss. \textbf{Right:} Cosine similarity to the final predictor direction.
  Y-axes are log-scale.}
  \label{fig:loss_corr}
\end{figure}

\begin{figure}[t]
  \includegraphics[width=1\textwidth]{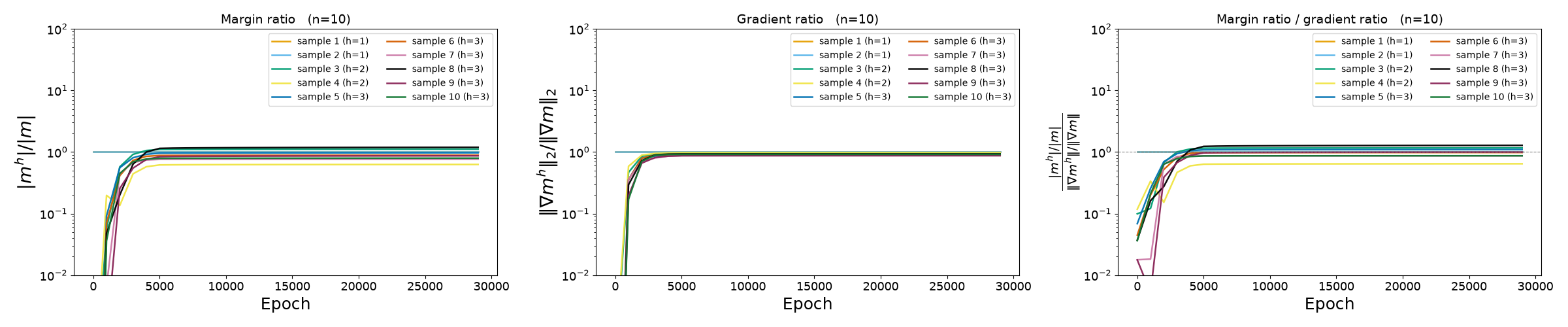}
  \caption{\textbf{Left:} Ratio between the effective degree homogenization of the margin and the margin, per example. \textbf{Middle:} The corresponding ratio for the norms of the gradients of the homogenization of the margin, and the margin. \textbf{Right:} The ratio of the two former ratios.
  Y-axes are log-scale.}
  \label{fig:assmp}
\end{figure}

In \Cref{fig:assmp,fig:loss_corr} we verify our assumptions.
First, we plot the loss to verify \Cref{assmp:minimization}. We see that the loss approaches $0$. Next, to verify \Cref{assmp:param_direction} we consider the correlation of the parameter direction with the final predictor, i.e. $\inner{\bar\bw^{(t)}, \bar\bw^\infty}$. Here, since we do not have access to the directional limit $\bar\bw^\infty$, we use the last iterate instead. We see that iterates become close to this direction quite early in training. For \Cref{assmp:effdeg}, we plot the following ratios for each training example. We plot $\abs{\frac{m_i^{\effdeg(m_i)}\left(\bw^{(t)}\right)}{m_i\left(\bw^{(t)}\right)}}$, likewise $\frac{\norm{\nabla m_i^{\effdeg(m_i)}\left(\bw^{(t)}\right)}}{\norm{\nabla m_i\left(\bw^{(t)}\right)}}$, and $\frac{\abs{m_i^{\effdeg(m_i)}\left(\bw^{(t)}\right)/m_i\left(\bw^{(t)}\right)}}{\norm{\nabla m_i^{\effdeg(m_i)}\left(\bw^{(t)}\right)}/\norm{\nabla m_i\left(\bw^{(t)}\right)}}$
where $m_i(\bw^{(t)}):=y_i\inner{\cp_i(\bw^{(t)}),\x_i}$. \Cref{assmp:effdeg} requires the first and second ratios to converge to the same constants, with the third ratio converging to 1.
For details about finding the effective degree, see \Cref{Section:Details_Experiments}.
\subsection{Convergence to Max-Margin}
\begin{figure}[t]
  \includegraphics[width=1\textwidth]{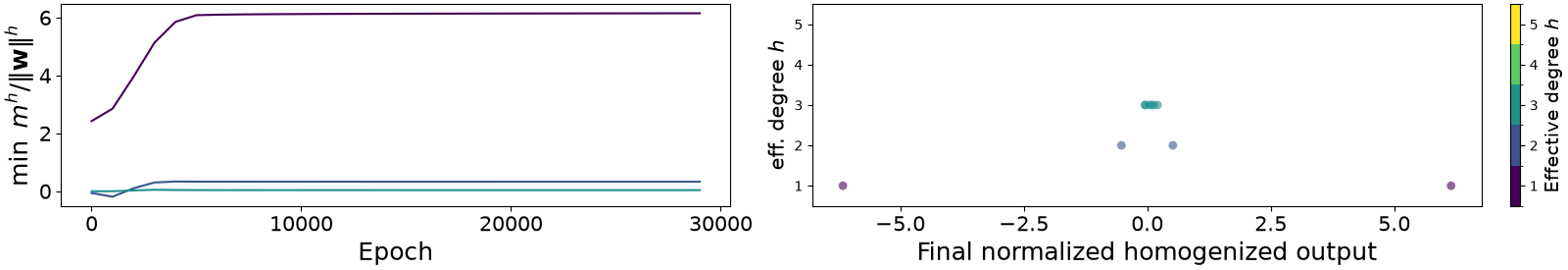}
  \caption{\textbf{Left:} Minimal normalized, homogenized margin per effective degree. \textbf{Right:} All points with their effective degrees and final normalized homogenized margins.}
  \label{fig:min_margin_scatter}
\end{figure}

In \Cref{fig:min_margin_scatter} we plot the minimum margins per effective degree. We see that margins of examples with the minimal effective degree indeed converge to higher values, in accordance to the constraints of \Cref{eq:OPT} and the results of \Cref{thm:main}. We also plot the outputs per effective degree after training. This verifies that examples with lower effective degree have margins bounded away from zero, and shows the concentration of margins in higher-effective degrees, which matches \Cref{eq:OPT}.


\section{Discussion and future work}
\label{Section:Discussion}

In this work, we characterized the implicit bias of gradient descent for polynomial predictors where each sample is labeled by a different predictor. It implies implicit bias results for Linear Autoregressive Models and linear SSMs, including linear Autoregressive SSMs. Finally, we shed light on the implicit bias of non-homogeneous models through the lens of polynomials, demonstrating that previous assumptions about their dynamics (\citet{bartlett}) may be overly strong in certain settings.

Some interesting questions remain for future work: First, can we completely characterize the implicit bias for the multi-polynomial case? We found that sometimes, the effective degree may be fractional (\Cref{Section:Fractional}). What is the implicit bias in those cases?
Second, how can we generalize these polynomial results to general non-homogeneous models? We consider only polynomials in this paper, but similar results may apply to other classes of predictors and be of much interest.

\subsubsection*{Acknowledgements}
GK is funded by the European Research Council (ERC) under the European Union’s Horizon 2022 research and innovation program (grant agreement No. 101041711), and the Simons Foundation (as part of the Collaboration on the Mathematical and Scientific Foundations of Deep Learning). GV is supported by the Israel Science Foundation (grant No. 2574/25), by a research grant from Mortimer Zuckerman (the Zuckerman STEM Leadership Program), and by research grants from the Center for New Scientists at the Weizmann Institute of Science, and the Shimon and Golde Picker -- Weizmann Annual Grant.
\clearpage

\bibliographystyle{authordate1}
\bibliography{bib}{}

\appendix
\section{Extra Discussions and Comparisons}
\subsection{Concrete Examples for Models} \label{subsection:concrete_ex}
\subsubsection{Linear Autoregressive}
Let $\x \in \reals^l$, $\bw\in\reals^D$. Suppose $D=l=2$ and consider $T=2$ autoregressive steps. We have that:
\begin{align*}
x_3 &= \inner{\bw, \x} = w_1x_1 + w_2x_2 &
x_4 &= \inner{\bw, \x_{2:D+T-1}} =\inner{\bw, \x_{2:3}}\\
&&&= w_1w_2x_1+(w_1+w^2_2)x_2
\end{align*}
 For our example the polynomial vector is $\cp(\bw)=\begin{pmatrix}
    w_1w_2 \\ w_1+w_2^2
\end{pmatrix}$,  and in our example $x_4=(w_1w_2, w_1+w_2^2)~ \x$.
\subsubsection{Autoregressive SSM}
For a concrete example, consider $l=2,T=2$. Then we have
\begin{align*}
    x_3 &= C^\top A B x_{1}+C^\top B x_{2} &
    x_4 &= C^\top A^2 B x_{1}+C^\top A B x_{2}+C^\top B x_{3} \\
    &&&=C^\top A^2 B x_{1}+C^\top A B x_{2}+C^\top B \left(C^\top A B x_{1}+C^\top B x_{2}\right)\\
    &&&=\left(C^\top A^2 B + C^\top BC^\top A B \right)x_{1}+\left(C^\top A B + \left(C^\top B\right)^2 \right)x_{2}
\end{align*}
That is the corresponding polynomial would be
\[
\cq(A,B,C)=\begin{pmatrix}
C^\top A^2 B + C^\top BC^\top A B
\\
    C^\top A B + \left(C^\top B\right)^2
\end{pmatrix}
\]
\subsection{Comparison to \citet{kunin2022asymmetric}} \label{subsection:kunin}
\citet{kunin2022asymmetric} define a model to be $\Lambda$-quasi-homogeneous if:
\[
f(\x, e^{\alpha\Lambda}\bw)=e^{\alpha}f(\x, \bw)
\]
For any scalar $\alpha>0$, input $\x$ and parameters $\bw$. These models satisfy a type of Euler-Homogeneous-Theorem. i.e.
\[
\inner{\nabla_\bw f(\x, \bw), \Lambda \bw}=f(\x, \bw)
\]
Under the setting of \Cref{Section:Linear}, let us consider the model with two autoregressive steps, with an example in 2D, i.e. $f(\x,\bw)=x_1w_1w_2+x_2(w_1+w_2^2)$. Since $f$ is linear in $\x$, and the identity must hold for any $\x$, then the identities
\begin{align*}
&\inner{\nabla_\bw w_1w_2, \Lambda \bw}=w_1w_2\\
&\inner{\nabla_\bw w_1+w_2^2, \Lambda \bw}=w_1+w_2^2
\end{align*}
must hold. Let $\Lambda=\begin{pmatrix}
    a & b\\ c & d
\end{pmatrix}$. We will start with the first identity
\[
\inner{\nabla_\bw w_1w_2, \Lambda \bw}=cw_1^2+(a+d)w_1w_2+bw_2^2~.
\]
Hence $b=c=0, a+d=1$. For the second identity, we have
\[
\inner{\nabla_\bw w_1+w_2^2, \Lambda \bw}=aw_1 +bw_2 + 2cw_1w_2+2dw_2^2~.
\]
Hence $a=1, d=1/2, b=c=0$. But now we have $a+d=3/2\neq 1$, contradiction to the first identity's requirements. Hence there exists no $\Lambda$ s.t. $f$ is quasi-homogeneous.
\subsection{Fractional Effective Degree}\label{Section:Fractional}
It is important to note that while \Cref{assmp:minimization,assmp:param_direction,assmp:effdeg} capture a wider variety of dynamical behaviors than \citet{bartlett}, they still do not capture the full array of possible dynamics. For example, our effective degree assumption does not always hold.

\begin{lemma} \label{lemma:all_fail}
There exists an example $(\x,y)$, under the setting in \Cref{Section:Linear}, using \Cref{eq:gradient_updates} after initializing at $\bw^{(0)}=(-2,1)$, with a constant step size $\eta>0$ s.t. its margin $m(\bw^{(t)})=\Theta\left(\norm{\bw^{(t)}}^{3/2}\right)$, violating both \Cref{assmp:bartlett} and \Cref{assmp:effdeg}.
\end{lemma}
We take $\x=\begin{pmatrix}
    0\\ -1/3
\end{pmatrix}, y=1$, and
\[
\cp(\bw)=\begin{pmatrix}
    w_1^2+w_1w_2^2 \\
    2w_1w_2 + w_2^3
\end{pmatrix}
\]
corresponding to 3 autoregressive steps. We have $\cl(\bw)=\exp(\frac{2}{3}w_1w_2+\frac{1}{3}w_2^3)$. The idea is that the optimization path follows a coordinate-relation of $w_1 \sim -w_2^2$, so we have $\bar\bw^{\infty}=\begin{pmatrix}
    -1\\0
\end{pmatrix}$ and $\norm{\bw}\sim w_2^2$, both since $w_1$ dominates $w_2$. Similarly, substituting $w_1$ with $-w_2^2$ we get that the margin $\sim w_2^{3}$. Hence effectively behaves like a $3/2$ degree polynomial - which is not an integer. A way to think about what happens is that $\bw^{(t)}/\norm{\bw^{(t)}}$ converges slow enough, that the radial speed in the direction of $\bar\bw^{\infty}$ is significantly slower than the radial speed of $\bw$. The full proof is in \Cref{Section:assumptions_fail}. Note that in this example \Cref{assmp:bartlett} does not hold as well, as by \Cref{lemma:bartlett} it requires the margin to scale like $\Theta\left(\norm{\bw^{(t)}}^2\right)$.

A natural question arises as to whether  \Cref{assmp:effdeg} can be generalized to fractional-effective degrees. This is an interesting future research direction. We note that it cannot simply be extended using our proof technique. First, note that there is no trivial generalization for homogenization to a fractional degree of a polynomial with integer powers, and \Cref{assmp:effdeg} requires the limit to equal a homogenization of the margin. Second, if we relax \Cref{assmp:bartlett} to only require $\lim_{t\to\infty}\frac{m(\bw^{(t)})}{\norm{\bw^{(t)}}^k}=C\neq 0$ for some fractional $k$, a new problem arises. Recall that by \Cref{assmp:param_direction} we have that $\frac{\bw^{(t)}}{\norm{\bw^{(t)}}}=\bar\bw^{\infty}+\delta^{(t)}$ with $\delta^{(t)}\to 0$, then using taylor approximation we have\begin{align*}
\nabla m\left(\bw^{(t)}\right)&=\sum_{i=1}^{d} \norm{\bw^{(t)}}^{i-1}\nabla m^i\left(\frac{\bw^{(t)}}{\norm{\bw^{(t)}}}\right)\\
&=\sum_{i=1}^{d} \norm{\bw^{(t)}}^{i-1}\nabla m^i\left(\bar\bw^{\infty}+\delta^{(t)}\right)\\
&=\sum_{i=1}^{d} \norm{\bw^{(t)}}^{i-1} \sum_{j=1}^i \nabla ^j m^i\left(\bar\bw^{\infty}\right)\left[(\delta^{(t)})^{\otimes j-1}\right]
\end{align*}
Where $(\delta^{(t)})^{\otimes j}=(\delta^{(t)},\ldots, \delta^{(t)})$ $i$ times. Now $\norm{\nabla ^j m^i\left(\bar\bw^{\infty}\right)\left[(\delta^{(t)})^{\otimes j}\right]}=O\left(\norm{\delta^{(t)}}^{j-1}\right)$, and hence if $\bar\bw^\infty$ zeros out some of the top homogenizations of $m$ and their gradients, the dominant terms may come from either the expansions of lower degree homogenizations, or expansions of those canceled top homogenizations, i.e. $\delta^{(t)}\norm{\bw^{(t)}}^k \overset{?}{\lessgtr} \norm{\bw^{(t)}}^{k-1}$ depending on how fast or slow $\delta^{(t)}\to0$.
All in all, we find that there's a wide range of possible dynamical behaviors when optimizing polynomials with the exponential loss, some of which change the resulting implicit bias as we see in \Cref{Section:Results}. We leave it for future work to investigate these cases, and focus on \Cref{assmp:effdeg}. We verify all of our assumptions experimentally in \Cref{Section:Experiments}.

\subsubsection{Proof of \Cref{lemma:all_fail}} \label{Section:assumptions_fail}
In the following example, we demonstrate that in a simple setting, our effective degree assumption may fail. Our example is in $2D$ and 3 linear autoregressive steps. We take $\x=\begin{pmatrix}
    0\\ -1/3
\end{pmatrix}, y=1$, and
\[
\cp(\bw)=\begin{pmatrix}
    w_1^2+w_1w_2^2 \\
    2w_1w_2 + w_2^3
\end{pmatrix}
\]

So we have $\cl(\bw)=\exp(\frac{2}{3}w_1w_2+\frac{1}{3}w_2^3)$. The idea is that the optimization path follows a relation of $w_1 \sim -w_2^2$, so we have $\bar\bw^{\infty}=\begin{pmatrix}
    -1\\0
\end{pmatrix}$ and $\norm{\bw}\sim w_2^2$ while the margin $\sim w_2^{3}$, hence the effective degree is $3/2$, not an integer. A way to think about what happens is that $\bw^{(t)}/\norm{\bw^{(t)}}$ converges slow enough, that the radial speed in the direction of $\bar\bw^{\infty}$ is significantly slower than the radial speed of $\bw$. Let us analyze this formally:
\begin{align*}
   &\frac{\partial \cl}{\partial w_1}=\frac{2}{3}\cdot\cl(\bw)w_2\\
   &\frac{\partial \cl}{\partial w_2}=\cl(\bw)\left(\frac{2}{3}w_1 +w_2^2\right)
\end{align*}
For a fixed learning rate $\eta$, denote $\alpha^{(t)}=\eta \cl(\bw^{(t)})$, then the gradient descent update gives
\begin{align*}
   &w_1^{(t+1)}=w_1^{(t)}-\alpha^{(t)}\frac{2}{3}w_2\\
   &w_2^{(t+1)}=w_2^{(t)}-\alpha^{(t)}\left(\frac{2}{3}w_1 +w_2^2\right)
\end{align*}
Let us start the optimization from $\bw=\begin{pmatrix}
-2 \\ 1
\end{pmatrix}$.
Note that if $d:=-\frac{2}{3}w_1 - w_2^2\geq 0$ and $w_2\geq 0 $, then $w_2$ is non-decreasing. We will show $d^{(t)}\geq0$ $\forall t$. Clearly $d^{(0)}=1/3$. Let us see how $d$ evolves:
\begin{align}
d^{(t+1)}&=-\frac{2}{3}w_1^{(t+1)} -\left(w_2^{(t+1)}\right)^2 \nonumber\\
&=-\frac{2}{3}w_1^{(t)}+\alpha^{(t)}\frac{4}{9}w^{(t)}_2-\left(w_2^{(t)}+\alpha^{(t)}d^{(t)}\right)^2\nonumber\\
&=-\frac{2}{3}w_1^{(t)}+\alpha^{(t)}\frac{4}{9}w^{(t)}_2-\left(w_2^{(t)}\right)^2-2w_2^{(t)}\alpha^{(t)}d^{(t)}-\left(\alpha^{(t)}d^{(t)}\right)^2\nonumber\\
&=d^{(t)} -\left(\alpha^{(t)}d^{(t)}\right)^2 + \alpha^{(t)}w_2^{(t)}\left(\frac{4}{9}-2d^{(t)}\right) \label{eq:evo}
\end{align}

So whenever $d^{(t)}=0$, $d^{(t+1)} =\frac{4}{9} \alpha^{(t)}w_2^{(t)}>0$ as long as $w_2^{(t)}$ is positive. First we will show that $d^{(t)}$ is upper bounded. If $d^{(t)}\geq 4/9$, then $4/9-2d^{(t)}\leq-d^{(t)}$ and so by \Cref{eq:evo}, $d^{(t+1)}\leq d^{(t)}-\alpha^{(t)}w_2^{(t)}d^{(t)}\leq d^{(t)}$ as long as $w_2^{(t)}$ is positive. Hence $d^{(t)}\leq \max\{4/9,d^{(0)}\}=4/9$.

Now whenever $d^{(t)}\geq 0$, we have
\[
d^{(t+1)} \geq d^{(t)}-\left(\alpha^{(t)}d^{(t)}\right)^2-2\alpha^{(t)}d^{(t)}w_2^{(t)}=d^{(t)}\left(1- d^{(t)}\left(\alpha^{(t)}\right)^2-2\alpha^{(t)}w_2^{(t)}\right)
\]
so it's enough to show $  -d^{(t)}\left(\alpha^{(t)}\right)^2-2\alpha^{(t)}w_2^{(t)} \geq-1$ for $d^{(t)}$ to stay non-negative.
Note that we can rewrite $\frac{2}{3}w_1^{(t)}w_2^{(t)}+\frac{1}{3}\left(w_2^{(t)}\right)^3=-\frac{2}{3}\left(w_2^{(t)}\right)^3-d^{(t)}w_2^{(t)}$. Let us bound
\begin{align}
    \alpha^{(t)}w_2^{(t)}&\leq \eta w_2^{(t)}\exp\left(\frac{2}{3}w_1^{(t)}w_2^{(t)}+\frac{1}{3}\left(w_2^{(t)}\right)^3\right)\nonumber\\
    &=\eta w_2^{(t)}\exp\left(-\frac{2}{3}\left(w_2^{(t)}\right)^3-d^{(t)}w_2^{(t)}\right) \nonumber\\
    &\leq\eta w_2^{(t)}\exp\left(-\frac{2}{3}\left(w_2^{(t)}\right)^3\right)\leq0.6\eta \label{eq:alphaw}
\end{align}
Where the last inequality is by finding the maximum of $w_2^{(t)}\exp\left(-\frac{2}{3}\left(w_2^{(t)}\right)^3\right)$ for $w_2^{(t)}\geq0$. Hence choosing for example $\eta=0.1$ we get
\[
-d^{(t)}\left(\alpha^{(t)}\right)^2-2\alpha^{(t)}w_2^{(t)} \geq-4/9\eta^2-2\alpha^{(t)}w_2^{(t)}\geq-0.1-0.2 \geq -1
\]
So we have that $d^{(t)}\geq 0$, and so $w_2^{(t)}$ is non-decreasing. We want to show that it's unbounded.
Assume towards contradiction that it is bounded, i.e. $w_2^{(t)} \leq U$, then we have that $\cl(\bw^{(t)})=\exp\left(-\frac{2}{3}\left(w_2^{(t)}\right)^3-d^{(t)}w_2^{(t)}\right) \geq \cl_{\inf}>0$. Now
\[
w_1^{(t)}-\frac{2}{3}\alpha^{(t)}w_2^{(t)} \leq w_1^{(t)}-\frac{2}{3}\alpha^{(t)}
\]
Since we start from $w_2^{(0)}=1$ and it's non-decreasing. Finally, we get
\[
w_1^{(t+1)}\leq w_1^{(t)}-\frac{2}{3}\eta \cl_{\inf}
\]
So $w_1^{(t)}\to -\infty$, and hence $d^{(t)}=-\frac{2}{3}w_1^{(t)}-\left(w_2^{(t)}\right)^2\to\infty$, contradicting its upper bound. Hence we have $w_2^{(t)}\to\infty$, and similarly, $\cl(\bw^{(t)})\to0$.

Now consider the following sequence
\[
e^{(t)}=d^{(t)}-2/9
\]
We want to show it converges to 0. Note that
\[
e^{(t+1)}-e^{(t)}=-\left(\alpha^{(t)}d^{(t)}\right)^2+\alpha^{(t)}w_2^{(t)}\left(\frac{4}{9}-2d^{(t)}\right)=-\left(\alpha^{(t)}d^{(t)}\right)^2-2\alpha^{(t)}w_2^{(t)}e^{(t)}
\]
Hence
\[
e^{(t+1)}=e^{(t)}\left(1-2\alpha^{(t)}w_2^{(t)}\right)-\left(\alpha^{(t)}d^{(t)}\right)^2
\]
Now from \Cref{eq:alphaw}, and positivity of $w_2^{(t)}$ we have that for large enough $t$, $\left(1-2\alpha^{(t)}w_2^{(t)}\right)\in(0,1)$. We want to show that $\sum_i\left(\alpha^{(t)}d^{(t)}\right)^2<\infty$. Recall that
\[
w_2^{(t+1)} -w_2^{(t)}=\alpha^{(t)}d^{(t)}\leq \frac{4}{9}\eta\exp\left(-\frac{2}{3}\left(w^{(t)}_2\right)^3\right)
\]
Let $I_n:=\{t: w_2^{(t)}\in[n, n+1)\}$, then for large enough $n$, we have that $\sum_{t\in I_n} \left(w_2^{(t+1)}-w_2^{(t)}\right)=\sum_{t\in I_n} \alpha^{(t)}d^{(t)}\leq 2$. Moreover, $\max_{t\in I_n}\alpha^{(t)}d^{(t)}\leq \frac{4}{9}\eta \exp\left(-\frac{2}{3}n^3\right)$ so overall
\[
\sum_{t\in I_n} \left(\alpha^{(t)}d^{(t)}\right)^2\leq\max_{t\in I_n}\alpha^{(t)}d^{(t)}\sum_{t\in I_n} \left(\alpha^{(t)}d^{(t)}\right)\leq \frac{8}{9}\eta\exp\left(-\frac{2}{3}n^3\right)
\]
and overall
\[
\sum_{t} \left(\alpha^{(t)}d^{(t)}\right)^2=\sum_n \sum_{t\in I_n}\left(\alpha^{(t)}d^{(t)}\right)^2\leq\sum_n \frac{8}{9}\eta\exp\left(-\frac{2}{3}n^3\right) < \infty
\]
Note that since $-\left(w^{(t)}_2\right)^2-\frac{2}{3}w^{(t)}_1=d^{(t)}\geq0$ we have that $w_1^{(t)}\to -\infty$. And
\[
w_1^{(t)}=w_1^{(0)}-\frac{2}{3}\sum_{k=0}^{t-1}\alpha^{(k)}w_2^{(k)}\to-\infty
\]
So overall $\sum_{k=0}^{t-1}\alpha^{(k)}w_2^{(k)}\to\infty$. Now let us define $e^{(t)}:=d^{(t)}-2/9$, we want to use \Cref{lemma:recursion-convergence} to show $e^{(t)}\to 0$. We have
\[
e^{(t+1)}=e^{(t)}(1-2\alpha^{(t)}w_2^{(t)})-\left(\alpha^{(t)}d^{(t)}\right)^2
\]
so indeed using \Cref{lemma:recursion-convergence} we get $d^{(t)}\to 2/9$. But now
\[
w_1^{(t)}=-\frac{3}{2}\left(w_2^{(t)}\right)^2-1/3 + o(1)
\]
and so
\[
\frac{2}{3}\left(w_2^{(t)}\right)^3+d^{(t)}w_2^{(t)} =\Theta\left( \left(w_2^{(t)}\right)^3\right)
\]
but $\norm{\bw^{(t)}}=\Theta\left(\left(w_2^{(t)}\right)^2\right)$, so the margin is $\Theta\left(\norm{\bw^{(t)}}^{3/2}\right)$

\subsection{Proof of \Cref{lemma:assump_implication}} \label{Section:implication_proof}
\Cref{lemma:bartlett} implies \Cref{assmp:minimization}. Theorem 6.2 in \citet{bartlett} (\Cref{thm:bartlett:62}) implies \Cref{assmp:param_direction}.
\paragraph{For the case where $q$ is $M$-homogeneous} these two imply \Cref{assmp:effdeg} immediately by homogeneity.
\paragraph{For the case where $q$ is not homogeneous and $\deg(p)\geq1$} let us bound the loss explicitly:
\begin{align}
 \cl(\bw^{(t)})&=\sum_{i=1}^N \exp\left(-y_i\inner{q(\bw^{(t)}),\x_i}\right) \nonumber\\
&\geq \max_{i\in[N]}\exp\left(-y_i\inner{q(\bw^{(t)}),\x_i}\right) \label{eq:loss_lowerbound}
\end{align}
We have the margin as defined in \citet{bartlett} is,
\begin{equation*}
\gamma(t):=\frac{\exp\left(\Phi\left(\exp\left(p\left(\norm{\bw^{(t)}}\right)\right)\cl\left(\bw^{(t)}\right)/N\right)\right)}{\norm{\bw^{(t)}}^M}
\end{equation*}
Where $M=\deg(q)$, and $\Phi(x)=\log(\log(1/Nx))+2/\log(Nx)$. Since $\exp\circ\Phi(x)$ is monotonous decreasing, using \Cref{eq:loss_lowerbound} and for large enough $t$ s.t. $p\left(\norm{\bw^{(t)}}\right)\geq 1$,
\begin{align}
\gamma(t)&\leq\frac{\exp\left(\Phi\left(\max_{i\in[N]}\exp\left(-y_i\inner{q(\bw^{(t)}),\x_i}\right)/N\right)\right)}{\norm{\bw^{(t)}}^M}\nonumber\\
&=\frac{\exp\left(\log\log\left(\frac{1}{\max_{i\in[N]}\exp\left(-y_i\inner{q(\bw^{(t)}),\x_i}\right)}\right)+\frac{2}{\log\left(\max_{i\in[N]}\exp\left(-y_i\inner{q(\bw^{(t)}),\x_i}\right)\right)}\right)}{\norm{\bw^{(t)}}^M}\nonumber\\
&=\frac{\exp\left(\log\left(\min_{i\in[N]}y_i\inner{q(\bw^{(t)}),\x_i}\right)+\frac{2}{\max_{i\in[N]}\left(-y_i\inner{q(\bw^{(t)}),\x_i}\right)}\right)}{\norm{\bw^{(t)}}^M} \nonumber\\
&\leq \frac{\exp\left(\log\left(\min_{i\in[N]}y_i\inner{q(\bw^{(t)}),\x_i}\right)\right)}{\norm{\bw^{(t)}}^M} \label{eq:implication_explain}\\
&=\frac{\min_{i\in[N]}y_i\inner{q(\bw^{(t)}),\x_i}}{\norm{\bw^{(t)}}^M} \label{eq:final_margin_lowerbound}
\end{align}
Where \Cref{eq:implication_explain} is due to $\frac{2}{\max_{i\in[N]}\left(-y_i\inner{q(\bw^{(t)}),\x_i}\right)}$ being negative for large enough $t$. Taking the limit on both sides of \Cref{eq:final_margin_lowerbound}, and using \Cref{thm:bartlett:61} showing that $\gamma(t)$ is positive and increasing for large enough $t$, and since we have that $\deg(q)=M$, then the normalized margin converges to some positive constant. For simplicity let us denote the margin polynomials $m_i(\bw):=y_i\inner{q(\bw^{(t)}),\x_i}$, then using homogeneity we have
\[
\frac{m(\bw^{(t)})}{\norm{\bw^{(t)}}^{M}}=\sum_{i=0}^{M}\norm{\bw^{(t)}}^{i-M}m^{i}\left(\frac{\bw^{(t)}}{\norm{\bw^{(t)}}}\right)
\]
Taking the limit, one summand must not go to zero (because we know the limit is positive), hence it must be the last one, and we get $\lim_{t\to\infty}\frac{m(\bw^{(t)})}{\norm{\bw^{(t)}}^{M}}=m^{M}\left(\bar \bw^{\infty}\right)>0$.
Similarly, from homogeneity we get that $ \inner{\bar\bw^{\infty},\nabla m^{M}\left(\bar \bw^{\infty}\right)}=Mm^{M}\left(\bar \bw^{\infty}\right)>0$, hence $\nabla m^{M-1}\left(\bar \bw^{\infty}\right)\neq 0$, and must dominate the gradient. Thus, we have concluded \Cref{assmp:effdeg}.

\paragraph{For the case where $q$ is not homogeneous and $\deg(p)=0$}
We have that $p\left(\norm{\bw^{(t)}}\right)$ is a constant and hence
\[
\gamma(t):=\frac{\exp\left(\Phi\left(C\cl\left(\bw^{(t)}\right)/N\right)\right)}{\norm{\bw^{(t)}}^M}
\]
For some $C>0$. The rest of the argument follows exactly as above, with a trailing constant $C$, that does not change the sign of the limits, leading to the same conclusion.
\section{Multidimensional Tokens}
\label{Section:MultidimToken}

\subsection{Linear Autoregressive Models}
Given an example $X\in \reals^{l\times D}$ which we interpret as the initial part of some sequence $X=X_{[1:D]} = (\x_1,\ldots, \x_D)$ (i.e $\x_i$ denote the $i^{th}$ column of $X$), we compute a sequence of $T$ autoregressive steps, the final's sign being our label. Formally, the first step will be $\x_{D+1}=X\bw$. The second step  $X_{D+2}=X_{2:D+1}\bw$, and at the $T^{th}$ step we'll have $x_{D+T}=X_{T:D+T-1}\bw$. For example, let $D=2$ and let us consider two autoregressive steps ($T=2$). We have that
\begin{align*}
\x_3 &= X\bw = \begin{pmatrix}
    w_1x_{1,1} + w_2x_{1,2}\\
    w_1x_{2,1} + w_2x_{2,2}
\end{pmatrix} \\
\x_4 &=  X_{[2:D+T-1]}\bw = X_{[2:3]}\bw\\ &=\begin{pmatrix}
    w_1 x_{1,2} + w_2 x_{1,3}\\
    w_1 x_{2,2} + w_2 x_{2,3}
\end{pmatrix} =  \\
&= \begin{pmatrix}
    w_1w_2x_{1,1}+(w_1+w^2_2)x_{1,2}\\
    w_1w_2x_{2,1}+(w_1+w^2_2)x_{2,2}
\end{pmatrix}
\end{align*}
Note that in contrast to \Cref{Section:Linear}, the resulting $\x_4$ cannot serve as a label by itself, and hence we must fix, or learn a projection vector $\bp\in \reals^l$, s.t. $y=\sign(\inner{\bp, \x_4})$.
We denote the vector of coefficients of $x_1, \ldots, x_D$ in $x_{D+T}$ as $\cp$, a polynomial in $\bw$, and optionally $\bp$. For our running example that vector would be $\cp(\bw,\bp)=\begin{pmatrix}
    w_1w_2p_1 & w_1p_1+w_2^2p_1 \\ w_1w_2p_2 & w_1p_2+w_2^2p_2
\end{pmatrix}$. Using this notation, we have that $\inner{\cp(\bw,\bp), X} = x_{D+T}$, and in our example $x_4=\tr\left(\begin{pmatrix}
    w_1w_2p_1 & w_1p_1+w_2^2p_1 \\ w_1w_2p_2 & w_1p_2+w_2^2p_2
\end{pmatrix} \begin{pmatrix}
    x_{1,1} & x_{1,2}\\
    x_{2,1} & x_{2,2}
\end{pmatrix}^\top\right)$.
Note that for a dataset $\{X_i, y_i\}_{n=1}^N$ we denote the polynomial corresponding to the $i^{th}$-example as $\cp_i$, but in this autoregressive setting, it is the case that $\cp_i=\cp_j$ for $i\neq j$, if both examples $i,j$ are labeled by the same number $T$ of autoregressive steps.
We state the theorem for the case of learning $\bp$, but it can easily be amended for fixed $\bp$.
\begin{theorem}[Implicit Bias of Multidimensional Linear Autoregressive Models]
    For parameters $\bw \in \reals^{D}$ and $\bp\in \reals^l$ denote $\theta:=(\bw, \bp)$, let $\{X_i,y_i\}_{i=1}^N$, where $y_i=\sign\inner{\cp_{i}(\theta), X_i}$. Consider the sequence of gradient updates according to \Cref{eq:gradient_updates}, with any non-increasing sequence of step sizes $\{\eta_t\}$,  minimizing the empirical risk. If \Cref{assmp:minimization,assmp:param_direction,assmp:effdeg}, then $\bar \theta^{\infty}:=\lim_{t\to\infty}\frac{\theta^{(t)}}{\norm{\theta^{(t)}}}$ is a positive scaling of a first order stationary point of the following optimization problem:

\begin{align*}
    {\min}_{\theta\in \reals^{D+l}}\norm{\theta}_2^2 \text{ s.t. } &\forall i, y_i \inner{X_i,\cp^{\nu_{i}}_{i}(\theta)} \geq 0 \\
    &\forall i\in Z, y_iC_i \inner{X_i,\cp^{\nu_{i}}_{i}(\theta)}\geq 1
\end{align*}
where  $Z=\arg\min_{i\in[n]}\left\{\effdeg_\theta\left(\inner{\cp_i ,\x_iy_i}\right)\right\}$.
\end{theorem}
\begin{proof}
    The proof follows immediately from \Cref{thm:main}, by concatenating $\theta=(\bw, \bp)$ into one vector, concatenating each $X_i$'s columns into a vector $\bv_i$, and concatenating $\cp_i$'s columns into a vector $\tilde\cp_i$ the same way, such that they match and then $\inner{\bv_i, \tilde \cp_i(\theta)}=\inner{X_i, \cp_i(\bw, \bp)}$.
\end{proof}
\subsection{Autoregressive SSMs}
Here we consider the full Autoregressive SSMs, as it subsumes the simpler SSM case. We repeat the notation for convenience. We denote the parameters of the model by $A\in \reals^{d_A\times d_A}$, $B\in \reals^{d_A\times d}$, $C\in \reals^{d_A \times d}$. Consider an example $X \in \reals^{d\times l}$, that we can think of as an input prompt, or as a beginning of some sequence $X = X_{1:l} = \begin{pmatrix}
x_{1,1}& \ldots& x_{1,l}
\\
\vdots & \ddots &\vdots
\\
x_{d,1}&\ldots& x_{d,l}
\end{pmatrix}$, we compute a sequence of $T$ autoregressive steps:
\begin{equation}
    (X)_{l+1}=C^\top \sum_{i=1}^{l}A^{l-i} B (X)_{i},\quad (X)_{l+2}=C^\top \sum_{i=1}^{l+1}A^{l+1-i} B (X)_{i},\quad\dots\quad, (X)_{l+T}=C^\top \sum_{i=1}^{l+T-1}A^{l+T-1-i} B (X)_{i}~.
\end{equation}
Where $(X)_i$ denotes the $i^{th}$ column of $X$
The label of the sequence $\x$ is defined to be $\sign(\inner{\x_{l+T}, \bp})$, where $\cp$ is some projection vector that is required in order to get a scalar prediction. This vector may be learned or fixed. We phrase the theorem for the general case, where $\cp$ can be learned
\begin{theorem}[Implicit bias of SSMs]
    For parameters $A\in \reals^{d_A\times d_A}$, $B\in \reals^{d_A\times d}$, $C\in \reals^{d_A\times d}$ and $\bp\in \reals^d$ denoted for simplicity by $\bw:=(A,B,C,\bp)$, let $\{X_i,y_i\}_{i=1}^N$, where $y_i=\sign\inner{\cq_{i}(\bw), X_i}$.
    Consider the sequence of gradient updates according to \Cref{eq:gradient_updates},
    with a non-increasing sequence of step sizes $\{\eta_t\}$, minimizing the empirical risk. If \Cref{assmp:minimization,assmp:param_direction,assmp:effdeg}, $\bar \bw^{\infty}:=\lim_{t\to\infty}\frac{\bw^{(t)}}{\norm{\bw^{(t)}}}$ is a positive scaling of a first order stationary point of the following optimization problem:
\begin{align*}
    {\min}_{\bw\in \reals^{d_A\times d_A + 2d_A\times d+d}}\norm{\bw}_2^2 \text{ s.t. } &\forall i, y_i \inner{X_i,\cq^{\nu_{i}}_{i}(\bw)} \geq 0 \\
    &\forall i\in Z, y_i C_i \inner{X_i,\cq^{\nu_{i}}_{i}(\bw)}\geq 1
\end{align*}
where $Z=\arg\min_{i\in[n]}\left\{\effdeg_{\bw}\left(\inner{\cq_i, \x_i}\right)\right\}$.
\end{theorem}
\begin{proof}
    The proof follows immediately from \Cref{thm:main}, by concatenating $\bw=(A,B,C, \bp)$ into one vector, and noting that since at each step we only multiply by the parameters and not by $X_i$, we can write a coefficient vector for every $x_{i,j}$ in terms of $A,B,C$ and $\bp$ only, then using that vector as our polynomial, and concatenating each $X_i$ to a long vector respecting these coefficients.
\end{proof}
\section{Proof of \Cref{lemma:bartlett_fails}} \label{Section:example_bartlett}

Consider the following simple toy example in 4D: Take one example $\x=\begin{pmatrix}
    0, 0, 0, 1
\end{pmatrix}, y=1$
with 4-autoregressive steps.
We have that
\begin{align*}
    x_5&=w_1x_1+w_2x_2+w_3x_3+w_4x_4\\
    x_6&=w_1x_2+w_2x_3+w_3x_4+w_4x_5 = w_3+w_4^2\\
    x_7&=w_1x_3+w_2x_4+w_3x_5+w_4x_6 = w_2+2w_3w_4+w_4^3\\
    x_8&=w_1x_4+w_2x_5+w_3x_6+w_4x_7 = w_1+2w_2w_4+w_3^2+3w_3w_4^2 +w_4^4
\end{align*}

So we have that
\[
\cl(\bw)=\exp(-w_1-2w_2w_4-w_3^2-3w_3w_4^2 -w_4^4)
\]
The idea behind this toy example is that
\[
-\nabla \cl(\bw)=\cl(\bw)\cdot\begin{pmatrix}
    1 \\ 2w_4 \\ 2w_3+3w_4^2 \\ 2w_2+6w_3w_4+4w_4^3
\end{pmatrix}
\]
Hence whenever $w_2=w_3=w_4=0$ we get that
\[
-\nabla \cl(\bw)=\cl(\bw)\cdot\begin{pmatrix}
    1 \\ 0 \\ 0 \\ 0
\end{pmatrix}
\]
That is, $w_2,w_3, w_4$ remain $0$. This produces the dynamic that while $\deg(\cp(\bw) ^\top \x y)=4$, we have that $\effdeg_{\bw}(\cp(\bw) ^\top \x y)=1$. Effectively the higher parts are not used during optimization.

We will first show why the assumptions of \citet{bartlett} fail, and proceed to show why our assumptions hold.
\subsection{Proof of \Cref{lemma:bartlett_fails}}
Let us first restate the assumption:
\begin{assumption} [\citet{bartlett}, simplified for our example]
At some point $t_0$ during optimization, $\cl(\bw^{(t_0)}) \leq \exp\left(-p\left(\norm{\bw^{(t_0)}}\right)\right)$ for some $p(x)$. If the predictor is a polynomial of the form $f(\bw;\x)=\inner{q(\bw),\x}$, then $\deg(p) = \deg(q)-1$.
\end{assumption}
They later use this assumption to prove the following:
\begin{lemma}[\citet{bartlett}, simplified]\label{lemma:bartlett}
$\forall t\geq t_0$ we have that $\cl(\bw^{(t)}) \leq \exp\left(-p\left(\norm{\bw^{(t)}}\right)\right)$
\end{lemma}
Let us analyze the example in light of \Cref{assmp:bartlett}.
First, recall that
\[
\cl(\bw)=\exp(-w_1-2w_2w_4-w_3^2-3w_3w_4^2 -w_4^4)
\]
and
\[
-\nabla \cl(\bw)=\cl(\bw)\cdot\begin{pmatrix}
    1 \\ 2w_4 \\ 2w_3+3w_4^2 \\ 2w_2+6w_3w_4+4w_4^3
\end{pmatrix}
\]
Initializing at $\bw^{(0)}=\begin{pmatrix}
    0\\0\\0\\0
\end{pmatrix}$
and whenever $w_2=w_3=w_4=0$, we get
\begin{equation}
-\nabla \cl(\bw)=\cl(\bw)\cdot\begin{pmatrix}
    1 \\ 0 \\ 0 \\ 0
\end{pmatrix} \label{eq:ex1:grad}
\end{equation}

So assuming a constant step size $\eta$,
\begin{align}
    w_1^{(t+1)}&=w_1^{(t)}+\eta \cl(\bw^{(t)})\nonumber\\
    w_j^{(t+1)}&=w_j^{(t)}+0 \quad j\in\{2,3,4\} \label{eq:ex1:w2}
\end{align}
Where the updates follow from inductively $w_2^{(t)}\equiv w_3^{(t)}\equiv w_4^{(t)}\equiv 0$ staying zero for any $t$. This means that during training we have:
\begin{equation}
\norm{\bw^{(t)}}=\abs{w_1^{(t)}}  \label{eq:ex1:norm}
\end{equation}
 Furthermore, $w_1^{(t)}$ is strictly increasing. We claim that $-w_1^{(t)}\to -\infty$. Assume otherwise, then $w_1^{(t)}\to L\in \reals$, but then the steps can be bounded from below i.e. $\bw_1^{(t)}-\bw_1^{(t-1)}\geq  \eta \exp^{-L}>0$, which is a contradiction to $w_1^{(t)}\to L$. So we have
\begin{equation}
    -w_1^{(t)}\to -\infty \label{eq:ex1:infinity}
\end{equation}

But then $\cl(\bw^{(t)}) = \exp(-w^{(t)}_1)=\exp\left(-\norm{\bw^{(t)}}\right)$. Now assume towards contradiction that \Cref{assmp:bartlett} holds, then by \Cref{lemma:bartlett} we get that there exists some $t_0$ s.t. $\forall t\geq t_0$, $\cl(\bw^{(t)})\leq \exp\left(-p(\norm{\bw})\right)$, but $\cl(\bw^{(t)})=\exp\left(-\norm{\bw^{(t)}}\right)\geq\exp(-p(\norm{\bw})$ contradiction, since $\norm{\bw^{(t)}}\to\infty$ and so for any $p$ s.t. $\deg(p)=3$ with positive leading coefficient, we have $\norm{\bw^{(t)}}=o\left(p(\norm{\bw})\right)$. Consequently  \Cref{assmp:bartlett} must not hold.

\subsection{Our Assumptions Hold}
We want to show \Cref{assmp:minimization,assmp:param_direction,assmp:effdeg} hold. By \Cref{eq:ex1:w2} and \Cref{eq:ex1:grad}, we have that \Cref{assmp:param_direction} holds, since $\nabla \cl(\bw)$ always points in the same direction. From \Cref{eq:ex1:infinity}  and \Cref{eq:ex1:w2} we have that $\cl(\bw^{(t)}) = \exp(-w^{(t)}_1)\to 0$ so \Cref{assmp:minimization} holds.
To show \Cref{assmp:effdeg} holds true, note that
\[
\lim_{t\to\infty} \frac{w^{(t)}_1+2w_2^{(t)}w_4^{(t)}+(w_3^{(t)})^2+3w_3^{(t)}(w_4^{(t)})^2 +(w_4^{(t)})^4 }{\norm{\bw^{(t)}}^2}=\frac{w^{(t)}_1}{(w_1^{(t)})^2}\to0
\]
On the other hand:
\[
\lim_{t\to\infty} \frac{w^{(t)}_1+2w_2^{(t)}w_4^{(t)}+(w_3^{(t)})^2+3w_3^{(t)}(w_4^{(t)})^2 +(w_4^{(t)})^4}{\norm{\bw^{(t)}}}=\frac{w^{(t)}_1}{\abs{w^{(t)}_1}}=1=\cp^1(\bar\bw^{\infty})^\top \x y
\]
Now for the corresponding gradient,
\[
\lim_{t\to\infty} \frac{\begin{pmatrix}
    1 \\ 0 \\ 0 \\ 0
\end{pmatrix}}{\norm{\bw^{(t)}}^0}=\begin{pmatrix}
    1 \\ 0 \\ 0 \\ 0
\end{pmatrix}=\nabla \cp^1(\bar\bw^{\infty})^\top\x y
\]
And hence \Cref{assmp:minimization,assmp:param_direction,assmp:effdeg} all hold.
\subsection{Constructing General Negative Examples to \Cref{assmp:bartlett}} \label{Section:remark}
We can use the principles from the proof techniques of \Cref{lemma:bartlett_fails}, i.e. \Cref{Section:example_bartlett}, to create polynomials for which similar dynamics hold. For simplicity, assume we have one example $\x=\begin{pmatrix}
        1\\1
    \end{pmatrix},y=1$. For every even $d\geq2$  and $k<d$ we construct a polynomial such that the margin would be with degree $d$ and effective degree $k$:
    Take $\cp(\bw)=\begin{pmatrix}
        -\left(\sum_{i\in A}w_i\right)^k\\
        \left(\sum_{i\in B}w_i\right)^{d}
    \end{pmatrix}$,
    for some $A\cap B=\emptyset$. The exponential loss would be:
    \[
    \cl(\bw)=\exp\left(-\left(\sum_{i\in A}w_i\right)^k+\left(\sum_{i\in B}w_i\right)^{d}\right)~.
    \]
    This will yield a similar type of dynamics. In fact, this will yield a stronger dynamic, since this will have effective degree $k$ for almost all initializations in the case of even $k$, and a constant portion of initializations in the case of odd $k$.
    \begin{lemma}[\Cref{assmp:bartlett} does not hold for almost all initializations]\label{lem:almostall}
    Let $d\geq2$ be an even integer, and $0<k<d$ another even integer. Let $\x=\begin{pmatrix}
        1\\1
    \end{pmatrix},y=1$, and consider $\cp(\bw)=\begin{pmatrix}
        -\left(\sum_{i\in A}w_i\right)^k\\
        \left(\sum_{i\in B}w_i\right)^{d}
    \end{pmatrix}$,
    for some $A\cap B=\emptyset$. Then for almost all initializations, optimizing
    \[
     \cl(\bw)=\exp\left(-\left(\sum_{i\in A}w_i\right)^k+\left(\sum_{i\in B}w_i\right)^{d}\right)~.
    \]
    With gradient descent yields $\cl(\bw^{(t)})=\exp\left(-\Theta\left(\norm{\bw^{(t)}}^k\right)\right)$
    \end{lemma}

    This can be further generalized to larger families of polynomials, building on the same insight. Importantly, our main result about implicit bias for non-homogeneous polynomials applies to all of these constructions \Cref{thm:main}.
    \begin{proof}
        First note that the gradient is of rank-2.
        Denote $u(\bw)=\sum_{i\in A}w_i$, $v(\bw)=\sum_{i\in B}w_i$. Then for every $i\in A$ we have
        \[
        \frac{\partial \cl}{\partial w_i}(\bw)=\cl(\bw)\frac{\partial (-u^k+v^d)}{\partial w_i}(\bw)=-\cl(\bw)ku^{k-1}
        \]
        Similarly for $i\in B$, we have
        \[
        \frac{\partial \cl}{\partial w_i}(\bw)=\cl(\bw)\frac{\partial (-u^k+v^d)}{\partial w_i}(\bw)=\cl(\bw)du^{d-1}
        \]
        Whenever $i\notin A \cup B$,
        \[
        \frac{\partial \cl}{\partial w_i}=0
        \]
        Hence $\forall i,i'\in A$  $w_i-w_{i'}$ is constant. Similarly, $\forall i,i'\in B$, $w_i-w_{i'}$ is constant.

        So we can re-write the update per $u,v$, i.e.
        \begin{align}
            u_{t+1}&=u_t+\eta |A| k u_t^{k-1}\exp(-u_t^k+v^d) \label{eq:u_update}\\
            v_{t+1}&=u_t+\eta |B| d v_t^{d-1}\exp(-u_t^k+v^d) \nonumber
        \end{align}
        The rest of $\bw$ is fixed. Hence optimizing $u$ and $v$ will yield similar dynamics to optimizing $\bw$.
        Denote $R:=|v_0|+1$, $M:=\max\{0, -u_0^k\}$, and choose $\eta \leq \frac{\exp(-R^d-M)}{|B|dR^{d-2}}$.

        Now, since $u_{t+1}=u_t+\eta |A| k u_t^{k-1}\exp(-u_t^k+v^d)$, we have that since $k$ is even, $u^k_t\geq0$ and so $-u_t^k\leq 0 \leq M$ hence $\exp(-u_t^k+v^d_t)\leq \exp(M+v_t^d)$.

        Now,
        \[
        \abs{v_{t+1}}=\abs{v_t}\abs{1-\eta  d \abs{v_t}^{d-2}|B|\exp(-u_t^k-v_t^d)}
        \]
        Assume $|v_t|\leq R$, then $\eta  d \abs{v_t}^{d-2}|B|\exp(-u_t^k-v_t^d)\leq1$ from our choice of $\eta$., hence $\abs{1-\eta  d \abs{v_t}^{d-2}|B|\exp(-u_t^k-v_t^d)}\in [0,1]$. So $\abs{v_{t+1}}\leq|v_t|\leq R$, hence by induction $|v_t|\leq R$ for all $t$.

        Now since $k$ is even, $k-1$ is odd and from \Cref{eq:u_update} we have that $u_t$ never changes sign along training and $|u_t|$ is strictly increasing.
        Now assume toward contradiction that $|u_t|$ is bounded, then we can bound its update by $\abs{\eta |A|ku_0^{k-1}\exp(-B-R^d)}\leq\abs{\eta |A| k u_t^{k-1}\exp(-u_t^k+v^d)}$, i.e. the steps of $u_t$ are lower bounded by a constant, in contradiction to it being bounded. Finally, since $u^t=\sum_{i\in A}w_i$ and those are the only coordinates that diverge, we have that $\cl(\bw^{(t)})\leq \exp\left(-\Theta\left(\norm{\bw^{(t)}}^k\right)\right)$
    \end{proof}
    \begin{lemma}
    Let $d\geq2$ be an even integer, and $0<k<d$ an odd integer. Let $\x=\begin{pmatrix}
        1\\1
    \end{pmatrix},y=1$, and consider $\cp(\bw)=\begin{pmatrix}
        -\left(\sum_{i\in A}w_i\right)^k\\
        \left(\sum_{i\in B}w_i\right)^{d}
    \end{pmatrix}$,
    for some $A\cap B=\emptyset$. Further assume that $\left(\sum_{i\in A}w_i\right)^k>0$ at initialization. Then, optimizing
    \[
     \cl(\bw)=\exp\left(-\left(\sum_{i\in A}w_i\right)^k+\left(\sum_{i\in B}w_i\right)^{d}\right)~.
    \]
    With gradient descent yields $\cl(\bw^{(t)})=\exp\left(-\Theta\left(\norm{\bw^{(t)}}^k\right)\right)$
    \end{lemma}
    \begin{proof}
    The proof is similar to the proof of \Cref{lem:almostall}, except the boundedness argument, in which case:
        \begin{itemize}
            \item If $k$ is even, $u^k_t\geq0$ and so $-u_t^k\leq 0 \leq M$
            \item If $k$ is odd, $u^{k-1}_t\geq0$, and hence $u_t$ is non-decreasing, so $-u^t_k\leq  -u_0^k \leq -M$.
        \end{itemize}
    The rest of the proof is similar since we assume $u_0>0$.
    \end{proof}

\section{Effective Degree Characteristics}
\subsection{The Effective Degree Depends on the Model, Data, and Initialization} \label{subsec:init_dependence}
First we formalize \Cref{lem:different_init}:
\begin{lemma}
Let $\bw=(u,v)\in \reals^2$, define the polynomial predictor $\cp(\bw)=\begin{pmatrix}
    q(\bw) \\
    q(\bw)^2\\
    \vdots \\
    q(\bw)^R
\end{pmatrix}$, with \[q(\bw)=q(u,v)=u^2+v^4-u^2v^2.\] For $r\in [R]$, let $\x_r=\be_r$, $y_r=1$. Then, $m_r(\bw)=y_r\inner{\cp(\bw), \x_r}=q(\bw)^r$. Consider gradient descent with a sufficiently small constant step size on $\cl(\bw)=\sum_{i=1}^R \exp(-q(\bw)^r)$. For the open sets of initializations
\[
U_2=\{u>2, |v|<1/2\}, \qquad U_4=\{|u|<1/2, v>2\}
\]
\[\effdeg(m_r)=\begin{cases}
    2r & \bw^{(0)}\in U_2\\
    4r & \bw^{(0)}\in U_4
\end{cases}.
\]
\end{lemma}
\begin{proof}
Let us denote $p_2=u^2$, $p_4=v^4-u^2v^2$ i.e. $q=p_2+p_4$, these polynomials will be the effective polynomials in the two regimes.

By \Cref{lemma:effdeg_char_asymptotics} we have that for $\bw^{(0)}\in U_2$, $u_t \to \infty$, $v_t\to 0$, $u_tv_t\to 0$, hence $\frac{\bw^{(t)}}{\norm{\bw^{(t)}}}\to \be_1$.
\[
\frac{q(\bw^{(t)})}{\norm{\bw^{(t)}}^2}\to 1=p_2(\be_1).
\]
We have
\[
\nabla q(u, v)=(2u(1-v^2), 2v(2v^2-u^2)),
\]
And so,
\[
\frac{\nabla q(\bw^{(t)})}{\norm{\bw^{(t)}}}\to (2, 0)=\nabla p_2(\be_1).
\]
That is, $\effdeg(q)=2$. In a similar manner, by \Cref{lemma:effdeg_char_asymptotics} if $\bw^{(0)}\in U_4$, then $u_t \to 0$, $v_t=\infty$. So, $\frac{\bw^{(t)}}{\norm{\bw^{(t)}}} \to \be_2$. Furthermore,
\[
\frac{q(\bw^{(t)})}{\norm{\bw^{(t)}}^4}\to 1=p_4(\be_2),
\]
and
\[
\frac{\nabla q(\bw^{(t)})}{\norm{\bw^{(t)}}^3}\to (0,4)=\nabla p_4(\be_2),
\]
that is, $\effdeg(q)=4$. Finally, recall that $m_r=q^r$. Using the fact that $\frac{m_r(\bw)}{\norm{\bw}^{rk}}=\left(\frac{q(\bw)}{\norm{\bw}^{k}}\right)^r$, and $\frac{m_r(\bw)}{\norm{\bw}^{rk-1}}=r\left(\frac{q(\bw)}{\norm{\bw}^{k}}\right)^{r-1}\frac{\nabla q(\bw)}{\norm{\bw}^{k-1}}$ we finish the proof.
\end{proof}
\begin{lemma}\label{lemma:effdeg_char_asymptotics}
Let $\bw=(u,v)\in \reals^2$, define the polynomial predictor $\cp(\bw)=\begin{pmatrix}
    q(\bw) \\
    q(\bw)^2\\
    \vdots \\
    q(\bw)^R
\end{pmatrix}$, with \[q(\bw)=q(u,v)=u^2+v^4-u^2v^2.\] For $r\in [R]$, let $\x_r=\be_r$, $y_r=1$. Then, $m_r(\bw)=y_r\inner{\cp(\bw), \x_r}=q(\bw)^r$. Consider gradient descent with a sufficiently small constant step size on $\cl(\bw)=\sum_{i=1}^R \exp(-q(\bw)^r)$. For the open sets of initializations
\[
U_2=\{u>2, |v|<1/2\}, \qquad U_4=\{|u|<1/2, v>2\},
\]
and a sufficiently small learning rate $\eta>0$, if $\bw^{(0)}\in U_2$, $\bw^{(t)}\in U_2$, $u_t\to\infty$, $v_t\to 0$, $u_tv_t\to0$.
If $\bw^{(0)}\in U_4$, $\bw^{(t)}\in U_4$, $u_t\to0$, $v_t \to \infty$.
\end{lemma}
The proof follows by showing the path is invariant to $U_2$ and $U_4$, and the dynamic in each one forces the corresponding asymptotic behavior of the coordinates.
\begin{proof}
Note that $\nabla m_r(\bw)=rq(bw)^{r-1}\nabla q(\bw)$, hence $\nabla \cl(\bw)$ is always a scalar multiple of $\nabla q(\bw)$, i.e.
\[
-\nabla \cl(\bw)=a(q(\bw))\nabla q(\bw)
\]
For $a(s)=\sum_{r=1}^R rs^{r-1}\exp(-s^r)$, which is positive whenever $t>0$, and in that case we can think of $\bw^{(t)}$ following a gradient-ascent path w.r.t. $q$, for appropriate step sizes.
The coordinate updates are then:
\begin{equation}\label{sec:characteristics:eq:update}
u_{t+1}=u_t\left(1+2\eta \cdot a(q(\bw^{(t)}))(1-v_t^2)\right), \qquad v_{t+1}=v_t\left(1+2\eta \cdot a(q(\bw^{(t)}))(2v_t^2-u_t^2)\right)
\end{equation}

Note that $a(s)$ decays exponentially, so we can find
\[
M_1=\sup_{s\geq3} s\cdot a(s)<\infty, \qquad M_{1/2}=\sup_{s\geq3} s^{1/2}\cdot a(s)<\infty,
\]
We will choose $\eta > 0$ s.t.
\[
\frac{8}{3}M_1 <1/2, \qquad \frac{8}{\sqrt{15}} \eta M_{1/2}<1/2.
\]
\paragraph{If $\bw^{(t)}\in U_2$:} Then $q(\bw^{(t)})=u_t^2(1-v_t^2)+v_t^4\geq \frac{3}{4}u_t^2>3$, and so $u_t^2\leq \frac{4}{3}q(\bw^{(t)})$. Furthermore, $1-v_t^2>0$, so $u_{t+1}>u_t$, and since $2v_t^2-u_t^2<0$, we have
\[
0>2\eta \cdot a(q(\bw^{(t)}))(2v_t^2-u_t^2)>-2\eta \cdot a(q(\bw^{(t)}))u_t^2 \geq -\frac{8}{3}\eta \cdot a(q(\bw^{(t)})) > -1/2,
\]
from the choice of $\eta$ and $u_t^2\leq \frac{4}{3}q(\bw^{(t)})$. Plugging in to the update for $v_{t+1}$ we get  $|v_{t+1}|<|v_t|$, so $\bw^{(t+1)}\in U_2$. We got that $u_t$ is increasing. If we assume it is bounded, we get that $q(\bw^{(t)})$ must be bounded, but then we can lower bound $a(q(\bw^{(t)})) \geq C>0$. Together with $1-v_t^2 \geq 3/4$ we get $u_{t+1}\geq u_t\left(1+\frac{3}{2}\eta C\right)$, contradiction. So $u_t \to \infty$.

Now we claim that
\begin{equation} \label{sec:characteristics:eq:u_2_divergence}
\sum_{i=0}^\infty 2\eta a( q(\bw^{(t)}))(u_t^2-2v_t^2)=\infty
\end{equation}. Indeed,
\begin{equation} \label{sec:characteristics:eq:u_2_divergence2}
\log\frac{u_{t+1}}{u_t}=\log(1+2a(q(\bw^{(t)})))\leq2a(q(\bw^{(t)})),
\end{equation}

and since the left hand side sums to infinity, so must the right hand side. On $U_2$ we have that $u_t^2-2v_t^2>2/7$, and hence the original sum must diverge.

Now for $\bw^{(t)}\in U_2$ we have that
\begin{align*}
     0&<2\eta a( q(\bw^{(t)}))(u_t^2-2v_t^2)\\
     &<2\eta a( q(\bw^{(t)}))u_t^2\\
     &\leq 2 \eta a( q(\bw^{(t)}))\frac{4}{3}q(\bw^{(t)})\\
     &=\frac{8}{3} q(\bw^{(t)})\eta a( q(\bw^{(t)}))\\
     &\leq \frac{8}{3} \eta M_1 < 1/2.
\end{align*}
Recall that $|v_{t+1}|=|v_t|(1-2\eta a( q(\bw^{(t)}))(u_t^2-2v_t^2))$. Together with  \eqref{sec:characteristics:eq:u_2_divergence} and the bound on the constant multiplying $|v_t|$, we get that $v_t\to 0$. Finally, by \eqref{sec:characteristics:eq:update}
\[
u_{t+1}|v_{t+1}|=u_t|v_t|\left(1+2\eta \cdot a(q(\bw^{(t)}))(2v_t^2-u_t^2)\right)\left(1+2\eta \cdot a(q(\bw^{(t)}))(1-v_t^2)\right).
\]
Using $\log(1+x)\leq x$,
\[
\log \frac{u_{t+1}|v_{t+1}|}{u_t|v_t|}\leq\left(1+2\eta \cdot a(q(\bw^{(t)}))(2v_t^2-u_t^2)\right)+2\eta \cdot a(q(\bw^{(t)}))(1-v_t^2)<-\frac{11}{2}\eta a(q(\bw^{(t)}))
\]
Where the last inequality is due to $\bw^{(t)}\in U_2$. Now summing the above equation over $t$ the r.h.s. diverges to $-\infty$, because the sum over $t$ in \Cref{sec:characteristics:eq:u_2_divergence2} diverges to infinity. The bound on the l.h.s. sum yields $|v_t|u_t \to 0$.
\paragraph{If $\bw^{(t)}\in U_4$:} then $u_t^2<1/4$, $v_t > 2$. Furthermore,
\begin{align*}
0&<2\eta a(q(\bw^{(t)}))(v_t^2-1)\\
&<    2\eta a(q(\bw^{(t)}))v_t^2 \\
&\overset{(*)}{\leq} \frac{8}{\sqrt{15}}\eta q(\bw^{(t)})^{1/2}a(q(\bw^{(t)})) < 1/2.
\end{align*}
Where $(*)$ is due to the fact that under $U_4$,
\begin{equation}\label{sec:characteristics:eq:lowerbound}
q(\bw^{(t)})\geq\frac{15}{16}v_t^4.
\end{equation}
Recalling the update rule in \Cref{sec:characteristics:eq:update}, we get $|u_{t+1}|<|u_t|$, and since $2v_t^2-u_t^2 > 0$, then $v_{t+1}>v_{t}$. That is, $\bw^{(t+1)} \in U_4$.

Similarly to the prior case, if $v_t$ were bounded then so were $q(\bw^{(t)})$, forcing geometric growth for $v_t$, which is a contradiction. Hence $v_t\to \infty$. In a similar technique to the prior case, we would like to show $\sum_{t=0}^\infty 2 \eta a(q(\bw^{(t)}))(v_t^2-1)$. First note that $2 \eta a(q(\bw^{(t)}))(v_t^2-1)\geq \frac{2}{3} \eta a(q(\bw^{(t)}))v_t^2$, so showing that summing the r.h.s. over $t$ diverges suffices. We have
\[
\log \frac{v_{t+1}}{v_t}\leq \log \left(1+2\eta a(q(\bw^{(t)}))(2v_t^2-u_t^2)\right)\leq4\eta a(q(\bw^{(t)}))v_t^2.
\]
Since summing the l.h.s. diverges to infinity, so does summing the r.h.s., and hence every scalar multiplication of it. Now $|u_{t+1}|=|u_t|(1-2 \eta a(q(\bw^{(t)}))(v_t^2-1))$. From the divergence above and the fact that
\begin{eqnarray*}
0&<&2 \eta a(q(\bw^{(t)}))(v_t^2-1)\\
&\leq&2 \eta a(q(\bw^{(t)}))v_t^2 \\
&\overset{\eqref{sec:characteristics:eq:lowerbound}}{\leq}&\frac{8}{\sqrt{15}}\eta q(\bw^{(t)})^{1/2}a(q(\bw^{(t)}))\\
&\leq & \frac{8}{\sqrt{15}}\eta M_{1/2}<1/2,
\end{eqnarray*}
we have that $u_t\to0$.
\end{proof}
\subsection{Determining the Effective Degree Without Solving the Dynamics} \label{subsec:simple_calc}
We next provide a polynomial predictor $\cp:\reals^D\to\reals^n$, and a dataset $\{\x_i, y_i\}_{i=1}^n$, such that when trained with gradient descent (\Cref{eq:gradient_updates}) with a constant step size $\eta>0$, the margins $m_i=y_i\inner{\cp, \x_i}$ have different effective degrees, and $\effdeg(m_i)$ can be derived by a short calculation.

Note that the proof works for any subset of the above dataset, with arbitrary labels, and the polynomial predictor can be extended to a family of polynomials for which the above holds. So this yields in fact a family of such examples.

Consider the parameter vector $(u,v)=\bw \in \reals^2$ and the polynomial $P:\reals^2 \to \reals ^N$
\[
P(u,v)=\begin{pmatrix}
u^{\nu_1}-v^d\\
u^{\nu_2}-v^d\\
\vdots \\
u^{\nu_N}-v^d
\end{pmatrix},
\]
for $2\leq \nu_1 <\ldots<\nu_N<d$, with $d$ even.
Let $y_i\in \{\pm1\}$ some labels, choose $\x_i=y_i\be_i\in \reals^N$.

Now $m_i(u,v)=u^{\nu_i}-v^d$. For  $0<\eta <\frac{1}{dN}$ and initializing with $u_0>1$, and $|v_0|<1$, we will show $\effdeg(m_i)=\nu_i$. The gradient updates are:
\begin{align*}\label{sec:effdeg:eq:update}
u_{t+1}&=u_t+\eta \sum_{i=1}^N\nu_iu_t^{\nu_i-1}\exp(v_t^d-u_t^{\nu_i})\\
v_{t+1}&=v_t-\eta d v_t^{d-1} \sum_{i=1}^N\exp(v_t^d-u_t^{\nu_i})\\
&=v_t\left(1-\eta d |v_t^{d-2}| \sum_{i=1}^N\exp(v_t^d-u_t^{\nu_i})\right) &&(\text{$d$ is even})
\end{align*}
Now while $u_t\geq 1$ and $|v_t|\leq 1$, we have $v_t^d-u_t^{\nu_i} \leq 0$. So
\[
0\leq\eta d |v_t^{d-2}| \sum_{i=1}^N\exp(v_t^d-u_t^{\nu_i}) \leq \eta d N \leq 1
\]
So overall $|v_{t+1}|<|v_t|$, and $u_{t+1}\geq u_t$ since all of the $u$ updates are positive.

We have that the optimization stays in the open interval $u_t >u_0 \geq 1$, $|v_t|<|v_0|<1$.

If we assume $u_t\leq M$, then
\[
u_{t+1}-u_t\geq\eta \nu_1 u_t^{\nu_1-1}\exp\left(v_t^d-u_t^{\nu_1}\right)\geq
\eta\nu_1 u_0^{\nu_1-1}e^{-M^{\nu_1}}>0.
\]
Hence $u_t\to \infty$, contradiction. A straight forward calculation concludes the proof.

\section{Proof of \Cref{thm:main}} \label{proof:main}
\subsection{Proof Intuition}
\label{Section:Intuition}
We have that $\bar\bw^{\infty}$ exists from \Cref{assmp:param_direction}. We use \Cref{assmp:minimization} to show this implies $\bar\bw^{\infty}$ must create positive margins, and hence primal feasibility. \Cref{assmp:effdeg} implies two different things:
\begin{itemize}
    \item The margins are dominated by their $\nu_i$-homogeneous parts, and hence $\norm{\cp_i}$ scale like $\norm{\bw}^{\nu_i}$.
    \item The margins' gradients are dominated by their $\nu_i-1$-homogeneous parts, and hence $\norm{\nabla\cp_i}$ scale like $\norm{\bw}^{\nu_i-1}$.
\end{itemize}
The first is then used to transform the non-homogeneous margin $m_i:=y_i\inner{\cp_i,\x_i}$ into the homogeneous polynomial in the KKT conditions, i.e. $m^{\nu_i}=\inner{y_iC_i\cp_i^{\nu_i}, \x_i }$, and also to show that the gradients of lower-effective-degree margins dominate the gradient, since their corresponding loss-exponent decays slower (the exponent of every loss-per-sample decays as $\norm{\bw}^{\nu_i}$). Among the lowest $\effdeg$ margins, we further show that those that converge to a lower normalized margin dominate their higher-normalized-margin counterparts. Overall this tells us that $\nabla\cl=\sum_{i\in S}\alpha_i\nabla m_i+o\left(\sum_{i\in S}\alpha_i\nabla m_i\right)$, where $S$ consists of the lowest effective degree polynomials with the lowest normalized margin, i.e. the gradient is dominated by a possibly dynamic convex sum of margins from $S$. From \Cref{assmp:effdeg}, $\nabla m_i\approx C_i\nabla m_i^{\nu_{Z}}$ where $\nu_{Z}$ is the minimal effective degree and hence $(\bar\bw^{\infty})^\top\sum_{i\in S}\alpha_i\nabla m_i\approx \gamma_{\min}\nu_{Z}$, where $\gamma_{\min}$ is the minimum margin in that defines $S$. i.e. the loss has a constant projection onto $\bar\bw^{\infty}$.
We then attempt to show that the projection onto the space orthogonal to $\bar\bw^{\infty}$ vanishes, at least in some subsequence of the gradients. Towards this we show that the GD updates induce an approximate exponential-multiplicative-weights step on $\alpha_i$, which can be thought of as the dual variables. We further show that this multiplicative-weights update is one that minimizes the gradient norm along some subsequence. Finally, we show that the minimizer must have no component orthogonal to $\bar\bw^{\infty}$.
This implies that there exists a subsequence for which $\nabla \cl \propto \bar\bw^{\infty}$, which shows optimality and complementary slackness. Dual feasibility is due to $\alpha_i$ being weights of a convex sum.
\subsection{Proof}
For convenience, let us restate the theorem:

\begin{theorem}
Let $\{\cp_i\}_{i=1}^N$ a collection of polynomials with parameters $\bw\in \reals^D$. For a dataset $\{\x_i, y_i\}_{i=1}^N$ where $y_i=\sign \left(\inner{\cp_{i}(\bw) ,\x_i}\right)$, consider the sequence of gradient updates (\Cref{eq:gradient_updates}) with a non-increasing sequence of step sizes $\{\eta_t\}_t$, minimizing the empirical risk.
If \Cref{assmp:minimization,assmp:param_direction,assmp:effdeg} hold we have that $\bar \bw^{\infty}:=\lim_{t\to \infty}\frac{\bw^{(t)}}{\norm{\bw^{(t)}}}$ is a positive scaling of a first order stationary point (KKT point) of the following optimization problem:
    \begin{align*}
    {\min}_{\bw\in \reals^d}\norm{\bw}_2^2 \text{ s.t. } &\forall i, y_i \inner{\x_i,\cp^{\nu_{i}}_{i}(\bw)} \geq 0 && \tag{\text{OPT}}\\
    &\forall i\in Z, y_i C_i \inner{\x_i,\cp^{\nu_{i}}_{i}(\bw)}\geq 1~,
    \end{align*}
     where $Z=\arg \min_{i\in[n]}\left\{\effdeg_{\bw}y_i\inner{\cp_i, \x_i}\right\}$.
\end{theorem}

We would like to show that $\bar \bw^{\infty}$ is in a direction of a KKT point of \Cref{eq:OPT}. First, let us state the KKT conditions:
\begin{itemize}
    \item Primal feasibility: $\forall i,  y_i \inner{\x_i,\cp^{\nu_i}_{i}(\bw)} \geq 0$, $\forall i\in Z, y_i C_i\inner{\x_i,\cp^{\nu_i}_{i}(\bw)}\geq 1$
    \item Dual feasibility and complementary slackness: There are $\{\mu_i\}_{i=1}^N$ where $\mu_i\geq 0$ and
    \begin{itemize}
        \item $\forall i \notin Z, \text{ if } y_i \inner{\x_i,\cp^{\nu_i}_{i}(\bw)} > 0, \;\mu_i=0$
        \item $\forall i \in Z, \text{ if } y_i C_i \inner{\x_i,\cp^{\nu_i}_{i}(\bw)} > 1, \;\mu_i=0$
    \end{itemize}

    \item Optimality: $\bw=\sum_{i=1}^{N}\mu_i\cdot C_i\nabla\cp_{i}^{\nu_i}(\bw)^\top  y_i\x_i$

\end{itemize}

First we will show that any positive scaling of $\bar\bw^{\infty}$ is a feasible solution.
\begin{lemma}[Primal Feasibility] \label{proof:exp:primal_feasibility}
$\bar\bw^{\infty}$ classifies all examples correctly, and for $C>0$ large enough, $C\bar\bw^{\infty}$ is a feasible solution to \Cref{eq:OPT} under \Cref{assmp:minimization,assmp:param_direction,assmp:effdeg}.
\end{lemma}
\begin{proof}
By \Cref{assmp:effdeg} we have that $\frac{\cp_i(\bw^{(t)})^\top \x_i}{\norm{\bw^{(t)}}^{\nu_i}}-\delta_i^{(t)}=C_i\cp_i^{\nu_i}(\bar \bw^\infty)^\top \x_i \neq 0$ with $\delta_i^{(t)}\to 0$. Hence from homogeneity $C_i\cp_i^{\nu_i}(C\bar \bw^\infty)^\top \x_i \neq 0$. Assume towards contradiction that $C_i\cp_i^{\nu_i}(C\bar \bw^\infty)^\top \x_iy_i<0$, then we have
$$
\cp_i(\bw^{(t)})^\top \x_iy_i=\left(\frac{C_i\cp_i^{\nu_i}(C\bar \bw^\infty)^\top \x_iy_i}{C^{\nu_i}}+\delta^{(t)}_i\right)\norm{\bw^{(t)}}^{\nu_i} \to - \infty
$$
Since $\delta^{(t)}_i\to 0$, contradicting \Cref{assmp:minimization}. All that is left is scaling $C$ such that all margins are in the feasible region.
\end{proof}
To show the rest of the KKT conditions, we want to understand the asymptotic direction of the gradient.
\begin{lemma}[Minimum Margin of Minimum Effective Degree Dominates the Gradient]
$-\overline{\nabla\cl(\bar \bw ^{(t)})}+\delta_{\nabla}^{(t)} \propto \sum_{i\in S}\alpha^{(t)}_i C_i\nabla\cp_i^{\nu_i}\left(\bw^{(t)}\right)^\top\x_iy_i$ for $\delta_{\nabla}^{(t)}\to 0$, where $S=\arg\min_{i\in Z}\{C_i \cp^{\nu_Z}_i(\bar\bw^{\infty})^\top \x_iy_i\}$, where $S$ is a set, and $Z=\arg\min_{i\in[n]}\left\{\effdeg\left(\cp_i^\top(\bar\bw^{\infty}) \x_iy_i\right)\right\}$.
\end{lemma}
\begin{proof}
Consider the gradient of the loss:
\begin{align*}
-\nabla\cl\left(\bw^{(t)}\right)&=
\sum_{i=1}^n \nabla\cp_i\left(\bw^{(t)}\right)^\top\x_i y_i \exp\left(-\cp_i\left(\bw^{(t)}\right)^\top\x_i y_i\right)\\
&=
\sum_{i=1}^n \nabla\cp_i\left(\bw^{(t)}\right)^\top\x_i y_i \exp\left(-g^{\nu_i}(t)\left(C_i\cp^{\nu_i}_i\left(\bar\bw^{\infty}\right)^\top\x_i y_i + \delta^{(t)}_i y_i\right)\right) && \text{(by \cref{assmp:effdeg})}
\end{align*}
Now denoting $\gamma_i:=C_i\cp^{\nu_i}_i\left(\bar\bw^{\infty}\right)^\top\x_i y_i>0$ (this is positive from \Cref{proof:exp:primal_feasibility}), we get
\begin{align*}
-\nabla\cl\left(\bw^{(t)}\right)&=
\sum_{i=1}^n \nabla\cp_i\left(\bw^{(t)}\right)^\top\x_i y_i \exp\left(-g^{\nu_i}(t)\left(\gamma_i + \delta^{(t)}_i y_i\right)\right)\\
\end{align*}

Now we want to divide the loss into two separate parts. Denote $\nu_{Z}:=\min_i\{\effdeg\left(\cp_i(\cdot)^\top\x_i\right)\}$ ($Z$ is chosen because for every $k\in Z$ $\effdeg\left(\cp_i(\cdot)^\top\x_i\right)=\nu_{Z}$). Let us write
\[
A(t)=\sum_{i\in Z}\nabla\cp_i\left(\bw^{(t)}\right)\x_iy_i^\top \exp\left(-g^{\nu_{Z}}(t)\left(\gamma_{i}+\delta_{i}^{(t)}\right)\right)
\]
and
\[
B(t)=\sum_{i\notin Z}\nabla\cp_i\left(\bw^{(t)}\right)\x_iy_i^\top \exp\left(-g^{\nu_i}(t)\left(\gamma_{i}+\delta_{i}^{(t)}\right)\right)
\]
Note that $A(t)+B(t)=-\nabla \cl(\bw^{(t)})$.
We will now further divide $A(t)$ into two parts. First denote $\gamma_{\min}:=\min_i \gamma_i$,
(recall they are all equal and minimal), and let $\frac{{\min}_{i\in Z\setminus S}\{\gamma_i\}+\gamma_{\min}}{2}=\hat \gamma$. The idea is that we define $\hat \gamma$ to be the average between the minimum margin (of the minimum effective degree), and the second-smallest margin (of the minimum $\effdeg$). Then, when we will have for $i\in Z \setminus S$ $\gamma_i-\hat\gamma>0$ while $\gamma_{\min}-\hat\gamma<0$. Concretely, denote:

\[
\mathrm{I}(t):=\frac{\sum_{i\in S}\exp\left(-g^{\nu_{Z}}(t)\left(\gamma_{\min}+\delta_{i}^{(t)}\right)\right)\nabla\cp_i(\bw^{(t)})^\top \x_i y_i}{\exp\left(-\hat \gamma \cdot g^{\nu_{Z}}(t)\right)}
\]

\[
\mathrm{II}(t):=\frac{\sum_{i\in Z \setminus S}\exp\left(-g^{\nu_{Z}}(t)\left(\gamma_i+\delta_{i}^{(t)}\right)\right)\nabla\cp_i(\bw^{(t)})^\top \x_i y_i}{\exp\left(-\hat \gamma \cdot g^{\nu_{Z}}(t)\right)}
\]
We chose $\mathrm{I}, \mathrm{II}$ such that $A(t)=\exp(-\hat \gamma \cdot g^{\nu_{Z}}(t))(\mathrm{I}(t)+\mathrm{II}(t))$.
We can then write
\begin{equation}
    \norm{A(t)}\geq\exp(-\hat \gamma \cdot g^{\nu_{Z}}(t))\cdot \abs{ \norm{\mathrm{I}(t)} - \norm{\mathrm{II}(t)}} \label{eq:bound_on_A_norm}
\end{equation}

Inspecting the norm
\[
\norm{\mathrm{I}(t)}=\norm{\sum_{i\in S}\exp\left(-g^{\nu_{Z}}(t)\left(\gamma_{\min}-\hat\gamma+\delta_{i}^{(t)}\right)\right)\nabla\cp_i(\bw^{(t)})^\top \x_i y_i}
\]
We note that for large enough $t$ since $\delta_i^{(t)}\to0$ and by definition of $\hat \gamma$
\begin{equation}
\gamma_{\min}-\hat\gamma+\delta_{i}^{(t)}<0 \label{eq:I_to_infty}
\end{equation}

We want to use this to show $\norm{\mathrm{I}(t)}\to \infty$. Here, a subtle point is to verify the linear combination doesn't asymptotically cancel out. Indeed
\begin{align*}
\mathrm{I}(t)&=\sum_{i\in S}\exp\left(-g^{\nu_{Z}}(t)\left(\gamma_{\min}-\hat\gamma+\delta_{i}^{(t)}\right)\right)\nabla\cp_i(\bw^{(t)})^\top \x_i y_i \\
&=\sum_{i\in S}\exp\left(-g^{\nu_{Z}}(t)\left(\gamma_{\min}-\hat\gamma+\delta_{i}^{(t)}\right)\right)\left(C_i\nabla\cp^{\nu_{Z}}_i(\bar\bw^{\infty})^\top\x_i y_i +\delta_{\nabla,i}^{(t)}\right) && \text{(\Cref{assmp:effdeg})}
\end{align*}
Now from homogeneity we have that
\[
(\bar\bw^\infty)^{\top}\ C_i\nabla\cp^{\nu_{Z}}_i(\bar\bw^{\infty})y_i\x_i=\nu_{Z}C_i\cp^{\nu_{Z}}_i(\bar\bw^{\infty})y_i\x_i=\nu_{Z}\gamma_i>0
\]
Hence
\[
 \frac{(\bar\bw^{\infty})^\top\mathrm{I}(t)}{\norm{\mathrm{I}(t)}}=\frac{\sum_{i\in S}\exp\left(-g^{\nu_{Z}}(t)\left(\gamma_{\min}-\hat\gamma+\delta_{i}^{(t)}\right)\right)}{\norm{\sum_{i\in S}\exp\left(-g^{\nu_{Z}}(t)\left(\gamma_{\min}-\hat\gamma+\delta_{i}^{(t)}\right)\right)\left(C_i\nabla\cp^{\nu_{Z}}_i(\bar\bw^{\infty})^\top\x_i y_i +\delta_{\nabla,i}^{(t)}\right)}} \left(\nu_Z\gamma_{\min} +\delta_{\nabla,i}^{(t)}\right)
\]
Is bounded away from $0$, and so is $\frac{\mathrm{I}(t)}{\norm{\mathrm{I}(t)}}$.
Together with \Cref{eq:I_to_infty}, this yields $\norm{\mathrm{I}(t)}\to \infty$.

Now for $\mathrm{II}(t)$,
\[
\norm {\mathrm{II}(t)}=\norm{\frac{\sum_{i\in Z \setminus S}\exp\left(-g^{\nu_{Z}}(t)\left(\gamma_i+\delta_{i}^{(t)}\right)\right)\nabla\cp_i(\bw^{(t)})^\top \x_i y_i}{\exp\left(-\hat \gamma \cdot g^{\nu_{Z}}(t)\right)}}
\]
here again by the definition of $\hat \gamma$ and for large enough $t$, we have that for any $i \in Z \setminus S$

\[
\gamma_i-\hat \gamma + \delta_{i}^{(t)}>0
\]
therefore we can bound $\norm{\mathrm{II}(t)}\leq C$ for large enough $t$. Now consider the expression

\begin{align}
\norm{B(t)}&=\norm{\sum_{i\notin Z}\nabla\cp_i\left(\bw^{(t)}\right)\x_iy_i^\top \exp\left(-g^{\nu_i}(t)\left(\gamma_{i}+\delta_{i}^{(t)}\right)\right)} \\
&\leq \sum_{i\notin Z}\exp\left(-g^{\nu_i}(t)\left(\gamma_{i}+\delta_{i}^{(t)}\right)\right)\norm{\nabla\cp_i\left(\bw^{(t)}\right)\x_i } \\
&\leq\sum_{i\notin Z}g^{\nu_i-1}(t)\exp\left(-g^{\nu_i}(t)\left(\gamma_{i}+\delta_{i}^{(t)}\right)\right)(r_{\max}+C)
\end{align}

For some constant $C>0$, and $r_{\max} = \max_i \norm{C_i\nabla\cp^{\nu_i}_i(\bar\bw^{\infty})^\top \x_i}$. Overall using the bound on $\norm{A(t)}$ (\ref{eq:bound_on_A_norm}), we get that

\begin{align*}
\frac{\norm{B(t)}}{\norm{A(t)}}&\leq \frac{\sum_{i\notin Z}g^{\nu_i-1}(t)\exp\left(-g^{\nu_i}(t)\left(\gamma_{i}+\delta_{i}^{(t)}\right)\right)(r_{\max}+C)}{\exp(-\hat \gamma g^{\nu_{Z}}(t))\cdot \abs{ \norm{\mathrm{I}(t)} - \norm{\mathrm{II}(t)}}} \\
&=\frac{\sum_{i\notin Z}g^{\nu_i-1}(t)\exp\left(-g^{\nu_i}(t)\left(\gamma_{i}+\delta_{i}^{(t)}\right)+\hat\gamma g^{\nu_{Z}(t)}\right)(r_{\max}+C)}{ \abs{ \norm{\mathrm{I}(t)} - \norm{\mathrm{II}(t)}}}
\end{align*}
Now for $i\notin Z$, $-g^{\nu_i}(t)\left(\gamma_{i}+\delta_{i}^{(t)}\right)+\hat\gamma g^{\nu_{Z}}(t)$ is negative for large enough $t$. Since $\hat \gamma, \gamma_i + \delta_{i}^{(t)}$ are bounded and positive for large enough $t$, and $g^{\nu_i}(t)$ grows asymptotically faster than $g^{\nu_{Z}}(t)$. Therefore the expression in the exponent is negative for large enough $t$, hence the numerator goes to zero. The denominator goes to infinity due to $\norm{\mathrm{I}(t)}\to \infty$, and that $\norm{\mathrm{II}(t)}$ is bounded, so overall:
\begin{equation*}
    -\frac{\nabla \cl(\bw^{(t)})}{\norm{\nabla \cl(\bw^{(t)})}} = \frac{A(t)+B(t)}{\norm{A(t)+B(t)}}= \frac{A(t)}{\norm{A(t)}} + \tilde \delta^{(t)}
\end{equation*}
For some $\tilde\delta^{(t)}\to 0$.
Now similarly consider $\frac{\exp(-\hat \gamma g^{\nu_{Z}}(t))\cdot\norm{\mathrm{II}(t)}}{\exp(-\hat \gamma g^{\nu_{Z}}(t))\cdot\norm{\mathrm{I}(t)}}=\frac{\norm{\mathrm{II}(t)}}{\norm{\mathrm{I}(t)}}\to 0$, hence
\begin{equation*}
    \frac{A(t)}{\norm{A(t)}} =\frac{\mathrm{I}(t)}{\norm{\mathrm{I}(t)}}+ \hat \delta^{(t)}
\end{equation*}
For $\hat\delta^{(t)}\to 0$.
Overall
\[
    -\frac{\nabla \cl(\bw^{(t)})}{\norm{\nabla \cl(\bw^{(t)})}} \to \frac{\mathrm{I}(t)}{\norm{\mathrm{I}(t)}}
\]
i.e. the normalized gradient is dominated by $ \mathrm{I}(t)$.
\end{proof}
For simplicity of notation, let us denote $m_i(\bw):=\cp_i(\bw)^\top \x_iy_i$.
Let us define $\forall i\in S,\alpha_i^{(t)}:=\frac{\exp(-m_i(\bw^{(t)}))}{\cl_{|S}(\bw^{(t)})}$, where $\cl_{|S}=\sum_{i\in S}\exp(-m_i)\leq \cl$. By \Cref{assmp:effdeg}, we have that $\nabla m_i(\bw^{(t)})=\norm{\bw^{(t)}}^{\nu_i}\left(C_i\nabla m_i^{\nu_i}(\bar\bw^{\infty})+o(1)\right)$. So overall
\[
-\nabla\cl(\bw^{(t)})=\cl_{|S}(\bw^{(t)})g^{\nu_{Z}-1}(t)\left(\sum_{i\in S}\alpha_i^{(t)}C_i\nabla m_i^{\nu_{Z}}(\bar \bw^{\infty})+o(1)\right)
\]
Now note that $\Delta \bw^{(t)}:=\bw^{(t+1)}-\bw^{(t)}=-\eta_t \nabla \cl(\bw^{(t)})\to 0$, hence using taylor approximation we have
\[
m_i(\bw^{(t+1)})=m_i(\bw^{(t)})+\inner{\nabla m_i(\bw^{(t)}), \Delta\bw^{(t)}}+E_i^{(t)}
\]
With
\begin{align}
\abs{E_i^{(t)}}&\leq K_i\norm{\Delta \bw^{(t)}}^2g^{\deg(m_i)-2}(t) \nonumber\\
&= K_i\eta_t^2\norm{\nabla\cl( \bw^{(t)})}^2g^{\deg(m_i)-2}(t) \nonumber \\
&\leq K_iB^2\eta_t^2 \cl_{|S}^2(\bw^{(t)}) g^{2\nu_{Z}+\deg(m_i)-4}(t)\nonumber
\end{align}
For some constants $K_i$, and for some $B$ that bounds $\left(\sum_{i\in S}\alpha_i^{(t)}C_i\nabla m_i^{\nu_{Z}}(\bar \bw^{\infty})+o(1)\right)$.
Now let us compute the inner product from the above expression:
\[\inner{\nabla m_i(\bw^{(t)}), \Delta\bw^{(t)}}=\eta_t\sum_j\exp(-m_j(\bw^{(t)}))\inner{\nabla m_i(\bw^{(t)}),\nabla m_j(\bw^{(t)})}\]
So we have for $i\in S$
\[
m_i(\bw^{(t+1)})-m_i(\bw^{(t)})=\eta_t\cl_{|S}(\bw^{(t)})g^{2\nu_{Z}-2}(t)\left(\sum_j \alpha_j^{(t)}\inner{C_i\nabla m_i^{\nu_i}(\bar\bw^{\infty}), C_j\nabla m_j^{\nu_j}(\bar\bw^{\infty})}\right) + o(\eta_t\cl_{|S}(\bw^{(t)})g^{2\nu_{Z}-2}(t))
\]
Denote $\beta^{(t)}:=\eta_t\cl_{|S}(\bw^{(t)})g^{2\nu_{Z}-2}(t)$,  $A_{i,j}:=\inner{C_i\nabla m_i^{\nu_i}(\bar\bw^{\infty}), C_j\nabla m_j^{\nu_j}(\bar\bw^{\infty})}$ a gram matrix, then we can re-write $m_i(\bw^{(t+1)})-m_i(\bw^{(t)})=\beta^{(t)}(A \alpha^{(t)})_i + o(\beta^{(t)})$. Exponentiating we get

\begin{equation}
\exp(-m_i(\bw^{(t+1)})+m_i(\bw^{(t)}))=\exp(-\beta^{(t)}(A \alpha^{(t)})_i + o(\beta^{(t)})) \label{eq:exponentiation}
\end{equation}
Now, let us calculate the $\alpha^{(t)}_i$ update:
\begin{align}
    \alpha^{(t+1)}_i&=\frac{\exp(-m_i(\bw^{(t+1)}))}{\sum_{j\in S} \exp(-m_j(\bw^{(t+1)}))} \nonumber \\
    &=\frac{\exp(-m_i(\bw^{(t)}))\exp(-m_i(\bw^{(t+1)})+m_i(\bw^{(t)}))}{\sum_{j\in S}\exp(-m_j(\bw^{(t)})) \exp(-m_j(\bw^{(t+1)})+m_j(\bw^{(t)}))} && \text{(adding \& subtracting)}\nonumber\\
    &=\frac{\exp(-m_i(\bw^{(t)}))\left(\exp(-\beta^{(t)} (A\alpha^{(t)})_i+o(\beta^{(t)})\right)}{\sum_{j\in S}\exp(-m_j(\bw^{(t)})) \left(\exp(-\beta^{(t)} (A\alpha^{(t)})_j+o(\beta^{(t)})\right)} &&(\text{by \Cref{eq:exponentiation}})\nonumber\\
    &=\frac{\alpha^{(t)}_i\left(\exp(-\beta^{(t)} (A\alpha^{(t)})_i+o(\beta^{(t)})\right)}{\sum_{j\in S}\alpha^{(t)}_j \left(\exp(-\beta^{(t)} (A\alpha^{(t)})_j+o(\beta^{(t)})\right)} &&(\exp(-m_i(\bw^{(t)}))=\alpha_i^{(t)}\cl_{|S}(\bw^{(t)})) \label{eq:alpha_update}
\end{align}
Note that this is an Normalized Exponentiated Gradient update with perturbations.
\begin{lemma} \label{lemma:perturbed_exponentiated_grad}
    Let $\alpha^{(t)}$ be a sequence of vectors on the simplex, let $\alpha^*$ be some vector on the simplex, and let $\Phi(\alpha)$ be some convex function to be optimized. Denote $\ell^{(t)}=\nabla\Phi(\alpha^{(t)})$ and assume
    \[
    \alpha^{(t+1)}_i=\frac{\alpha^{(t)}_i\exp\left(-\beta^{(t)} \ell^{(t)}_i+\varepsilon^{(t)}_i\right)}{\sum_{j}\alpha^{(t)}_j \exp\left(-\beta^{(t)} \ell^{(t)}_j+\varepsilon^{(t)}_j\right)}
    \]
    For $\varepsilon^{(t)}\to 0$, then,
    \begin{equation} \label{eq:KL_ineq}
     \kl{\alpha^*}{\alpha^{(t+1)}}-\kl{\alpha^*}{\alpha^{(t)}}\leq \beta^{(t)}\left(\Phi(\alpha^*)-\Phi(\alpha^{(t)})\right)+(\beta^{(t)}\Delta^{(t)})^2/8 + 2\norm{\varepsilon^{(t)}}_\infty
    \end{equation}
    Where $\Delta^{(t)}:=\max_i \ell^{(t)}_i-\min_i \ell^{(t)}_i$.
\end{lemma}
\begin{proof}
    First, we will show
    \begin{equation}
     \kl{\alpha^*}{\alpha^{(t+1)}}-\kl{\alpha^*}{\alpha^{(t)}}\leq \beta^{(t)}\inner{\ell^{(t)}, \alpha^*-\alpha^{(t)}}+(\beta^{(t)}\Delta^{(t)})^2/8 + 2\norm{\varepsilon^{(t)}}_\infty
    \end{equation}
    For some generic $\ell^{(t)}$.
    Denote $Z_t:=\sum_j\alpha^{(t)}_j \exp\left(-\beta^{(t)} \ell^{(t)}_i+\varepsilon^{(t)}_i\right)$, so
    \[
    \alpha^{(t+1)}_i=\frac{\alpha^{(t)}_i\exp\left(-\beta^{(t)} \ell^{(t)}_i+\varepsilon^{(t)}_i\right)}{Z_t}
    \]
    Taking $\log$ over the update equation, we get $\log \alpha^{(t+1)}_i=\log(\alpha_i^{(t)})-\beta^{(t)}\ell^{(t)}_i+\varepsilon^{(t)}_i-\log(Z_t)$
    Hence,
    \begin{equation}
    \log \frac{\alpha_i^{(t)}}{\alpha^{(t+1)}_i}=\beta^{(t)}\ell^{(t)}_i-\varepsilon^{(t)}_i+\log(Z_t)
    \label{eq:logratio}
    \end{equation}

    So overall,
    \begin{align}
    \kl{\alpha^*}{\alpha^{(t+1)}}-\kl{\alpha^*}{\alpha^{(t)}}&=\sum_i\alpha_i^*\log\frac{\alpha_i^*}{\alpha^{(t+1)}_i}-\sum_i\alpha_i^*\log\frac{\alpha_i^*}{\alpha^{(t)}_i} \nonumber\\
    &=\sum_i\alpha_i^*\left(\log\frac{\alpha_i^*}{\alpha^{(t+1)}_i}+\log\frac{\alpha^{(t)}_i}{\alpha^*_i}\right)\nonumber\\
    &=\sum_i\alpha_i^*\log\frac{\alpha_i^{(t)}}{\alpha^{(t+1)}_i} \nonumber\\
    &=\sum_i\alpha_i^*\left(\beta^{(t)}\ell^{(t)}_i-\varepsilon^{(t)}_i+\log(Z_t)\right) && \text{(\cref{eq:logratio})}\nonumber\\
    &=\beta^{(t)}\inner{\alpha^*, \ell^{(t)}}-\inner{\alpha^*, \varepsilon^{(t)}} + \log Z_t \label{eq:kl_equality}
    \end{align}

    Now let us bound $\log Z_t$. Denote $\mu_t:=\inner{\alpha^{(t)}, \ell^{(t)}}$ (We can think as $\mu_t=\E_i \ell^{(t)}_i$, where $i$ is distributed by $\alpha^{(t)}$ -- a simplex vector). Then adding and subtracting $\mu_t$ we get:
    \[
    Z_t=\sum_i \alpha^{(t)}_i \exp(-\beta^{(t)}(\ell^{(t)}_i-\mu_t))\exp(\varepsilon^{(t)}_i)\exp(-\beta^{(t)}\mu_t)
    \]
    Hence,
    \begin{align*}
    \log Z_t&=-\beta^{(t)}\mu_t +\log \sum_i \alpha^{(t)}_i \exp(-\beta^{(t)}(\ell^{(t)}_i-\mu_t))\exp(\varepsilon^{(t)}_i)\\
    &\leq-\beta^{(t)}\mu_t +\log\left( \max_i\exp(\varepsilon^{(t)}_i)\sum_i \alpha^{(t)}_i \exp(-\beta^{(t)}(\ell^{(t)}_i-\mu_t))\right)\\
    &=-\beta^{(t)}\mu_t +\norm{\varepsilon^{(t)}}_\infty+\log\left(\sum_i \alpha^{(t)}_i \exp(-\beta^{(t)}(\ell^{(t)}_i-\mu_t))\right)
    \end{align*}
    Now again, since $\alpha^{(t)}$ is a simplex vector (or a probability distribution) we can use Hoeffding's Lemma (\Cref{lemma:hoeffding}), and get $\log\left(\sum_i \alpha^{(t)}_i \exp(-\beta^{(t)}(\ell^{(t)}_i-\mu_t))\right)\leq \frac{\left(\beta^{(t)}\Delta^{(t)}\right)^2}{8}$. Hence,
    \begin{equation*}
        \log Z_t \leq  -\beta^{(t)}\mu_t +\norm{\varepsilon^{(t)}}_\infty+\frac{\left(\beta^{(t)}\Delta^{(t)}\right)^2}{8}
    \end{equation*}
    Plugging it in \Cref{eq:kl_equality} we get
    \begin{align*}
    \kl{\alpha^*}{\alpha^{(t+1)}}-\kl{\alpha^*}{\alpha^{(t)}}&\leq \beta^{(t)}\inner{\ell^{(t)}, \alpha^*-\alpha^{(t)}}+\norm{\varepsilon^{(t)}}_\infty+\frac{\left(\beta^{(t)}\Delta^{(t)}\right)^2}{8}-\inner{\alpha^{*}, \varepsilon^{(t)}}\\
    &\leq \beta^{(t)}\inner{\ell^{(t)}, \alpha^*-\alpha^{(t)}}+2\norm{\varepsilon^{(t)}}_\infty+\frac{\left(\beta^{(t)}\Delta^{(t)}\right)^2}{8}\\
    \end{align*}
    Finally, take $\ell^{(t)}=\nabla\Phi(\alpha^{(t)})$. From the convexity of $\Phi$ we get $\Phi(\alpha^{*})-\Phi(\alpha^{(t)})\geq \inner{\nabla \Phi(\alpha^{(t)}), \alpha^*-\alpha^{(t)}}=\inner{\ell^{(t)}, \alpha^*-\alpha^{(t)}}$. Substitute to get what's required.
\end{proof}
\begin{lemma} \label{lemma:beta_stuff}
    \begin{enumerate}
        \item $\beta^{(t)} \to 0$ \label{lemma:betta_stuff:item1}
        \item $\sum_{t=0}^\infty \beta^{(t)}=\infty$ \label{lemma:betta_stuff:item2}
    \end{enumerate}
\end{lemma}
\begin{proof}
For \Cref{lemma:betta_stuff:item1}, recall that $\beta^{(t)}=\eta_t\cl_{|S}(\bw^{(t)})g^{2\nu_{Z}-2}(t)$. So while $g^{2\nu_{Z}-2}(t)$ is polynomial in $\norm{\bw^{(t)}}=g(t)$, $\cl_{|S}(\bw^{(t)})$ is exponentially decaying in $g(t)$. Since $g(t)\to\infty$ we have $\beta^{(t)}\to 0$. For \Cref{lemma:betta_stuff:item2} let us analyze the radial increments $\Delta g^{\nu_{Z}}(t):=g^{\nu_{Z}}(t+1)-g^{\nu_{Z}}(t)$. Recall that $\Delta\bw^{(t)}=-\eta_t\nabla\cl(\bw^{(t)})$. Using tailor expansion on the norm we get
\begin{equation} \label{eq:norm_taylor_expansion}
   g(t+1)=\norm{\bw^{(t)}+\Delta\bw^{(t)}}=g(t)+\inner{\bar \bw^{(t)}, \Delta\bw^{(t)}}+ O\left(\frac{\norm{\Delta\bw^{(t)}}^2}{g(t)}\right)
\end{equation}

Now using taylor expansion on $g(t)\mapsto g(t)^{\nu_{Z}}$ for a small increment
\begin{align*}
g(t+1)^{\nu_{Z}}&=g(t)^{\nu_{Z}}+\nu_{Z}g(t)^{\nu_{Z}-1}(g(t+1)-g(t)) + O\left(g(t)^{\nu_{Z}-2}(g(t+1)-g(t))^2\right) \\
&=g(t)^{\nu_{Z}}+\nu_{Z}g(t)^{\nu_{Z}-1}\inner{\bar \bw^{(t)}, \Delta\bw^{(t)}} + O\left(g(t)^{\nu_{Z}-2}\norm{\Delta\bw^{(t)}}^2\right) &&\text{(by \Cref{eq:norm_taylor_expansion})}
\end{align*}
Now note that
\begin{align*}
\nu_{Z}g(t)^{\nu_{Z}-1}\inner{\bar \bw^{(t)}, \Delta\bw^{(t)}}&=\eta_t \cl_{|S}(\bw^{(t)})\nu_{Z}g(t)^{2\nu_{Z}-2}\left(\inner{\frac{\bw^{(t)}}{\norm{\bw^{(t)}}}, \sum_{i\in S}\alpha_i^{(t)}C_i \nabla m_i^{\nu_{Z}}(\bar\bw^{\infty})}+o(1)\right) \\
&=\beta^{(t)}\nu_{Z}\left(\inner{\frac{\bw^{(t)}}{\norm{\bw^{(t)}}}, \sum_{i\in S}\alpha_i^{(t)} C_i\nabla m_i^{\nu_{Z}}(\bar\bw^{\infty})}+o(1)\right) && \text{(def $\beta^{(t)}$)}\\
&\leq\beta^{(t)}\nu_{Z}\left(\norm{\sum_{i\in S}\alpha_i^{(t)} C_i\nabla m_i^{\nu_{Z}}(\bar\bw^{\infty})}+o(1)\right) && \tiny{\text{(Cauchy-Schwartz)}}\\
&\leq\beta^{(t)}\nu_{Z}C_M\\
\end{align*}
For $C_M=1+\max_{i\in S}\norm{C_i \nabla m_i^{\nu_{Z}}(\bar\bw^{\infty})}$. As for the remainder:
\begin{align}
g^{\nu_{Z}-2}(t)\norm{\Delta\bw^{(t)}}^2 &=g^{\nu_{Z}-2}(t)\norm{\cl_{|S}(\bw^{(t)})\eta_tg^{\nu_{Z}-1}(t)\left(\sum_{i\in S} \alpha_i^{(t)}C_i\nabla m_i^{\nu_{Z}}(\bar\bw^{\infty}) + o(1)\right)}^2 \nonumber\\
&\leq g^{\nu_{Z}-2}(t)\left(\beta^{(t)}\right)^2\norm{\sum_{i\in S} \alpha_i^{(t)}C_i\nabla m_i^{\nu_{Z}}(\bar\bw^{\infty}) + o(1)}^2 \nonumber\\
&\leq g^{\nu_{Z}-2}(t)\left(\beta^{(t)}\right)^2C_M^2 \nonumber\\
&=o(\beta^{(t)}) \nonumber
\end{align}
Where the last inequality is due to $\beta^{(t)}$ decaying exponentially with $g^{(t)}$. Combining both bounds we get

\[
g(t+1)^{\nu_{Z}}-g(t)^{\nu_{Z}}\leq \nu_{Z}C_M\beta^{(t)}+o(\beta^{(t)})\leq \tilde C \beta^{(t)}
\]
For some $\tilde C>0$. Now summing from large enough $t$ to $\infty$ we get $\sum_{t=t_0}^{\infty}g(t+1)^{\nu_{Z}}-g(t)^{\nu_{Z}} =  \infty$ and hence also $\sum_{t=t_0}^\infty\tilde C \beta^{(t)} = \infty$.
\end{proof}
Using \Cref{lemma:perturbed_exponentiated_grad} with $\Phi(\alpha)=\frac{1}{2}\alpha^\top A \alpha=\frac{1}{2}\norm{\sum_{i\in S}\alpha_i^{(t)}C_i \nabla m_i^{\nu_{Z}}(\bar\bw^{\infty})}^2$, and \Cref{lemma:betta_stuff:item1} of \Cref{lemma:beta_stuff} to substitute $\varepsilon^{(t)}=o(\beta^{(t)})$, we get our update \Cref{eq:alpha_update}. Hence summing up \Cref{eq:KL_ineq} from $t=0$ to $T-1$, using $\alpha^*$ some minimizer of $\Phi$,  we get
\[
\sum_{t=0}^{T-1}\beta^{(t)}\left(\Phi(\alpha^{(t)})-\Phi(\alpha^{*})\right)\leq \kl{\alpha^*}{\alpha^{(0)}}+\sum_{t=0}^{T-1}(\beta^{(t)}\Delta^{(t)})^2/8 + 2\sum_{t=0}^{T-1}\norm{\varepsilon^{(t)}}_\infty
\]
Note that by the definition of $\Delta\bw^{(t)}$ we have that $\sum_{t=0}^{T-1}\frac{(\beta^{(t)}\Delta^{(t)})^2}{8} = o(\sum_{t=0}^{T-1}\beta^{(t)})$ and by \Cref{eq:alpha_update}  we have $2\sum_{t=0}^{T-1}\norm{\varepsilon^{(t)}}_\infty=o(\sum_{t=0}^{T-1}\beta^{(t)})$, so dividing both sides by $\sum_{t=0}^{T-1}\beta^{(t)}$ we get
\[
\frac{\sum_{t=0}^{T-1}\beta^{(t)}\left(\Phi(\alpha^{(t)})-\Phi(\alpha^{*})\right)}{\sum_{t=0}^{T-1}\beta^{(t)}} \to 0
\]
Thereby $\lim\inf_{t\to\infty}\left(\Phi(\alpha^{(t)})-\Phi(\alpha^*)\right)=0$, so there exists a subsequence $t_k$ s.t. $\Phi(\alpha^{t_k})\to\Phi(\alpha^*)$.

Notice that since $\forall i\in S$ $m_i$ has a minimal effective degree and the minimal margin among those with the minimal effective degree, hence we have that
\[
(\bar\bw^{\infty})^\top\sum_{i\in S}\alpha_i^{(t)} C_i\nabla m_i^{\nu_{Z}}(\bar\bw^{\infty})=\sum_{i\in S}\alpha_i^{(t)} \nu_{Z}C_im_i^{\nu_{Z}}(\bar\bw^{\infty})=\nu_{Z}\gamma_{\min}
\]
In other words,
\[
\sum_{i\in S}\alpha_i^{(t)}C_i \nabla m_i^{\nu_{Z}}(\bar\bw^{\infty})=(\nu_{Z}\gamma_{\min})\bar\bw^{\infty}+\bv^{(t)}
\] For $\bv^{(t)} \perp \bar\bw^{\infty}$ with a constant $\bar\bw^{\infty}$ element. So minimizing $\norm{\sum_{i\in S}\alpha_i^{(t)} C_i \nabla m_i^{\nu_{Z}}(\bar\bw^{\infty})}^2$ is akin to minimizing $\norm{\bv^{(t)}}^2$ for $\bv\in \conv_{i\in S} \{C_iP_\perp \nabla m_i^{\nu_{Z}}(\bar \bw^\infty)\}$ where $P_\perp$ the projection map to the space orthogonal to $\bar\bw^\infty$.
\begin{lemma} \label{lemma:minimal_projection}
    $\min_{\bv \in \conv_{i\in S} \{C_iP_\perp \nabla m_i^{\nu_{Z}}(\bar \bw^\infty)\}}\norm{\bv}^2=0$
\end{lemma}
\begin{proof}
    Assume towards contradiction that $0\notin \conv_{i\in S}\{C_iP_\perp \nabla m_i^{\nu_{Z}}(\bar\bw^{\infty})\}$, then by the separating hyperplane theorem there exists $\delta>0$ and $\bq \perp \bar\bw^{\infty}$ (since the whole convex hull lives orthogonally to $\bar \bw^\infty$) s.t. $\bq^\top  \bv\geq \delta$ for any $\bv\in \conv_{i\in S}\{C_iP_\perp \nabla m_i^{\nu_{Z}}(\bar\bw^{\infty})\}$. Recall that $\Delta\bw^{(t)}=\bw^{(t+1)}-\bw^{(t)}=\eta_t\cl_{|S}(\bw^{(t)})g^{\nu_{Z}-1}(t)\left(\sum_{i\in S} \alpha_i^{(t)}C_i\nabla m_i^{\nu_{Z}}(\bar\bw^{\infty}) + o(1)\right)$, hence for large enough $t_0$, we have for any $t\geq t_0$ that
    \begin{align*}
         \inner{\bq, \Delta \bw^{(t)}}&\geq (\delta/2)\eta_{t}\cl_{|S}(\bw^{(t)})g^{\nu_{Z}-1}(t)\\
        \inner{\bar\bw^\infty, \Delta \bw^{(t)}}& \leq \frac{3}{2}\nu_{Z}\gamma_{\min}\eta_{t}\cl_{|S}(\bw^{(t)})g^{\nu_{Z}-1}(t)
    \end{align*}
    Summing up we get
    \[
    \frac{\inner{\bq, \bw^{(t)}-\bw^{(t_0)}}}{\inner{\bar \bw^{\infty}, \bw^{(t)}-\bw^{(t_0)}}}\geq \frac{\delta\sum_{t=t_0}^T\eta_{t}\cl_{|S}(\bw^{(t)})g^{\nu_{Z}-1}(t)}{3\nu_{Z}\gamma_{\min}\sum_{t=t_0}^T\eta_{t}\cl_{|S}(\bw^{(t)})g^{\nu_{Z}-1}(t)}\overset{T\to\infty}{\to}\delta/3\nu_{Z}\gamma_{\min}>0
    \]
    Contradicting $\bar\bw^{(t)}\to \bar\bw^{\infty}$, since we got that $\frac{\bw^{(t)}-\bw^{(t_0)}}{\norm{\bw^{(t)}}}\to \bar \bw^{\infty}$ has a nonvanishing projection on $\bq\perp \bar \bw^{\infty}$.
\end{proof}
From \Cref{lemma:minimal_projection} we have that the subsequence $t_k$ has $\sum_{i\in S}\alpha_i^{(t_k)} C_i\nabla m_i^{\nu_{Z}}(\bar\bw^{\infty})\to \nu_{Z}\gamma_{\min}\bar \bw^\infty$.
Denote the subsequence limits $\alpha^{(t_k)}_i\to\alpha_i^{t_\infty}$, and $\overline{\nabla\cl(\bw^{(t_\infty)})}:=\lim_{k\to\infty}\frac{\nabla\cl(\bw^{(t_k)})}{\norm{\nabla\cl(\bw^{(t_k)})}}$.
So overall we have $\overline{\nabla\cl(\bw^{(t_\infty)})}\propto\sum_{i\in S}\alpha^{t_\infty}_i C_i\nabla \cp^{\nu_{Z}}_i(\bar \bw ^\infty)^\top y_i \x_i\propto \bar\bw^\infty$ for $\mu_i\geq 0$. We use $\alpha^{t_\infty}_i$ as the dual certificate (letting $\forall i\notin S$ $\alpha^{t_\infty}_i=0$), that satisfies dual feasibility and complementary slackness.  Therefore $\bar \bw ^{\infty} \propto \sum_{i=1}^{n}\alpha^{t_\infty}_i
C_i\nabla\cp_{i}^{\nu_i}(\bar \bw^{\infty})^\top \x_i y_i$. Scaling $\bar \bw ^{\infty}$ s.t. $\gamma_{\min}=1$, we conclude the proof.
\section{Auxiliary Material}
\begin{theorem}[Theorem 6.1 from \citet{bartlett}, simplified] \label{thm:bartlett:61}
Under \Cref{assmp:bartlett}, the expression
\begin{equation*}
\gamma(t):=\frac{\exp\left(\Phi\left(\exp\left(p\left(\norm{\bw^{(t)}}\right)\right)\cl\left(\bw^{(t)}\right)/n\right)\right)}{\norm{\bw^{(t)}}^M}
\end{equation*}
Where $M=\deg(q)$ is the degree of the predictor, and $\Phi(x)=\log(\log(1/nx))+2/\log(nx)$,
is positive and increasing from some time $t_0$ from \Cref{assmp:bartlett}.
\end{theorem}
\begin{theorem}[Theorem 6.2 from \citet{bartlett}, simplified] \label{thm:bartlett:62}
    Under \Cref{assmp:bartlett} the parameters $\bw^{(t)}$ converge in direction to a KKT point of
    \begin{equation*}
        \min \norm{\bw}^2,\quad\forall i, \quad y_if_H(\bw, \x_i)\geq 1
    \end{equation*}
    Where $f_H$ represents the top-degree homogenization of the margin (depending on $\x)$.
\end{theorem}
\begin{lemma} [Hoeffding's Lemma]\label{lemma:hoeffding}
Let $X$ be a RV over the reals, with $a\leq X\leq b$ almost surely. Then $\forall \lambda\in \reals$
\[
\E \exp(\lambda(X-\E X))\leq \exp(\lambda^2(b-a)^2/8)
\]
\end{lemma}

\begin{lemma}\label{lemma:recursion-convergence}
Let $(e_t)_{t=1}^\infty, (\beta_t)_{t=1}^\infty,(u_t)_{t=1}^\infty$, s.t.
\[
e_{t+1}=(1-\beta_t)e_t+u_t
\]
And $0<\beta_t<1$, $\sum_{t=1}^\infty \beta_t = \infty$, $\sum_{t=1}^\infty |u_t| < \infty$. Then $e_t \to 0$.
\end{lemma}
\begin{proof}
Iterating the recursion gives
\[
e_t
=
e_1\prod_{k=1}^{t-1}(1-\beta_k)
+
\sum_{j=1}^{t-1} u_j \prod_{k=j+1}^{t-1}(1-\beta_k),
\]
where an empty product is understood to be equal to $1$.

First,
\[
0 \leq \prod_{k=1}^{t-1}(1-\beta_k)
\leq
\prod_{k=1}^{t-1} e^{-\beta_k}
=
\exp\left(-\sum_{k=1}^{t-1}\beta_k\right)
\to 0,
\]
since $\sum_{k=1}^\infty \beta_k=\infty$. Hence
\[
e_1\prod_{k=1}^{t-1}(1-\beta_k)\to 0.
\]

It remains to treat the second term. Let $\varepsilon>0$. Since
$\sum_{j=1}^\infty |u_j|<\infty$, there exists $N$ such that
\[
\sum_{j=N+1}^\infty |u_j|<\varepsilon.
\]
Now split
\[
\sum_{j=1}^{t-1} u_j \prod_{k=j+1}^{t-1}(1-\beta_k)
=
\sum_{j=1}^{N} u_j \prod_{k=j+1}^{t-1}(1-\beta_k)
+
\sum_{j=N+1}^{t-1} u_j \prod_{k=j+1}^{t-1}(1-\beta_k).
\]

For each fixed $j$, we have
\[
\prod_{k=j+1}^{t-1}(1-\beta_k)
\leq
\exp\left(-\sum_{k=j+1}^{t-1}\beta_k\right)
\to 0,
\]
so the finite sum
\[
\sum_{j=1}^{N} u_j \prod_{k=j+1}^{t-1}(1-\beta_k)
\to 0.
\]

For the tail, since $0<1-\beta_k<1$,
\[
\left|
\sum_{j=N+1}^{t-1} u_j \prod_{k=j+1}^{t-1}(1-\beta_k)
\right|
\leq
\sum_{j=N+1}^{t-1} |u_j|
\leq
\sum_{j=N+1}^{\infty} |u_j|
<\varepsilon.
\]
Thus the second term tends to $0$. Combining this with the convergence of the first term, we conclude that
\[
e_t \to 0.
\]
\end{proof}

\section{Additional Experiments and Experimental Details}\label{Section:Details_Experiments}
For all experiments we use Pytorch (\citet{paszke2019pytorch}). We fix 20 seeds (for $i\in[4], j\in[5]$, $\text{Seeds}=\{i \cdot 10^{j-1}\}$), filter experiments that did not reach interpolation or those with loss higher than $10^{-3}$, and then plot averages of 5 seeds from the remaining experiments. Each training run is $30000$ epochs long. The training is run on RTX2080 GPUs, one for each experiment. Each run used an instance 8 cpus, and 12GB memory. We note that our experiments use relatively small data due to hardware memory and time constraints.
We use normalized gradient descent to accelerate convergence of the dynamics, since with standard gradient descent convergence requires many more epochs, and it does not affect the implicit bias.

\paragraph{Effective degree.}  To estimate the effective degree of each example, we track $\min\left(\abs{\frac{m_i(\bw^{(t)})}{m^{h}_i(\bw^{(t)})}},\abs{\frac{m^{h}_i(\bw^{(t)})}{m_i(\bw^{(t)})}}, \frac{\norm{\nabla m^{h}_i(\bw^{(t)})}}{ \norm{\nabla m_i(\bw^{(t)})}}, \frac{\norm{\nabla m_i(\bw^{(t)})}}{\norm{\nabla m^h_i(\bw^{(t)})}}\right)$ for each example $\x_i$ and $1\leq h\leq \deg(\cp_i)$, and choose the $h$ that maximizes this quantity. Asymptotically, we expect only the effective degree to not vanish, since it goes to a constant $C$. Otherwise, at least one of the ratios should go to zero.
\subsection{Linear Autoregressive}
 The data $\{\x_i,y_i\}_{i=1}^{10}$ is random gaussian with $\x_i\in\reals^{100}$ ($10\times$ overparameterization), where the labels are set by fixing a random vector of parameters $\bw$, each coordinate sampled i.i.d. from a gaussian of scale $10^{-19}$, applying it to every example $t_i$ times, and taking the result's sign. We use a learning rate of $3\cdot 10^{-3}$. We assign $t_i$ uniformly random from a set of indices $T\subset \naturals$. We run experiments with the  $T=\{1,2,5\},T=\{1,3,5\}, T=\{1,4\}, T=\{2,5\}$. Each of which we run with 20 different seeds, summing up to 80 experiments.
 \subsubsection*{$T=2, 5$}
 We note that the effective degree often does not equal the algebraic degree. This is significant, since our effective-degree estimator is biased toward higher degrees.
 \begin{figure}[H]
  \includegraphics[width=1\textwidth]{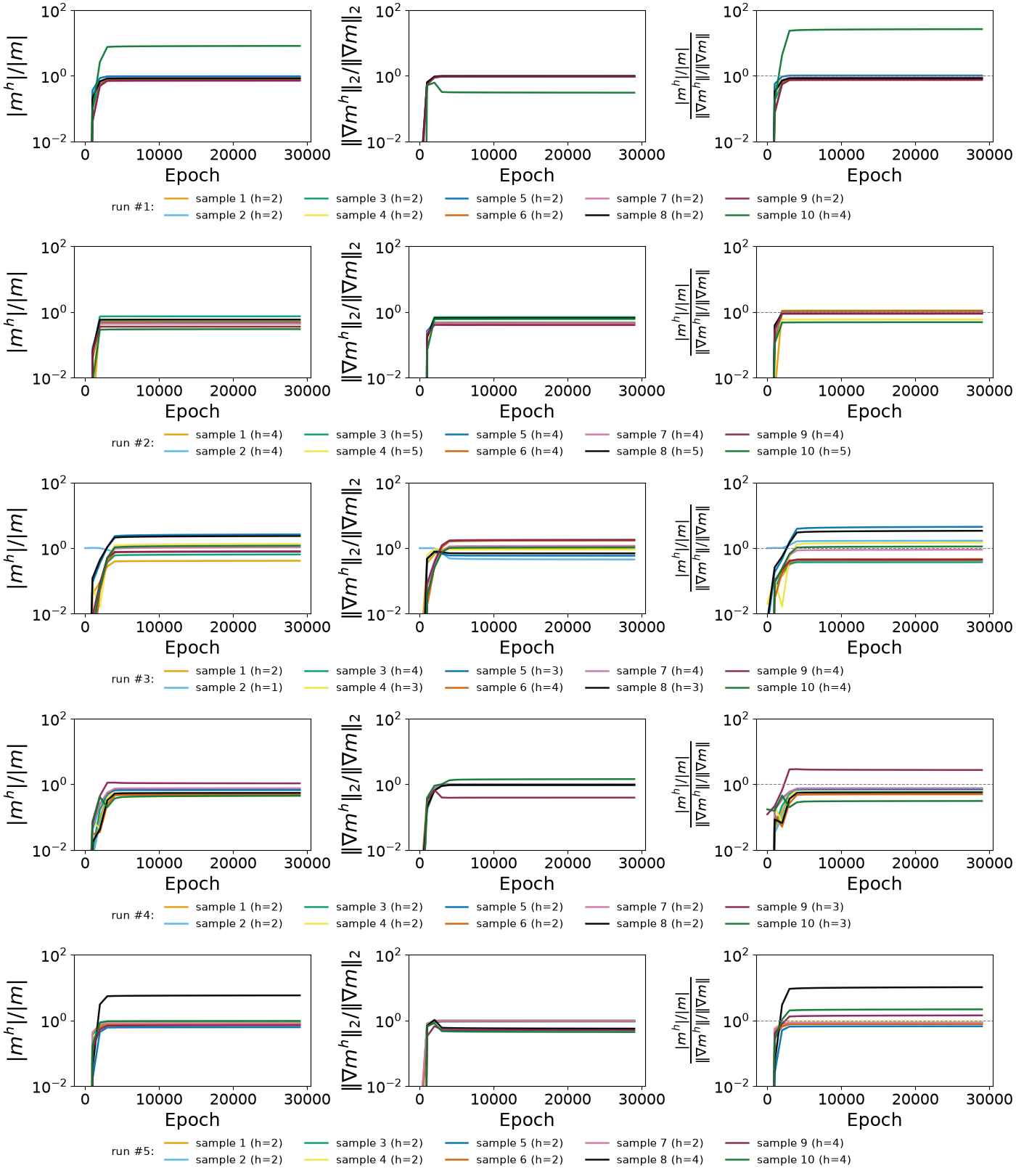}
  \caption{\textbf{Left:} Ratio between the effective-degree homogenization of the margin and the margin, $\abs{m_i^{h}/m_i}$. \textbf{Middle:} The corresponding ratio for the gradients, $\norm{\nabla m_i^{h}}/\norm{\nabla m_i}$. \textbf{Right:} The ratio of the two former ratios. Each row corresponds to a different seed and shows 10 random training examples, with $h$ the effective degree of each. Y-axes are log-scale. Figure corresponding to $T=2, 5$.}
\end{figure}
\begin{figure}[H]
  \includegraphics[width=1\textwidth]{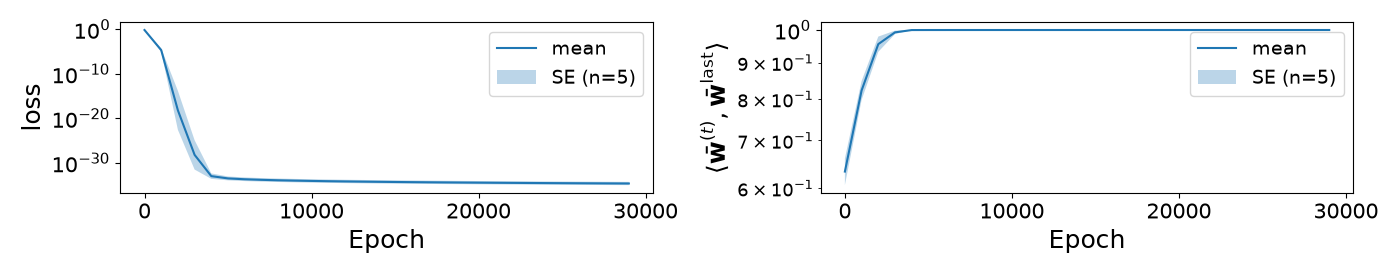}
  \caption{\textbf{Left:} The loss. \textbf{Right:} Cosine similarity to the final predictor direction. Y-axes are log-scale. Figure corresponding to $T=2, 5$.}
\end{figure}
\begin{figure}[H]
  \includegraphics[width=1\textwidth]{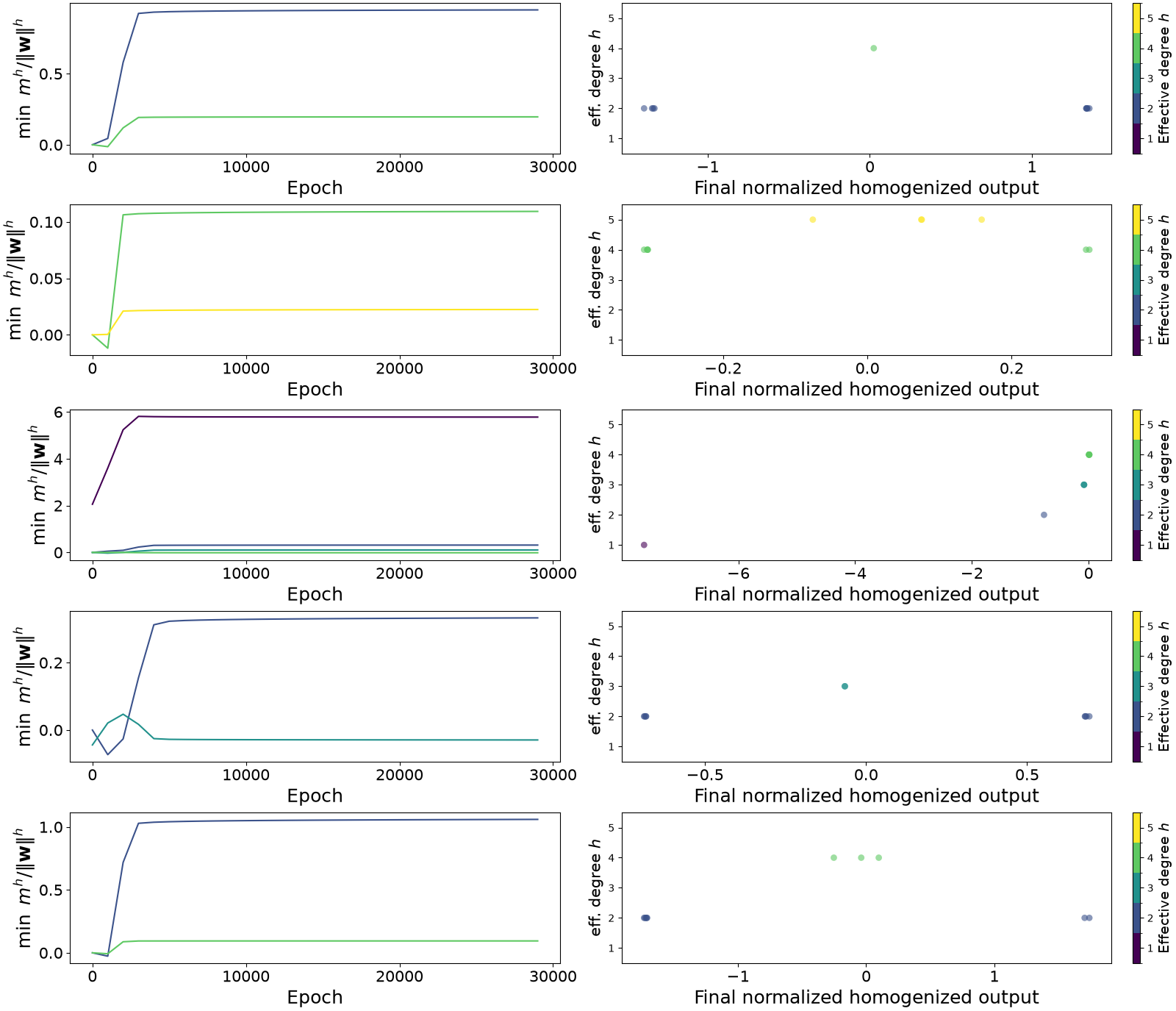}
  \caption{\textbf{Left:} Minimal normalized, homogenized margin per effective degree. \textbf{Right:} All points with their effective degrees and final normalized homogenized margins. Each row corresponds to a different seed. Figure corresponding to $T=2, 5$.}
\end{figure}
\subsubsection*{$T=1,4$}
\begin{figure}[H]
  \includegraphics[width=1\textwidth]{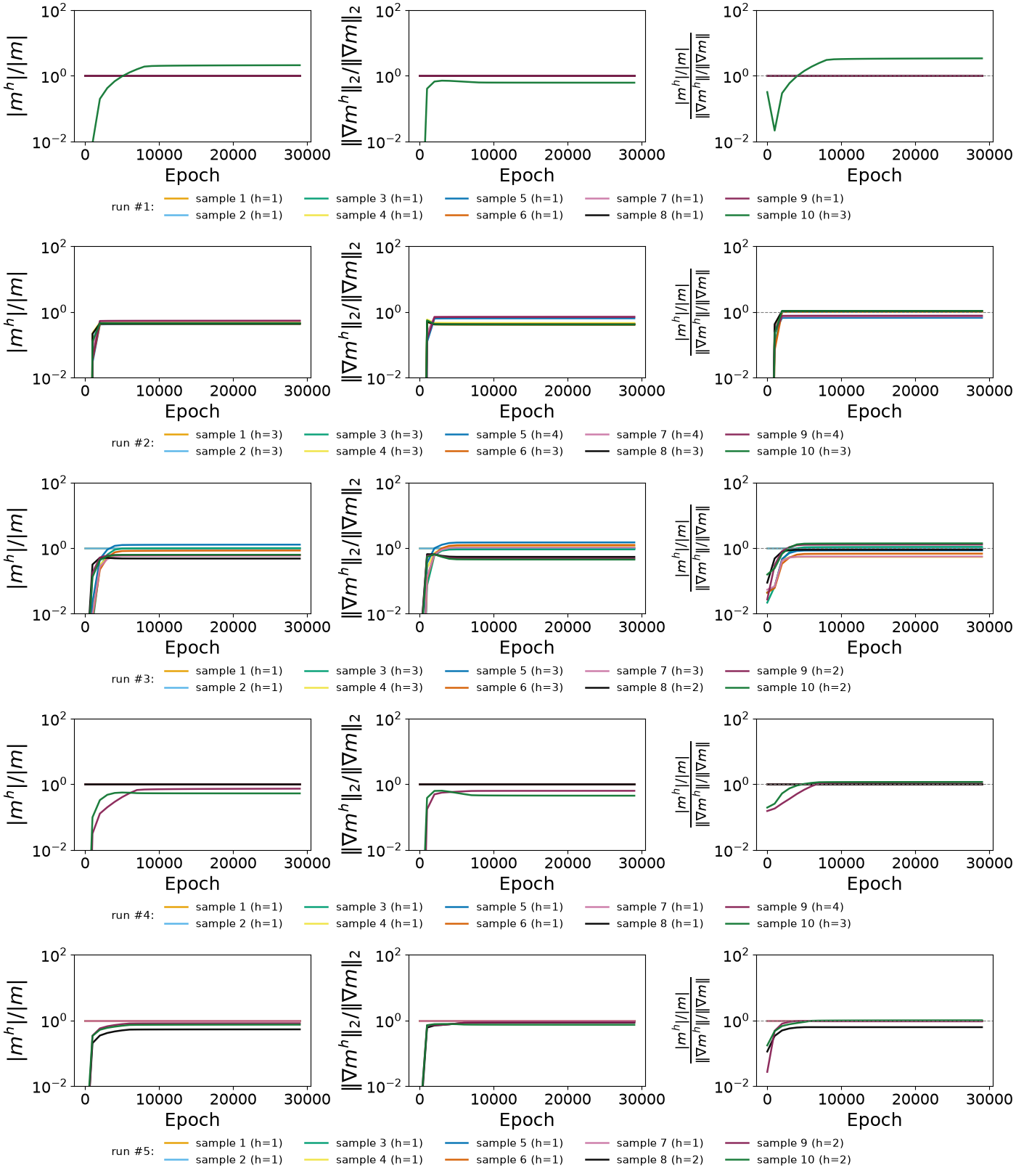}
  \caption{\textbf{Left:} Ratio between the effective-degree homogenization of the margin and the margin, $\abs{m_i^{h}/m_i}$. \textbf{Middle:} The corresponding ratio for the gradients, $\norm{\nabla m_i^{h}}/\norm{\nabla m_i}$. \textbf{Right:} The ratio of the two former ratios. Each row corresponds to a different seed and shows 10 random training examples, with $h$ the effective degree of each. Y-axes are log-scale. Figure corresponding to $T=1,4$.}
\end{figure}
\begin{figure}[H]
  \includegraphics[width=1\textwidth]{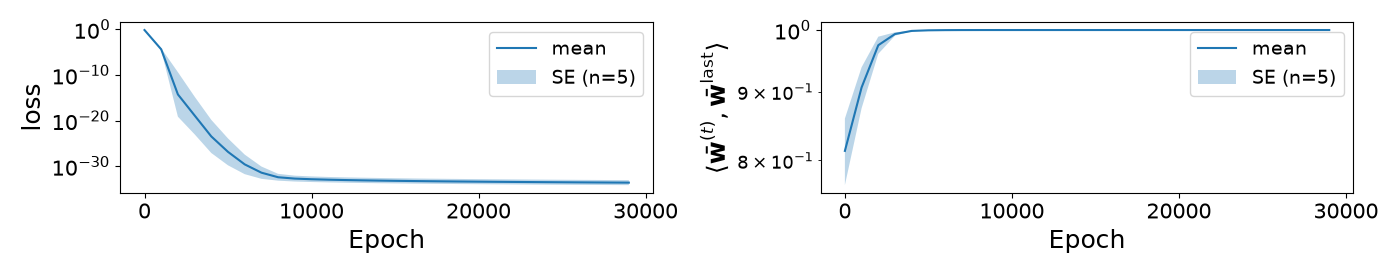}
  \caption{\textbf{Left:} The loss. \textbf{Right:} Cosine similarity to the final predictor direction. Y-axes are log-scale. Figure corresponding to $T=1,4$.}
\end{figure}
\begin{figure}[H]
  \includegraphics[width=1\textwidth]{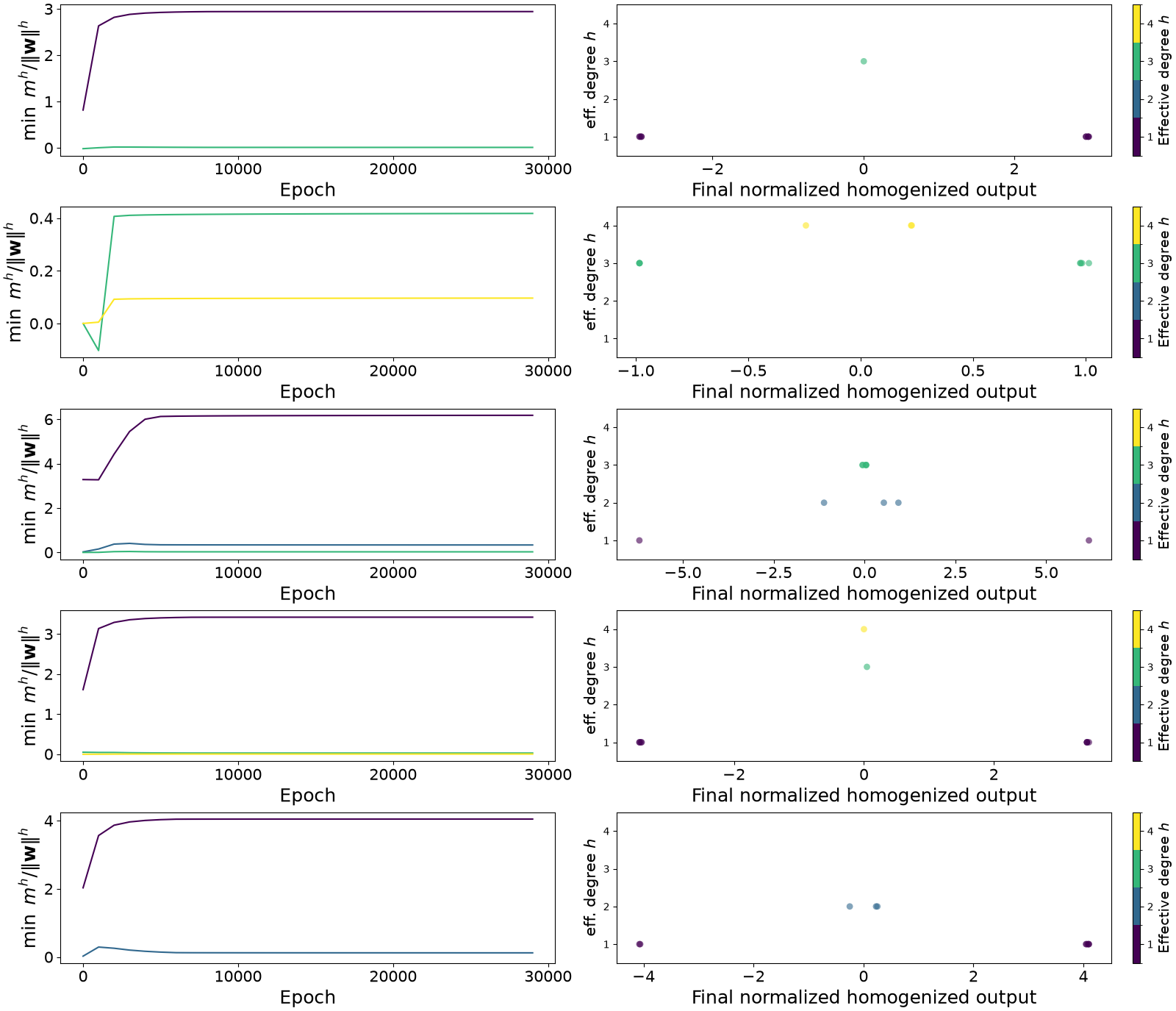}
  \caption{\textbf{Left:} Minimal normalized, homogenized margin per effective degree. \textbf{Right:} All points with their effective degrees and final normalized homogenized margins. Each row corresponds to a different seed. Figure corresponding to $T=1,4$.}
\end{figure}
\subsubsection*{$T=1,3,5$}
\begin{figure}[H]
  \includegraphics[width=1\textwidth]{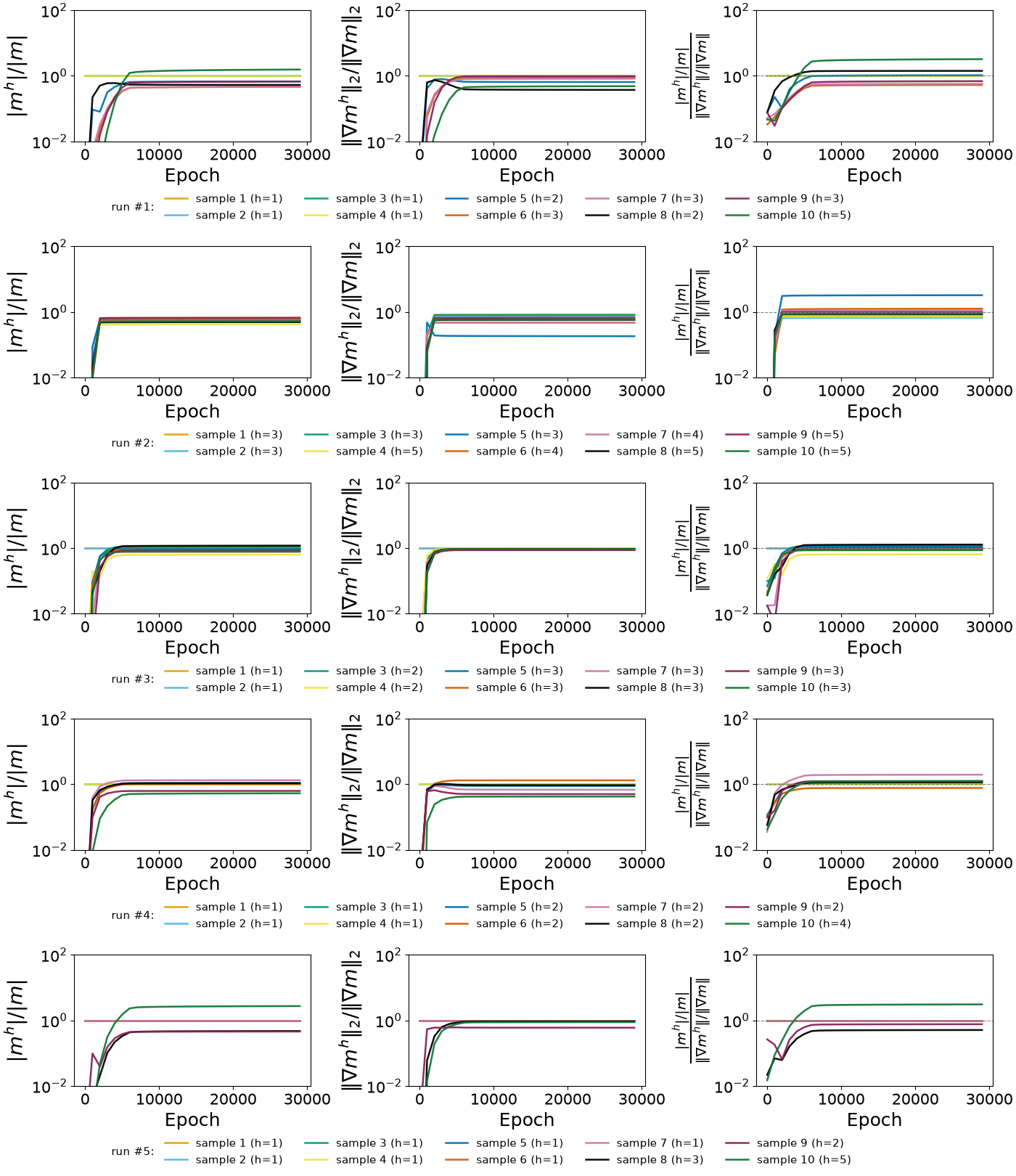}
  \caption{\textbf{Left:} Ratio between the effective-degree homogenization of the margin and the margin, $\abs{m_i^{h}/m_i}$. \textbf{Middle:} The corresponding ratio for the gradients, $\norm{\nabla m_i^{h}}/\norm{\nabla m_i}$. \textbf{Right:} The ratio of the two former ratios. Each row corresponds to a different seed and shows 10 random training examples, with $h$ the effective degree of each. Y-axes are log-scale. Figure corresponding to $T=1,3,5$.}
\end{figure}
\begin{figure}[H]
  \includegraphics[width=1\textwidth]{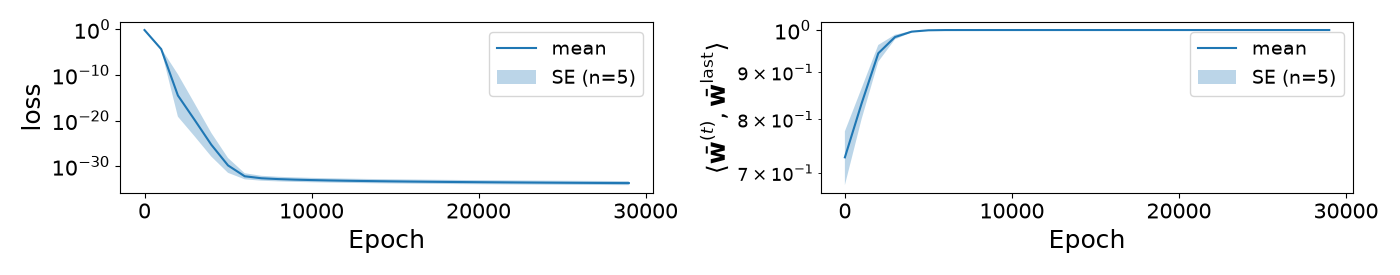}
  \caption{\textbf{Left:} The loss. \textbf{Right:} Cosine similarity to the final predictor direction. Y-axes are log-scale. Figure corresponding to $T=1,3,5$.}
\end{figure}
\begin{figure}[H]
  \includegraphics[width=1\textwidth]{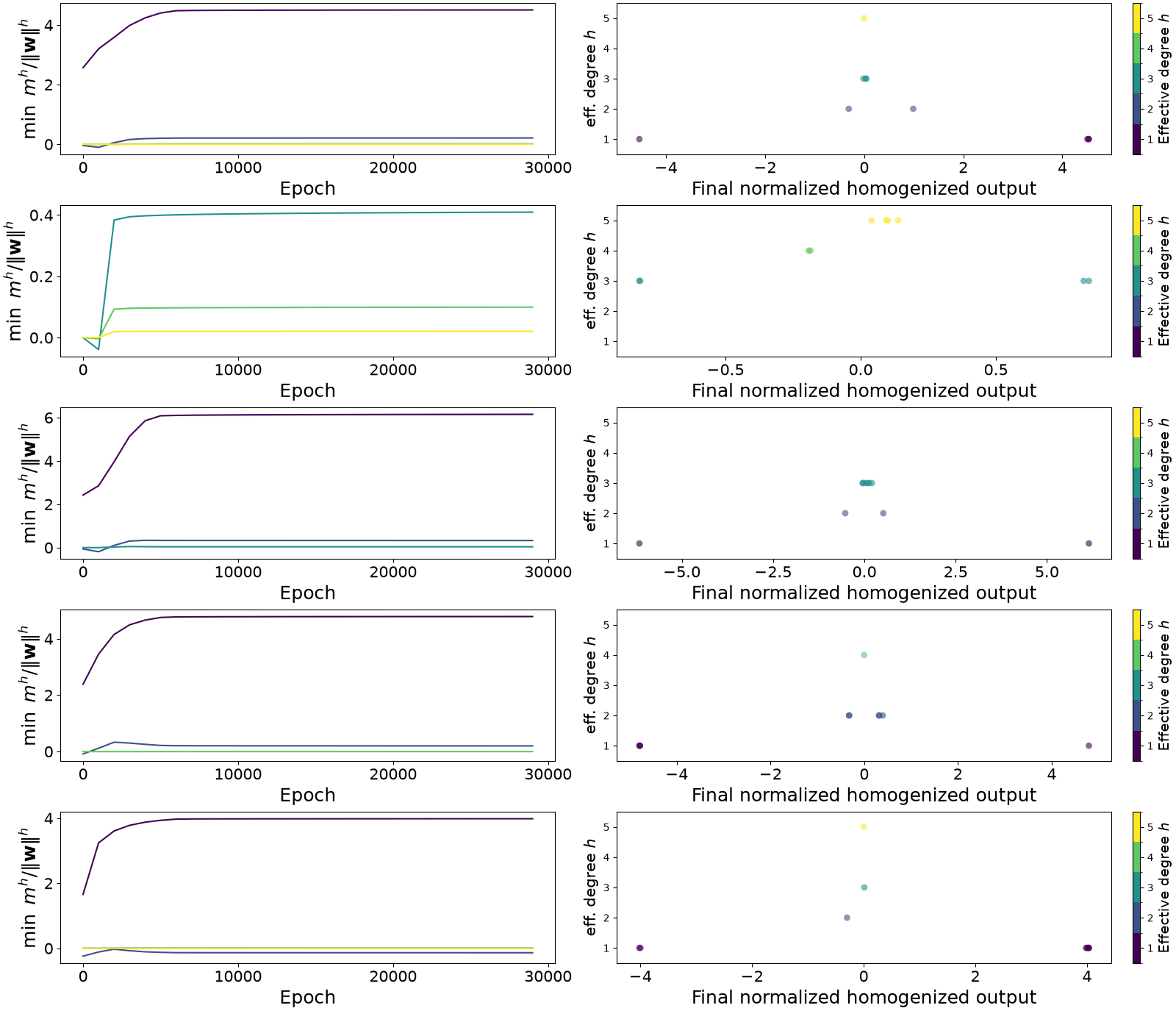}
  \caption{\textbf{Left:} Minimal normalized, homogenized margin per effective degree. \textbf{Right:} All points with their effective degrees and final normalized homogenized margins. Each row corresponds to a different seed. Figure corresponding to $T=1,3,5$.}
\end{figure}
\subsubsection*{$T=1,2,5$}
\begin{figure}[H]
  \includegraphics[width=1\textwidth]{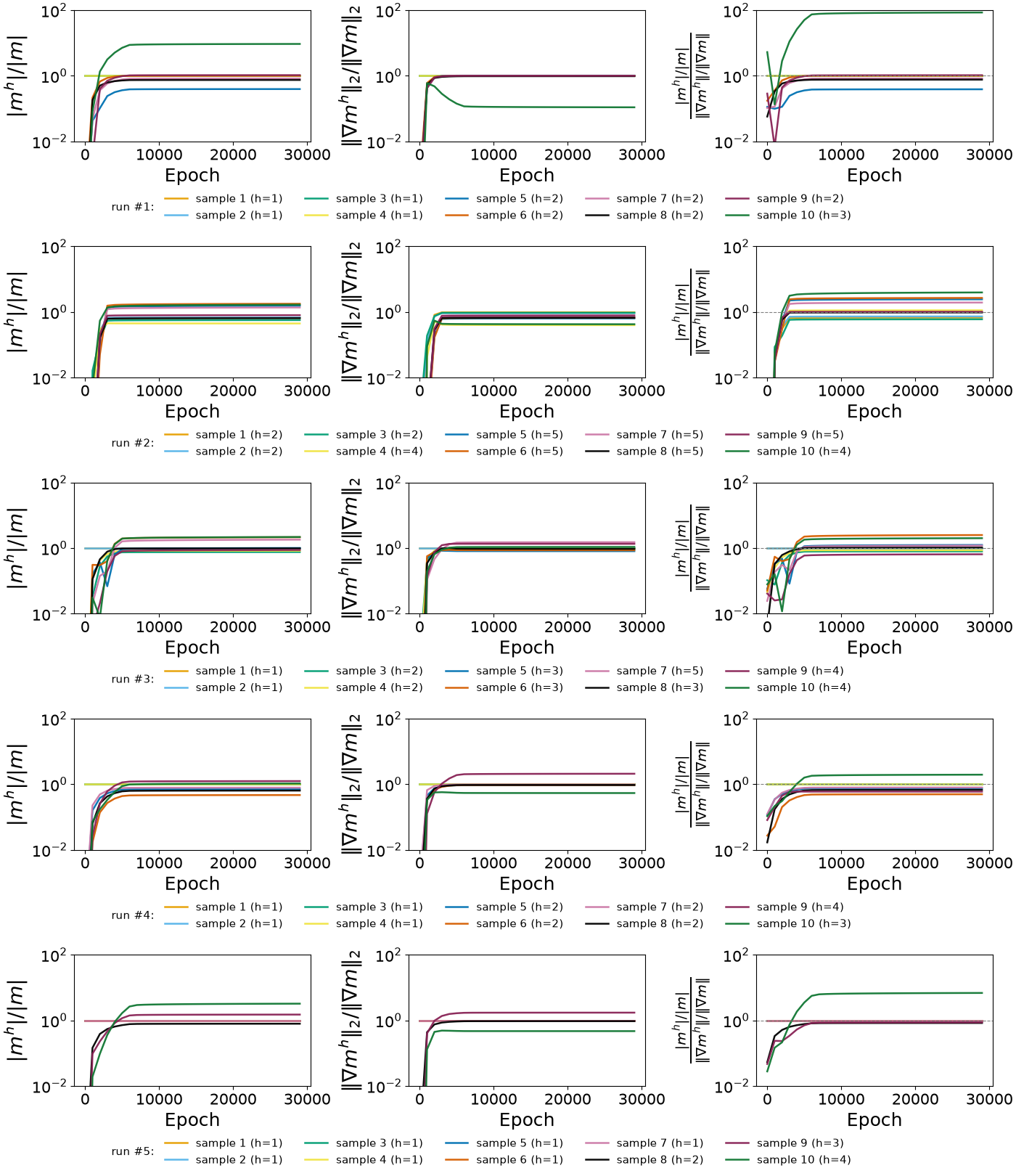}
  \caption{\textbf{Left:} Ratio between the effective-degree homogenization of the margin and the margin, $\abs{m_i^{h}/m_i}$. \textbf{Middle:} The corresponding ratio for the gradients, $\norm{\nabla m_i^{h}}/\norm{\nabla m_i}$. \textbf{Right:} The ratio of the two former ratios. Each row corresponds to a different seed and shows 10 random training examples, with $h$ the effective degree of each. Y-axes are log-scale. Figure corresponding to $T=1,2,5$.}
\end{figure}
\begin{figure}[H]
  \includegraphics[width=1\textwidth]{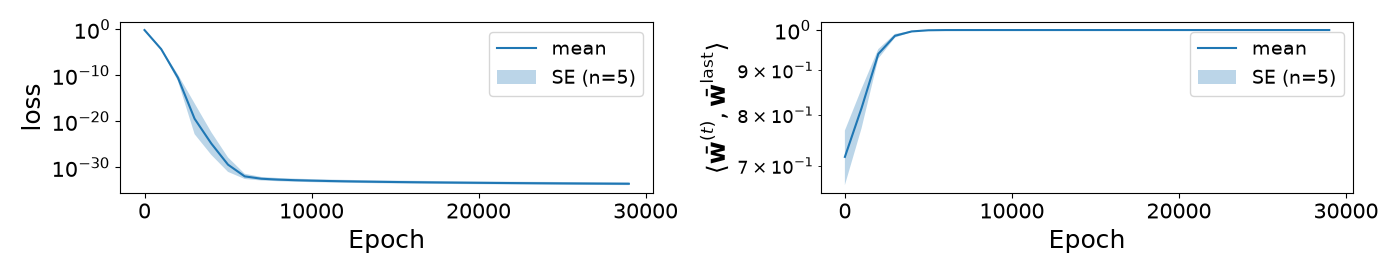}
  \caption{\textbf{Left:} The loss. \textbf{Right:} Cosine similarity to the final predictor direction. Y-axes are log-scale. Figure corresponding to $T=1,2,5$.}
\end{figure}
\begin{figure}[H]
  \includegraphics[width=1\textwidth]{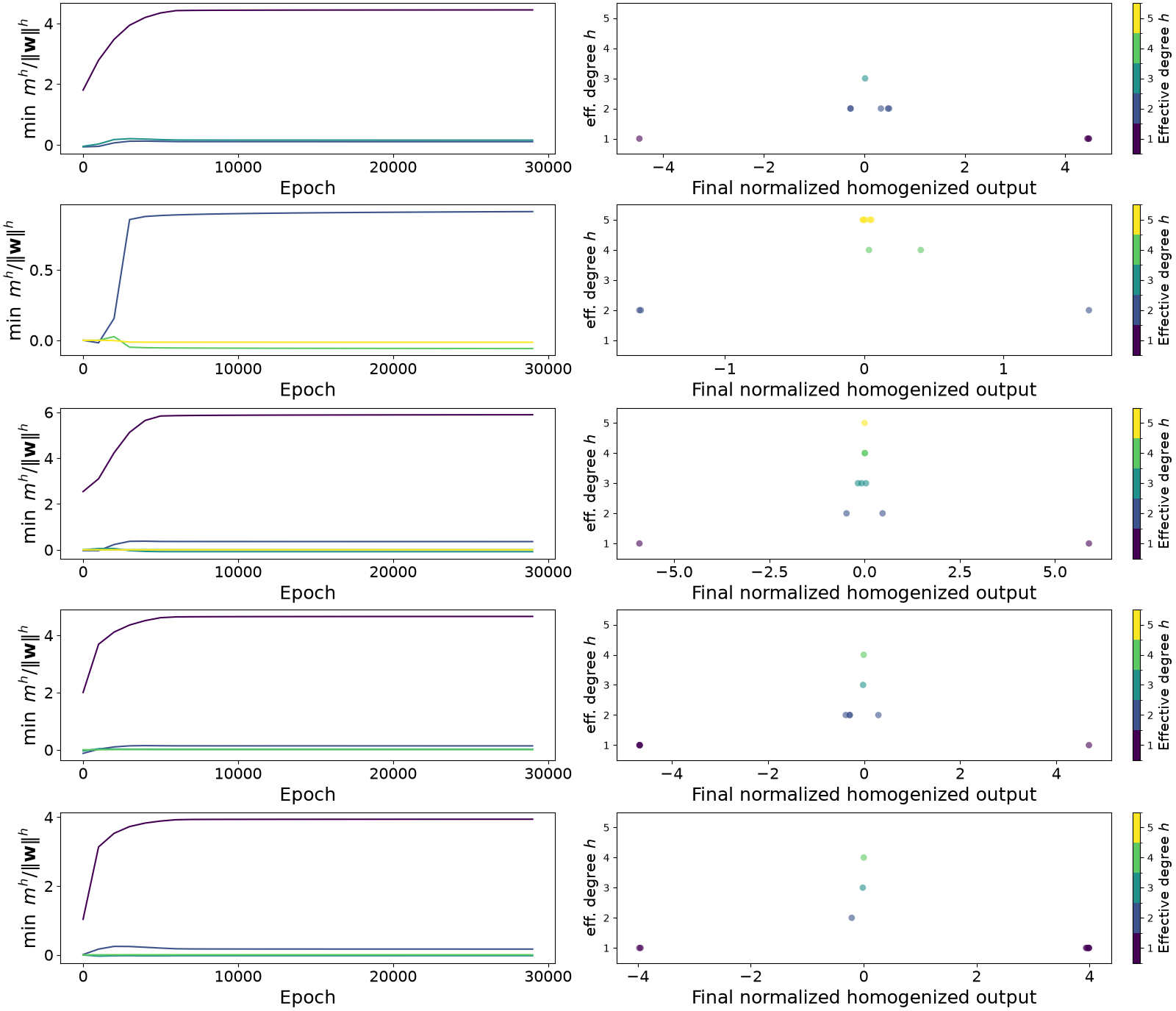}
  \caption{\textbf{Left:} Minimal normalized, homogenized margin per effective degree. \textbf{Right:} All points with their effective degrees and final normalized homogenized margins. Each row corresponds to a different seed. Figure corresponding to $T=1,2,5$.}
\end{figure}
\subsection{SSM}
The data $\{X_i,y_i\}_{i=1}^{10}$ is random gaussian with $X_i\in\reals^{4\times l_i}$, where the labels are set by fixing a random projection vector $\bp\in\reals^4$ (which is equal to the ``student's'' $\cp$ and stays fixed during training), random parameters $A\in \reals^{100\times 100}$, $B\in \reals^{100\times 4}, C\in\reals^{100\times 4}$, each coordinate sampled i.i.d. from a gaussian of scale $10^{-19}$, applying it to every example $t_i$ times, and taking the result's sign. We assign $l_i,t_i$ uniformly random from a set of indices $T\subset \naturals\times\naturals$. We run experiments with the sets $T=\{(3,1),(3,2), (5,2)\},T=\{(1,1);(2,2);(3,3)\},T=\{(3,1),(2,2),(1,3)\}$. Each of which we run with 20 different seeds, summing up to 80 experiments.
\subsubsection*{$l,T=(3,1), (3,2), (5, 2)$}
\begin{figure}[H]
  \includegraphics[width=1\textwidth]{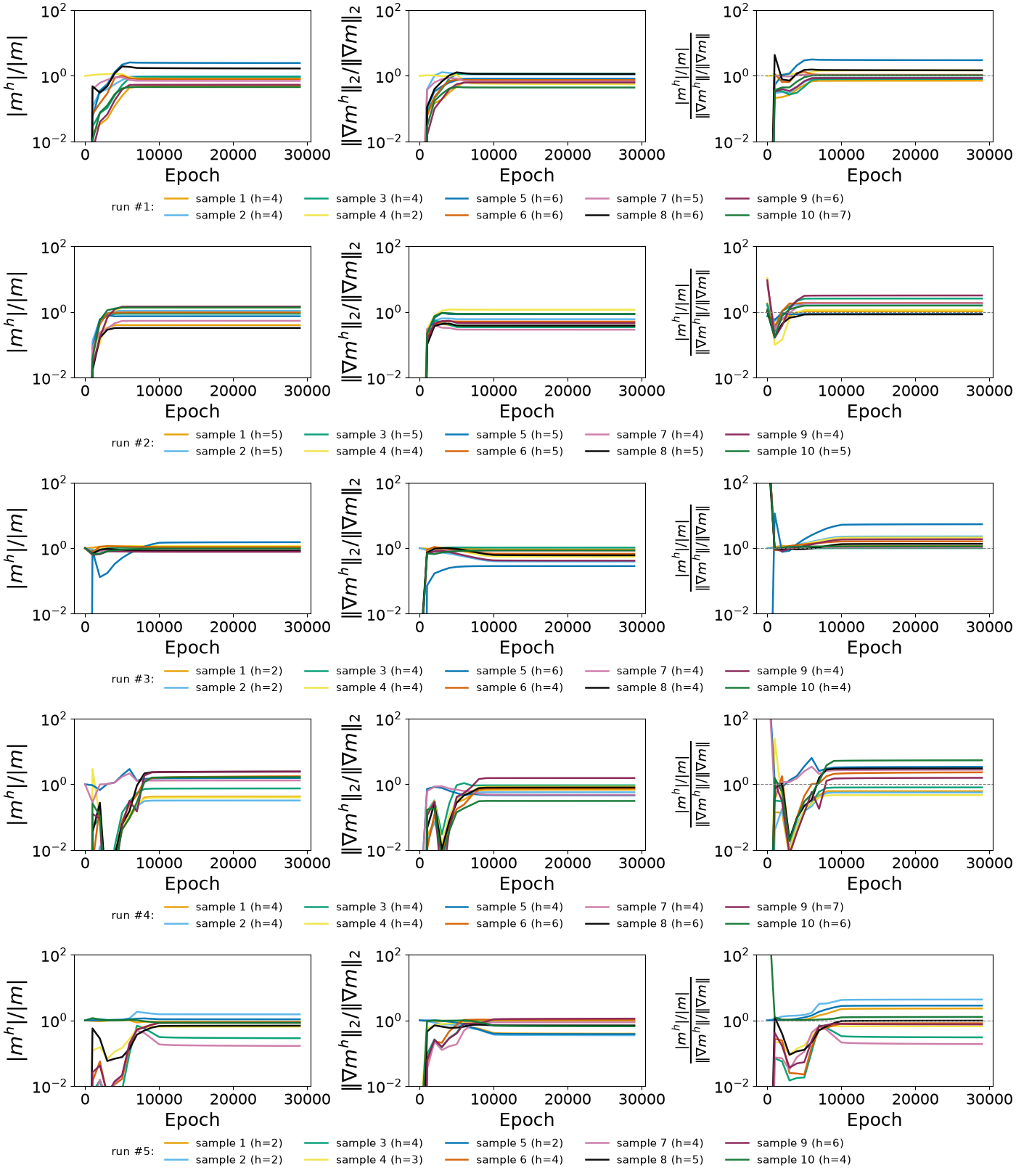}
  \caption{\textbf{Left:} Ratio between the effective-degree homogenization of the margin and the margin, $\abs{m_i^{h}/m_i}$. \textbf{Middle:} The corresponding ratio for the gradients, $\norm{\nabla m_i^{h}}/\norm{\nabla m_i}$. \textbf{Right:} The ratio of the two former ratios. Each row corresponds to a different seed and shows 10 random training examples, with $h$ the effective degree of each. Y-axes are log-scale. The figure is for $l,T=(3,1), (3,2), (5, 2)$.}
\end{figure}
\begin{figure}[H]
  \includegraphics[width=1\textwidth]{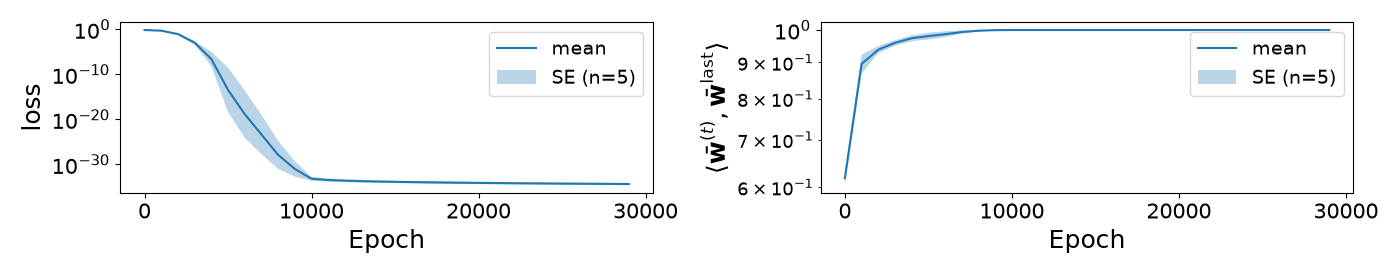}
  \caption{\textbf{Left:} The loss. \textbf{Right:} Cosine similarity to the final predictor direction. Y-axes are log-scale. The figure is for $l,T=(3,1), (3,2), (5, 2)$.}
\end{figure}
\begin{figure}[H]
  \includegraphics[width=1\textwidth]{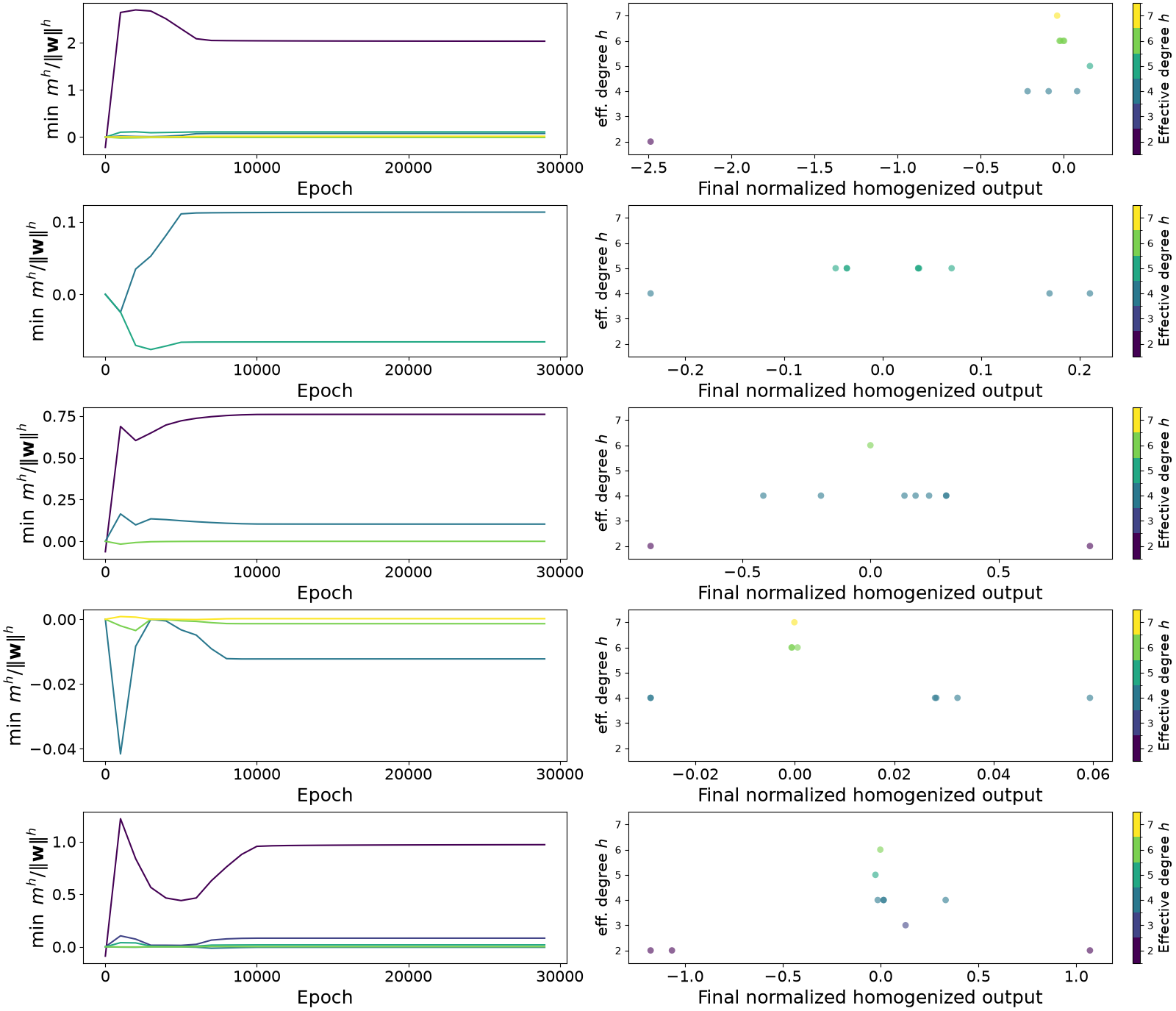}
  \caption{\textbf{Left:} Minimal normalized, homogenized margin per effective degree. \textbf{Right:} All points with their effective degrees and final normalized homogenized margins. Each row corresponds to a different seed. The figure is for $l,T=(3,1), (3,2), (5, 2)$.}
\end{figure}

\subsubsection*{$l,T=(1,1), (2,2), (3, 3)$}
\begin{figure}[H]
  \includegraphics[width=1\textwidth]{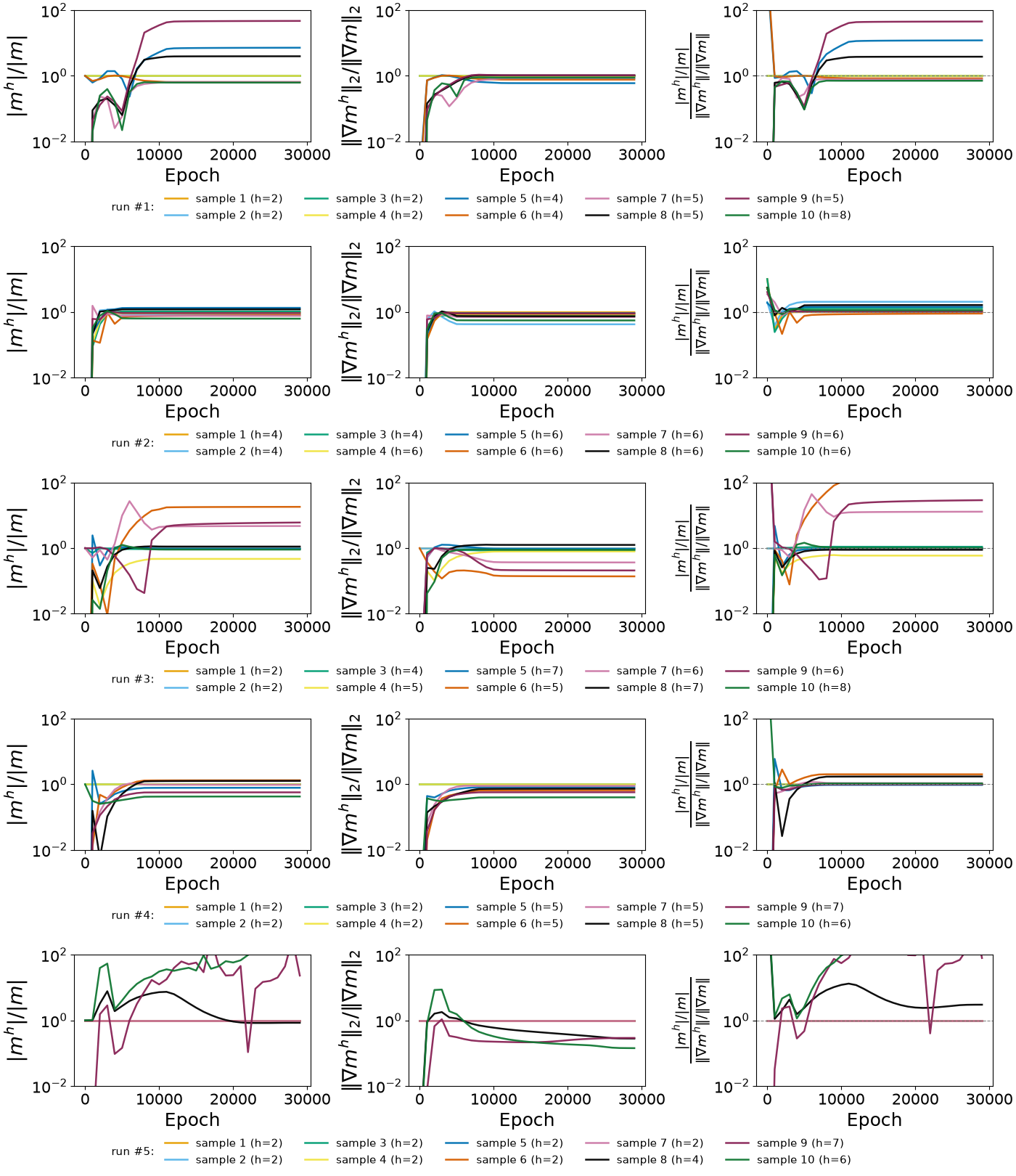}
  \caption{\textbf{Left:} Ratio between the effective-degree homogenization of the margin and the margin, $\abs{m_i^{h}/m_i}$. \textbf{Middle:} The corresponding ratio for the gradients, $\norm{\nabla m_i^{h}}/\norm{\nabla m_i}$. \textbf{Right:} The ratio of the two former ratios. Each row corresponds to a different seed and shows 10 random training examples, with $h$ the effective degree of each. Y-axes are log-scale. The figure is for $l,T=(1,1), (2,2), (3, 3)$.}
\end{figure}
\begin{figure}[H]
  \includegraphics[width=1\textwidth]{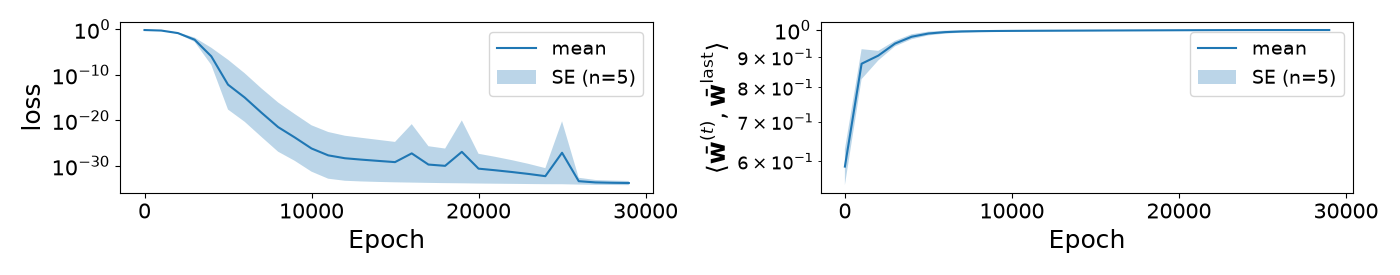}
  \caption{\textbf{Left:} The loss. \textbf{Right:} Cosine similarity to the final predictor direction. Y-axes are log-scale. The figure is for $l,T=(1,1), (2,2), (3, 3)$.}
\end{figure}
\begin{figure}[H]
  \includegraphics[width=1\textwidth]{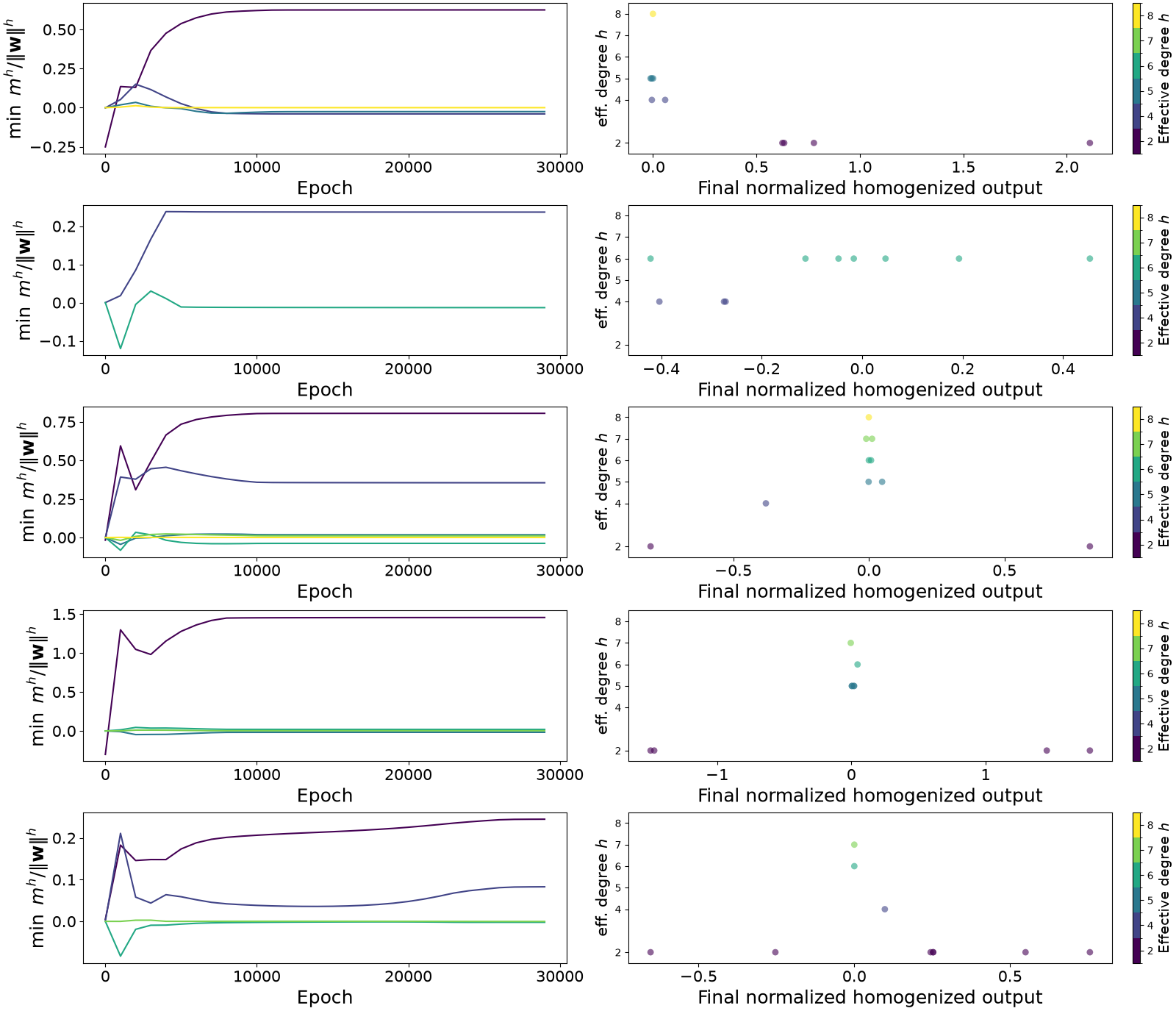}
  \caption{\textbf{Left:} Minimal normalized, homogenized margin per effective degree. \textbf{Right:} All points with their effective degrees and final normalized homogenized margins. Each row corresponds to a different seed. The figure is for $l,T=(1,1), (2,2), (3, 3)$.}
\end{figure}

\subsubsection*{$l,T=(3,1), (2,2), (1, 3)$}
\begin{figure}[H]
  \includegraphics[width=1\textwidth]{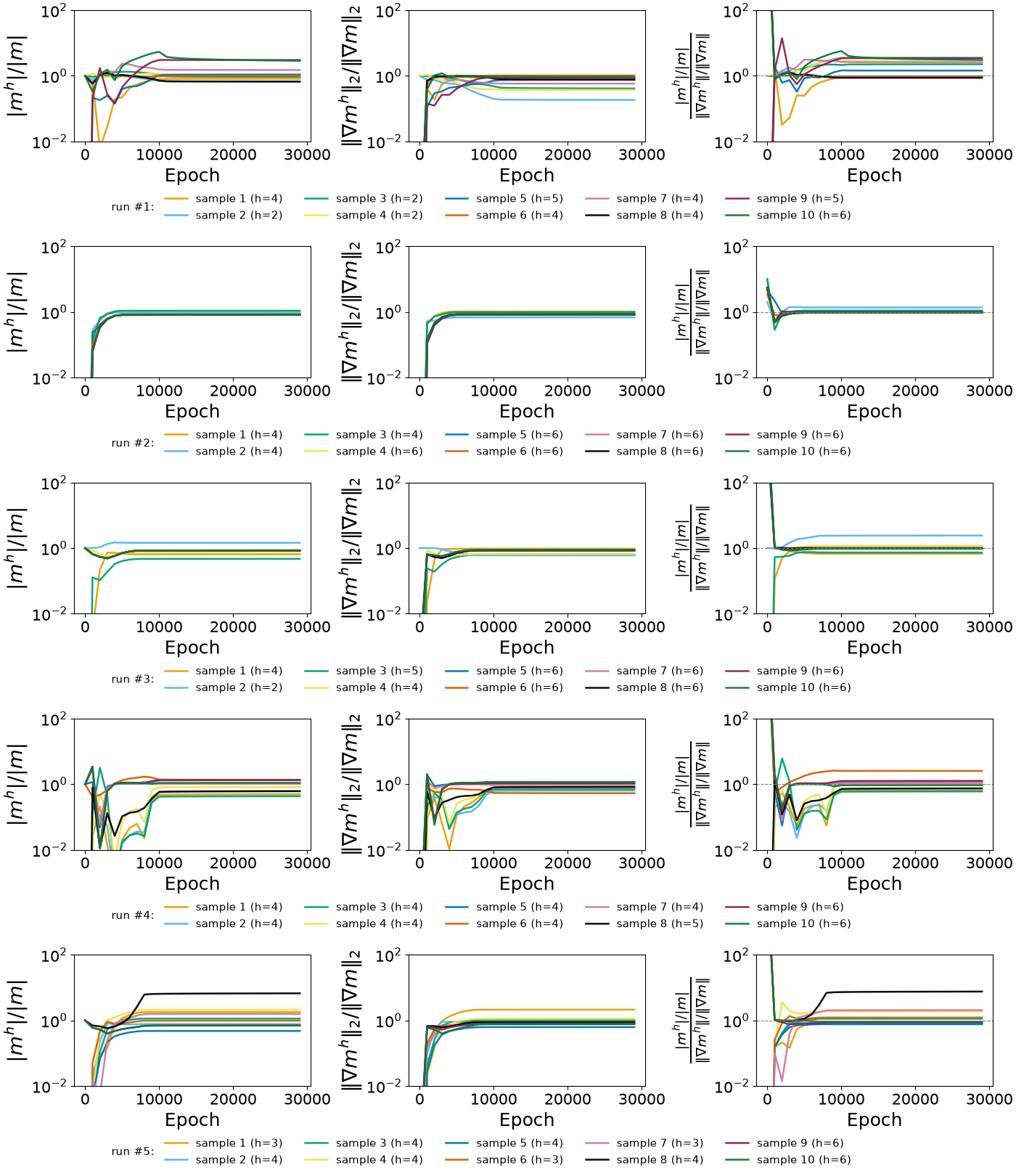}
  \caption{\textbf{Left:} Ratio between the effective-degree homogenization of the margin and the margin, $\abs{m_i^{h}/m_i}$. \textbf{Middle:} The corresponding ratio for the gradients, $\norm{\nabla m_i^{h}}/\norm{\nabla m_i}$. \textbf{Right:} The ratio of the two former ratios. Each row corresponds to a different seed and shows 10 random training examples, with $h$ the effective degree of each. Y-axes are log-scale. The figure is for $l,T=(3,1), (2,2), (1, 3)$.}
\end{figure}
\begin{figure}[H]
  \includegraphics[width=1\textwidth]{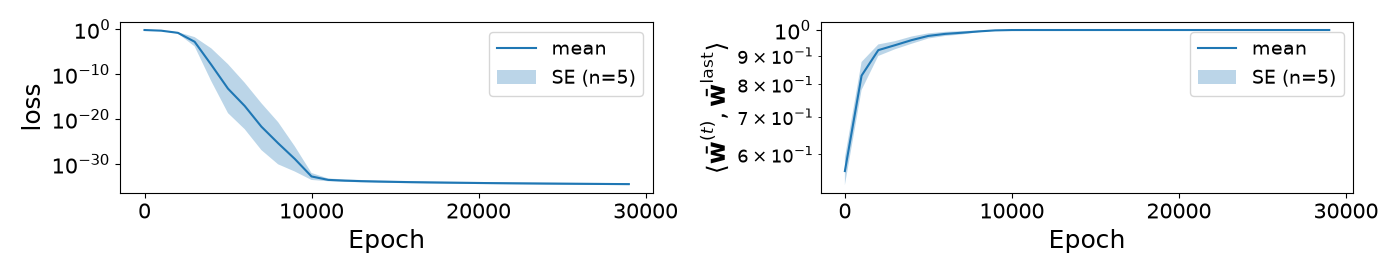}
  \caption{\textbf{Left:} The loss. \textbf{Right:} Cosine similarity to the final predictor direction. Y-axes are log-scale. The figure is for $l,T=(3,1), (2,2), (1, 3)$.}
\end{figure}
\begin{figure}[H]
  \includegraphics[width=1\textwidth]{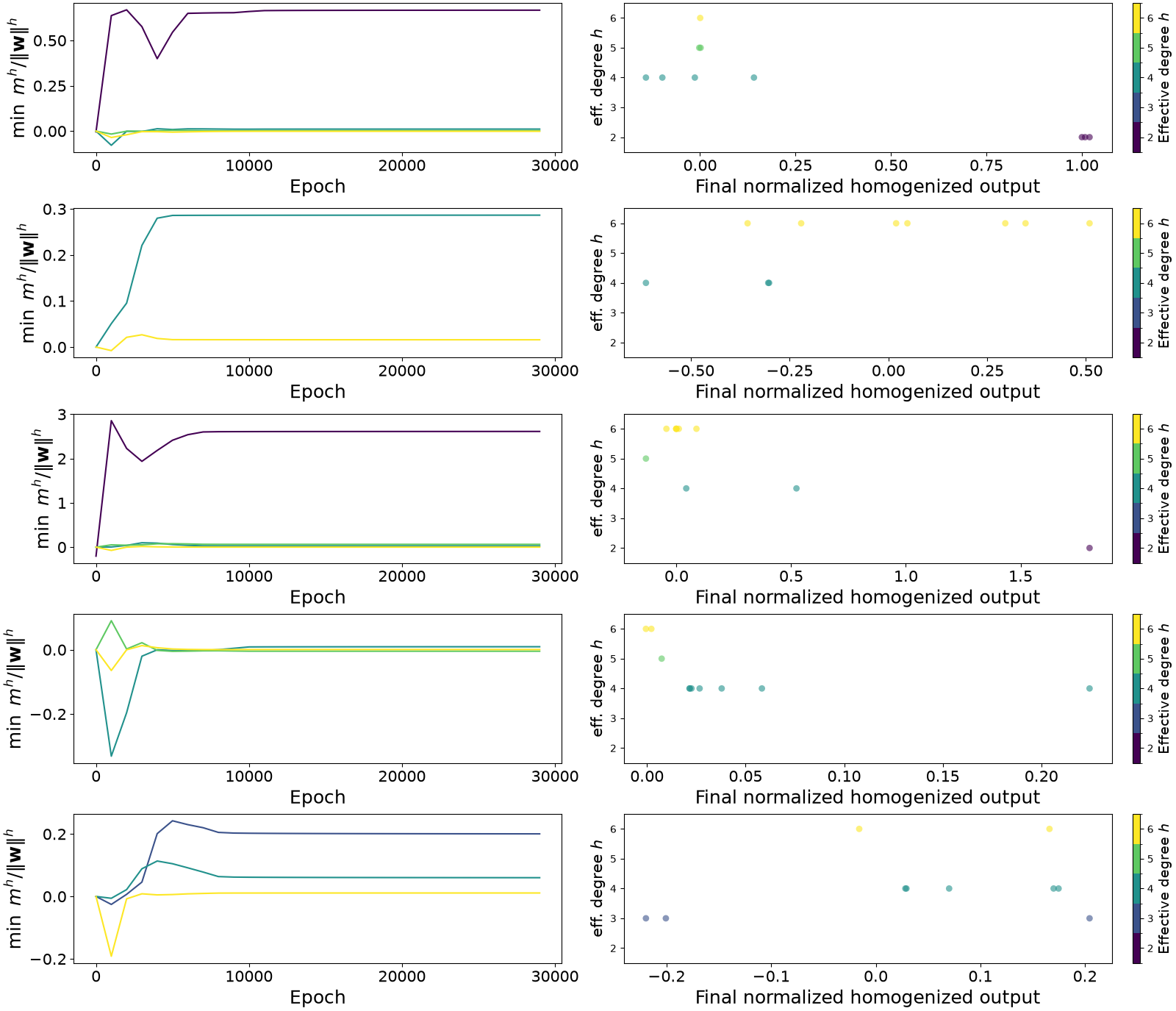}
  \caption{\textbf{Left:} Minimal normalized, homogenized margin per effective degree. \textbf{Right:} All points with their effective degrees and final normalized homogenized margins. Each row corresponds to a different seed. The figure is for $l,T=(3,1), (2,2), (1, 3)$.}
\end{figure}

\subsection{MNIST SSM}
We use data from MNIST \citep{lecun1998mnist}. The data $\{X_i,y_i\}_{i=1}^{10}$ divided to be half of images of the number $8$ (and positive label), and half of the number $1$ (and negative label). $X_i\in\reals^{28\times l_i}$,
$l_i$ is chosen uniformly at random from $\{14,21,28\}$, and we truncate the rows up to that point (data is first transposed so dimensions would fit). That is, we refer to the image's rows as a sequence, each row is a token of dimension $28$, and some examples dont use all the rows, in order for the result to have diverse effective degrees. Random parameters $A\in \reals^{64\times 64}$, $B\in \reals^{64\times 28}, C\in\reals^{64\times 28}$, $\bp \in\reals^{28}$, each coordinate sampled i.i.d. from a gaussian of scale $10^{-19}$. We run the experiment with 20 different seeds with learning rate $10^{-3}$.

\begin{figure}[H]
  \includegraphics[width=1\textwidth]{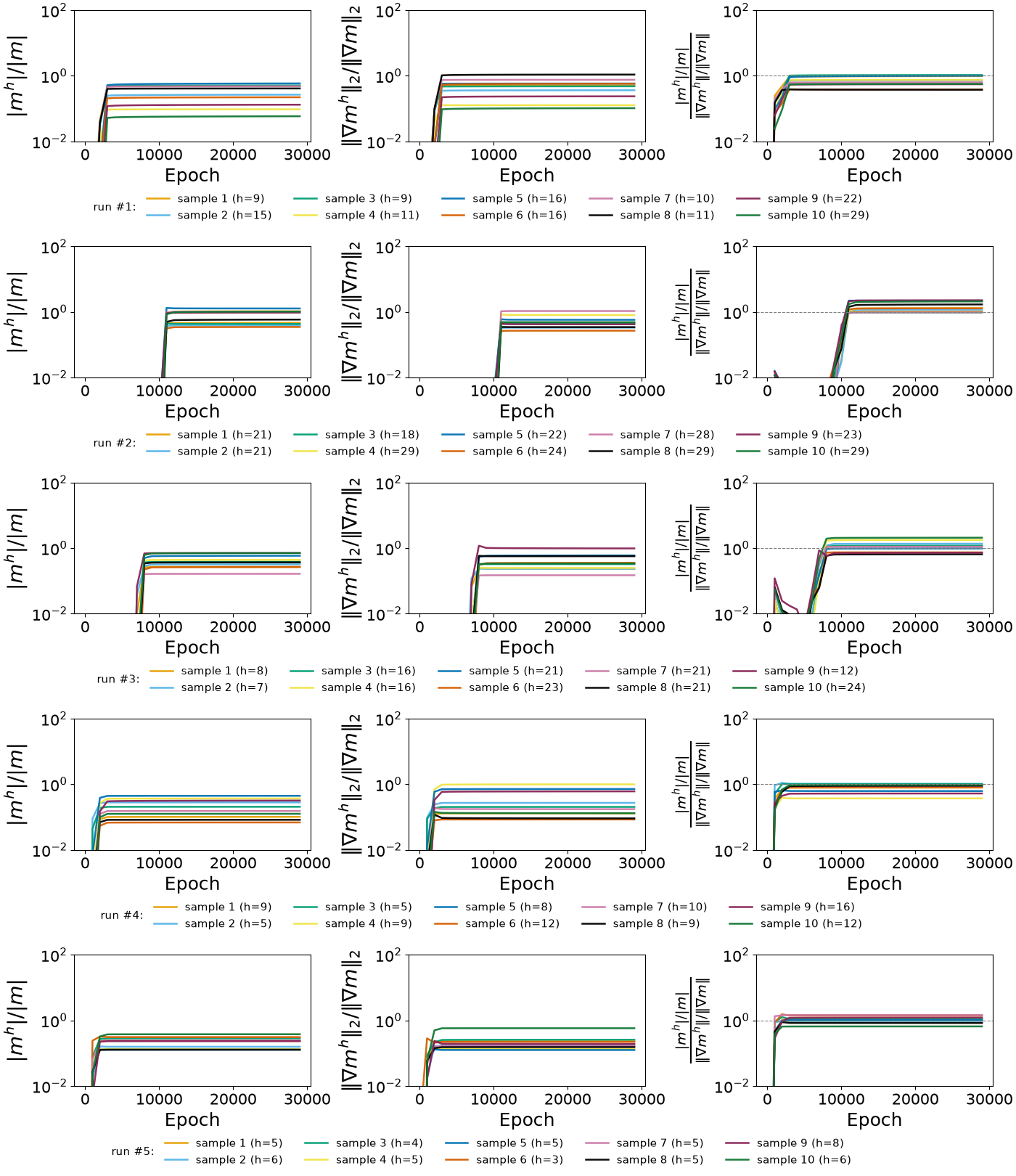}
  \caption{\textbf{Left:} Ratio between the effective-degree homogenization of the margin and the margin, $\abs{m_i^{h}/m_i}$. \textbf{Middle:} The corresponding ratio for the gradients, $\norm{\nabla m_i^{h}}/\norm{\nabla m_i}$. \textbf{Right:} The ratio of the two former ratios. Each row corresponds to a different seed and shows 10 random training examples, with $h$ the effective degree of each. Y-axes are log-scale. The figure is for SSM trained on MNIST with $l=\{14,21,28\}$.}
\end{figure}
\begin{figure}[H]
  \includegraphics[width=1\textwidth]{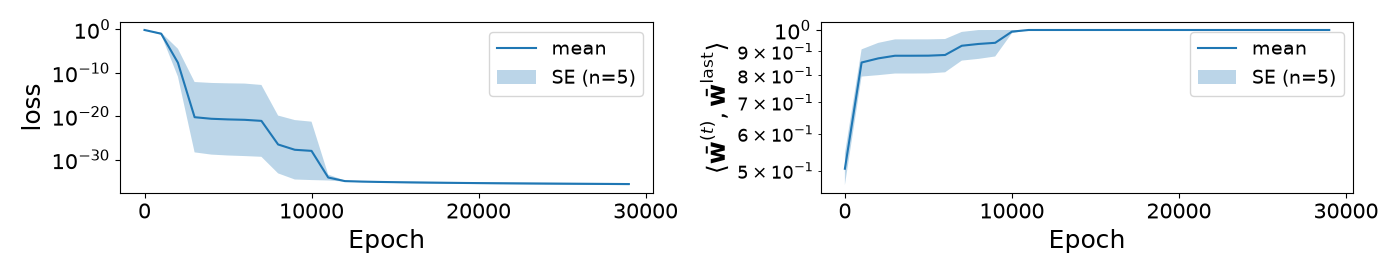}
  \caption{\textbf{Left:} The loss. \textbf{Right:} Cosine similarity to the final predictor direction. Y-axes are log-scale. The figure is for SSM trained on MNIST with $l=\{14,21,28\}$.}
\end{figure}
\begin{figure}[H]
  \includegraphics[width=1\textwidth]{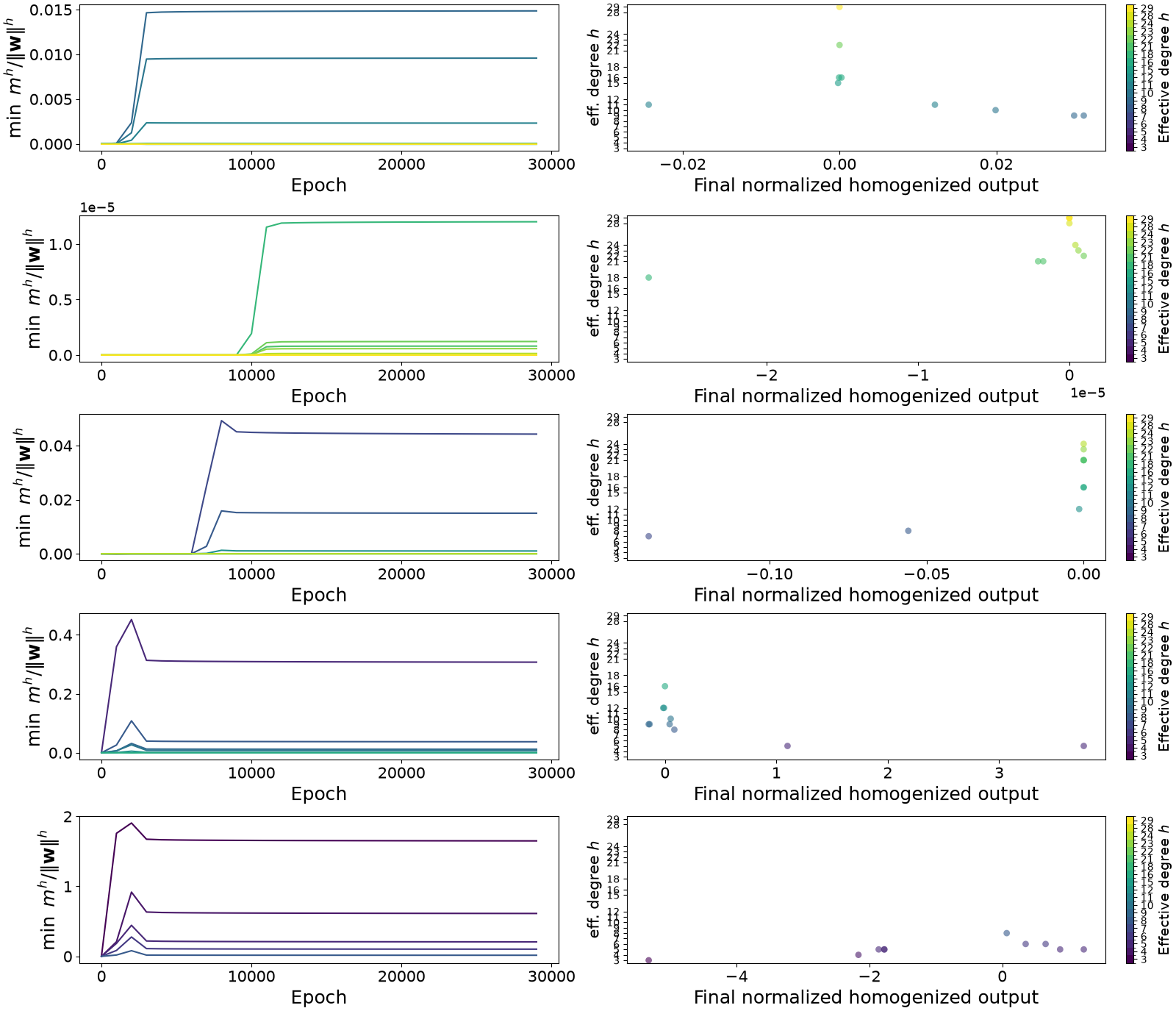}
  \caption{\textbf{Left:} Minimal normalized, homogenized margin per effective degree. \textbf{Right:} All points with their effective degrees and final normalized homogenized margins. Each row corresponds to a different seed. The figure is for SSM trained on MNIST with $l=\{14,21,28\}$.}
\end{figure}

\end{document}